# 硕士研究生学位论文

题目： 基于域泛化技术的水下机器人鲁棒目标检测

姓　　名： 宋品皓

学　　号： 1901213096

院　　系： 深圳研究生院

专　　业： 计算机应用技术

研究方向： 人机交互与机器人系统

导　　师： 刘宏 教授

二〇二二年六月

# 版权声明

# 摘要

目标检测旨在获得给定图片中特定物体的类别和位置，其包含两个子任务：分类和定位。近年来，研究者倾向于将目标检测应用在搭载了视觉系统的水下机器人上，以完成包括海产品捕捞、鱼类养殖、生物多样性监控等任务。然而水下的场景的多样性和复杂性，给目标检测技术提出了新的挑战。首先，生物群居的特性导致水下生物之间存在遮挡现象。其次，生物拟态导致了水下生物往往和环境的色调非常接近，使得目标和背景之间难以区分。再者，多样的水质环境和多变的、极端的光照条件，导致了水下相机得到了失真的、低对比度的、偏蓝或者偏绿的图像，不断变化的水质和光照引发了域迁移问题，而深度模型对于域迁移普遍是脆弱的。另外，水下机器人的运动会造成所捕获的图像出现模糊，也会导致周围水质混浊引起的可见度低的问题。本文针对了上述复杂水下环境中的目标检测问题展开研究，旨在设计一种高性能的、鲁棒的水下目标检测器，主要研究内容如下：

基于困难样本挖掘的思路提出了一种概率型二阶段水下目标检测器 Boosting R-CNN。构建了一个强力的一阶段检测器作为区域提议网络，给出较为准确的先验概率，并从贝叶斯的角度出发阐释二阶段目标检测，还原边缘分布，修正最终分类分数，有效提升模型准确率。最后，提出了一个困难样本挖掘方法——提升再权重模块，其可以从区域提议网络的错误中进行学习。当区域提议网络错误估计了某一个样本的先验概率，提升再权重模块会根据错误的程度放大其在最终损失的大小，集中训练了水下的困难样本，使模型更加鲁棒。实验证明，本方法在水下数据集能够达到更高的检测性能。

提出了建立通用性的水下目标检测器的概念，其意味着目标检测器一旦训练结束便可以无缝地在任何水域下实时使用。目前，由于水下数据集规模较小且采集困难，使模型无法获得足够的鲁棒性，阻碍了通用性的水下目标检测器的发展。本文使用了合成数据，揭示了由水质变化导致的模型脆弱性是一个亟待解决的关键问题，并且借助了数据增强、对抗训练和无关风险最小化理论构建了检测器 DG-YOLO，相比于其他方法在跨水质情况获得了极大的鲁棒性提升。

提出了一种域泛化训练范式，利用图像风格迁移和域混合操作，可以在域流形上由源域构成的域凸包中尽可能多地采样域数据，并且通过一个参数共享的骨干网络，对不同域的数据有选择性地进行正则化，以捕捉域无关的特征，构建一个强鲁棒性的水下检测器。实验结果表明，本文提出的方法在合成的域泛化水下数据集上能够实现最高的检测鲁棒性。

关键词：水下目标检测、概率建模、困难样本挖掘、域泛化、数据增强

# Robust Object Detection of Underwater Robot based on Domain Generalization

Pinhao Song (Computer Applied Technology)
Directed by Prof. Hong Liu

## ABSTRACT

Object detection aims to obtain the location and the category of specific objects in a given image, which includes two tasks: classification and location. In recent years, researchers tend to apply object detection to underwater robots equipped with vision systems to complete tasks including seafood fishing, fish farming, biodiversity monitoring and so on. However, the diversity and complexity of underwater environments bring new challenges to object detection. First, aquatic organisms tend to live together, which leads to severe occlusion. Second, the aquatic organisms are good at hiding themselves, which have the similar color with the background. Third, the various water quality and changeable and extreme lighting conditions lead to the distorted, low contrast, blue or green images obtained by the underwater camera, resulting in domain shift. And the deep model is generally vulnerable facing domain shift. Fourth, the movement of the underwater robot leads to the blur of the captured image and make the water muddy, which results in low visibility of the water. This paper investigates the problems brought by the underwater environment mentioned above, and aims to design a high-performance and robust underwater object detector. The main research contents are as follows:

Based on the idea of hard example mining, a probabilistic two-stage underwater object detector Boosting R-CNN is proposed. It builds a powerful one-stage detector as a Regional Proposal Network (RPN) to give accurate a prior probability. Interpreting the two-stage object detection from the perspective of Bayes, it restores the margin distribution, modifies the final classification scores, and effectively improves the accuracy of the model. Finally, a hard example mining method named Boosting Reweighting is proposed, which can help R-CNN learn from the error of RPN. When the RPN incorrectly estimates the aprior probability of a sample, the Boosting Reweighting will amplify its final loss according to the degree of error. The model who focuses more on training of the hard examples will be more robust. Experiments show that the Boosting R-CNN can achieve higher detection performance in underwater datasets.

General Underwater Object Detector (GUOD) is proposed, which means that the detector

can process in real time and be used in any water area if it is trained. Currently, because of the limited underwater dataset and the great difficulty of the data collection, the model can not obtain enough robustness, which hinders the development of underwater object detection. This paper uses synthetic data to reveal that the model's vulnerability caused by the changing of water qualities is an essential problem. The detector DG-YOLO is proposed, which leverages data augmentation, adversarial training and Invariant Risk Minimization (IRM) theory, greatly improving the cross-domain robustness compared with the baseline.

A domain generalization training paradigm is proposed, which uses the style transfer model and interpolation at the feature level to sample as much domain as possible in the domain convex hull constructs by the source domain in the domain manifold. Besides, it selectively regularizes the data from different domains through a siamesed network with shared parameters to capture domain-invariant features. As a result, a robust underwater detector is constructed. Experimental results show that the proposed method can achieve the highest detection robustness on the synthetic underwater domain generalization dataset.

KEY WORDS: Underwater object detection, Probabilistic modeling, Hard example mining, Domain generalization, Data augmentation

# 目录

# 第一章　绪论

## 1.1　研究背景

### 1.1.1 水下机器人的发展现状

海洋占地球总面积的 71%，海洋拥有着异常丰富的生物资源和矿产资源。然而长期以来，由于技术上的限制，人类一直集中于开发陆地上的资源，海洋中丰富的资源仍未有进行大规模开采。我国海岸线曲折漫长，拥有着面积巨大的海域，海洋中丰富的资源对我国的经济、民生、军事建设具有非常大的意义。在我国，从“十三五”规划纲要首次提出拓展蓝色经济空间的总体要求，到党的十九大报告进一步提出“坚持陆海统筹，加快建设海洋强国”的战略部署，再到“十四五”规划和 2035 年远景目标纲要在“优化区域经济布局促进区域协调发展”篇中对拓展海洋经济发展空间作出部署，陆海统筹理念贯穿于建设海洋强国总体布局之中。可见我国战略上对海洋资源开发的重视程度。

“中国智能机器人之父”蒋新松院士自改革开放之后，出国考察，了解到国外自动化领域的一个重要方向便是机器人，提出将极限作业机器人作为主攻方向，即在危险、肮脏、人所不能达到的环境下进行作业的机器人。而极限机器人开发的第一步，便选择的是水下机器人。1985 年 12 月 12 日，沈阳自动化研究所研制的“海人一号”水下机器人在旅顺港首航成功。自此，我国在水下机器人领域有着长足的发展。

目前，根据工作方式的不同，水下机器人可分为了两种类别： 自主式水下机器人（Autonomous Underwater Vechicle，AUV）和遥控式水下机器人（Remote Operated Vechicle，ROV）。水下机器人技术的研发为人类探索、开发海洋提供了巨大的帮助，例如海洋生物检测、海产品捕捞、海底矿藏调查、海底遗物搜寻和探索以及海底估计测绘和重建等[1]。

为了完成复杂的工作，高效的水下感知能力是必不可少的。现阶段水下机器人对于水下环境的感知根据传感技术的不同分类了两大类：声学感知和视觉感知。声学感知主要利用声纳等设备，对水下声波信息进行收集，生成水下声纳图像进行分析，从而判断水下环境的状态。视觉感知主要利用了水下视觉相机进行光学图像的捕捉，对图像中的近距离环境进行分析。相比于声学感知，视觉感知具有成本低廉、反应敏捷、符合人类感知逻辑的优点[2]。因此，本文的主体工作围绕视觉感知的水下机器人，分析水下视觉图像中的难点，提出对应的解决方法。

### 1.1.2 水下目标检测的研究背景与意义

目标检测（Object Detection）是计算机视觉基本的识别任务之一，其目的是为了获得给定图片中特地物体的类别和位置，因此目标检测可以分成两个子任务：分类（Classification）和定位（Localization）。

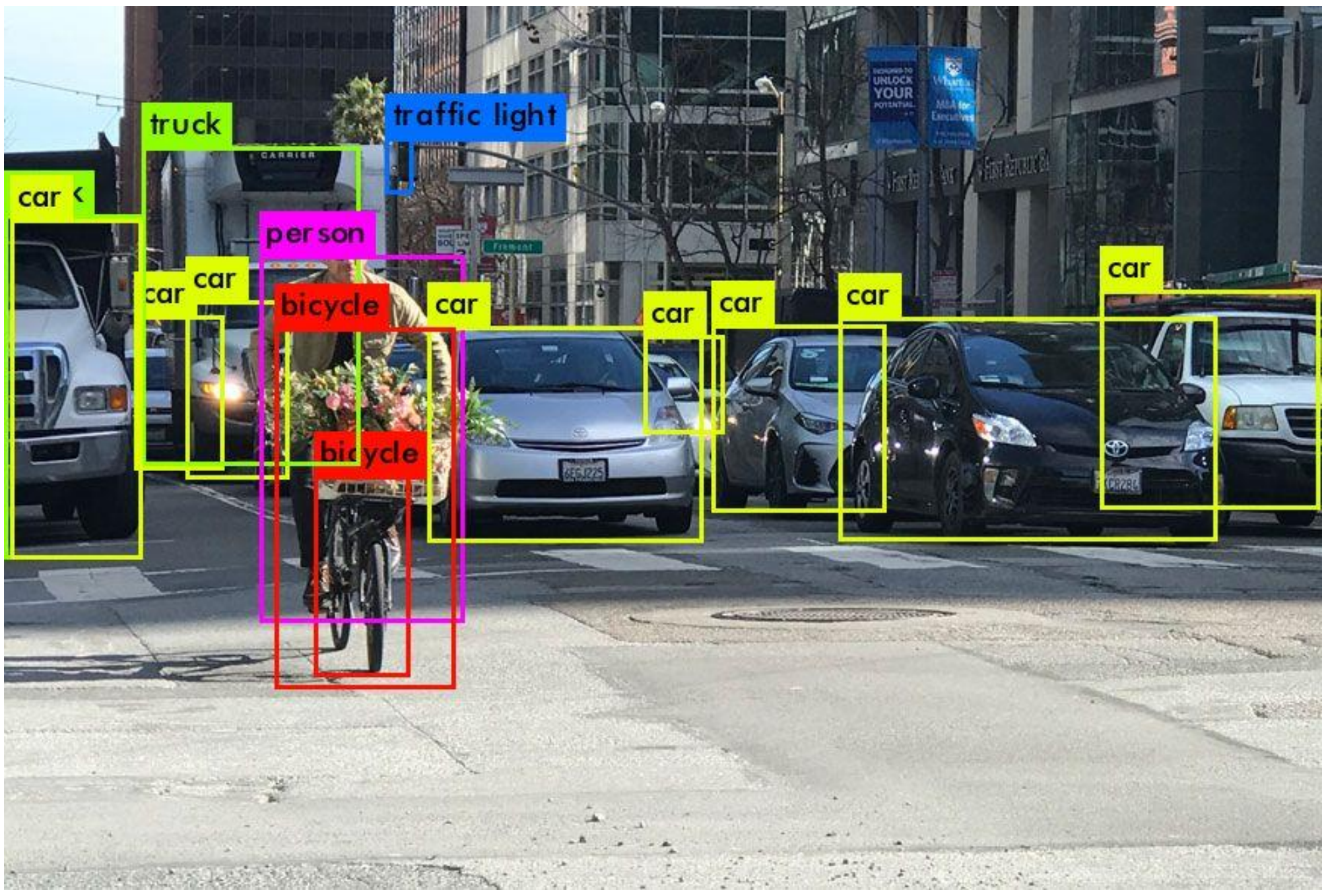

图 1.1 目标检测的识别结果
Figure 1.1 Results of Object Detection

由于目标检测技术的不断成熟，其已经被越来越广泛地应用到工业和日常生活中，例如：

（1）智能超市：近年来，随着计算机技术的不断进步和深度学习在各种视觉任务上实现的巨大性能突破，无人超市已经并非一个纯粹的概念，成为了一个可以切实落地的项目方向，成为了各大互联网巨头争相抢夺的战场。通过在超市内部增加各种传感器，结合最前沿的人工智能技术，赋能超市获得“智能”。其中，目标检测技术便占据了整个智能超市的核心。目标检测技术可以获取摄像头中顾客的边界框，快速定位顾客位置，同时边界框的顾客图像可以进一步执行下游任务：顾客身份识别、顾客行为识别、顾客姿态估计等，实现更加细粒和智能的顾客监控。同时，目标检测技术可以辅助进行货架上的货品管理，当顾客拿走商品，目标检测能辅助系统把该商品和该顾客联系起来，将账单加入到该顾客中，实现客货关联。

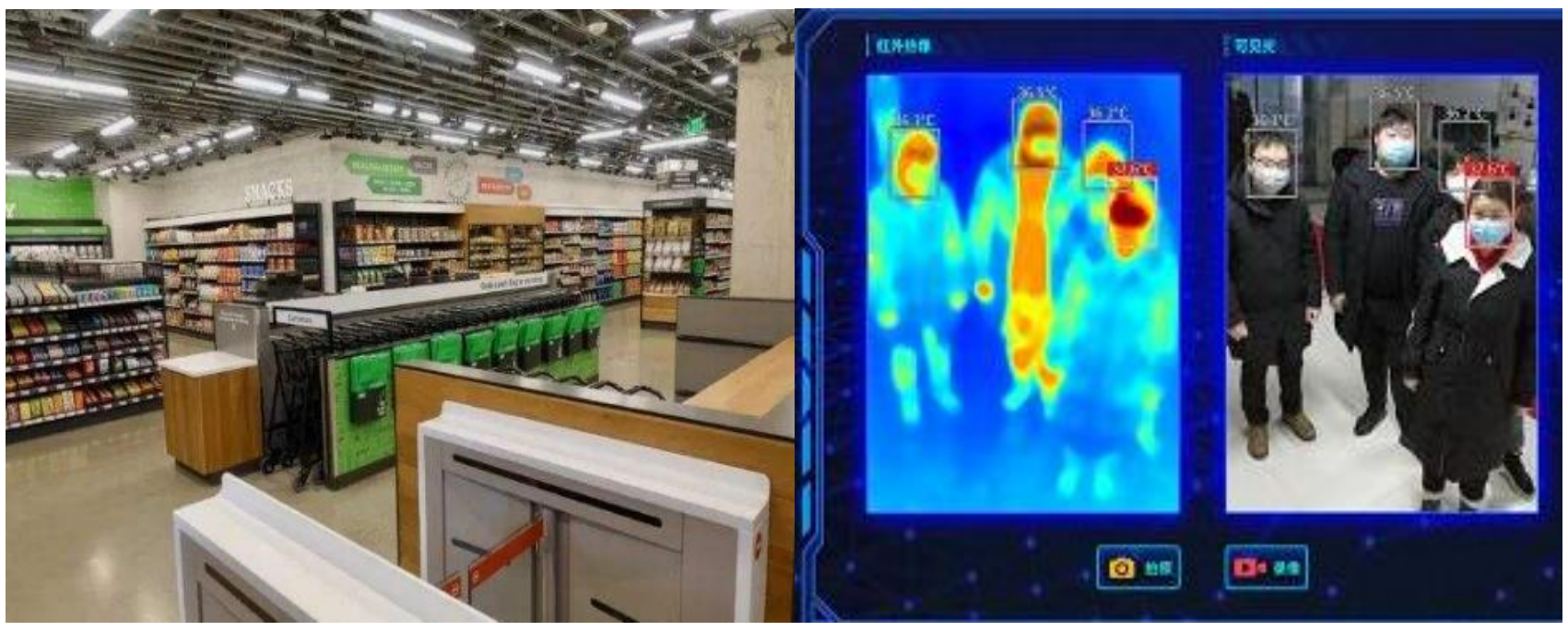

（a）智能超市 （b）体温监测

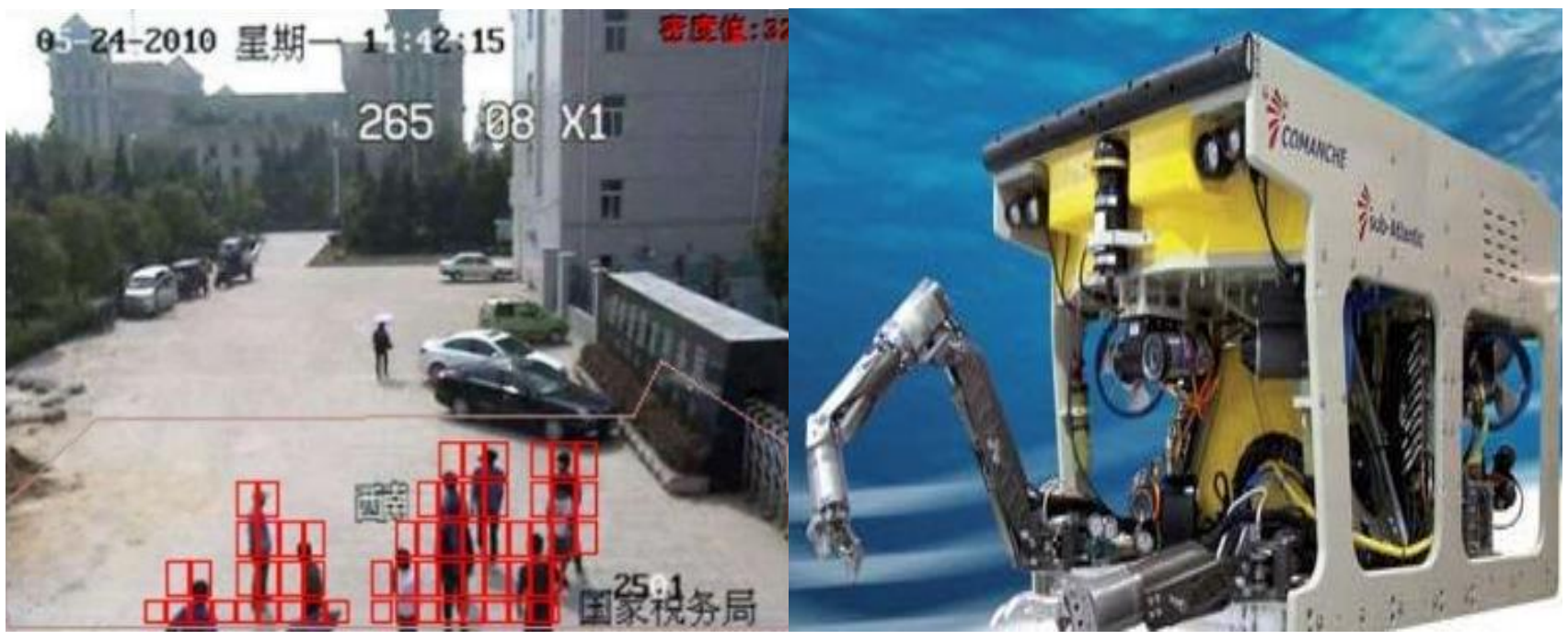

（c）智能监控 （d）水下机器人

图 1.2 目标检测的应用领域示例

Figure 1.2 The Application of Object Detection

(a) Smart shop; (b) Temperature detection; (c) Smart surveillance; (d) Underwater robot

（2）疫情下的体温监测：在新冠疫情期间，为了在做好疫情防护的同时，不影响人们的正常生活，需要在大型的商业中心和人群密集的服务场所（医院、学校、体育馆等）进行体温监测。但人流过大的时候，完全采用人工进行检测，成本大且效率低，因此基于目标检测的体温监测便能够大大提升效率，降低成本。目标检测能够迅速定位摄像头中人的位置，并将人的边界框截取此人的图像进行视觉体温检测，整个过程能够实时进行。若体温不正常者进入商场，系统能够迅速感知到并且做出报警。

（3）智能监控：伴随着人们的安全意识的提高以及监控设备成本的降低，越来越多的场所都配备了监控摄像头，如医院、学校、商场等。监控信息可以给公安部门提供线索，抓捕不法分子。但是监控设备的增加也带来的巨大工作量，而目标检测技术就可以帮助公安部门得到监控下有人的时间段，迅速定位人的位置，结合人体目标重识别技术和人脸识别技术，更可以快速从大量的监控素材中找到犯罪分子的行踪，大大降低执法成本，降低犯罪率。

此外，目标检测也可以用于水下机器人，实现水下目标检测。近年来，越来越多的研究团队考虑在水下机器人上搭载视觉系统，以进行更加复杂更加智能的水下工作。例如在海产品捕捞上，水下机器人可以通过摄像头得到需要捕捞的海胆、海参、扇贝等

生物的位置，操控特殊的机械臂或捕捞器对海产品进行捕捞，能够大大降低人力成本，提升捕捞效率。随着对于水下机器人研究的深入，研究者开始将目标检测技术和搭载着视觉系统的水下机器人进行结合，赋予水下机器人智能，使其能够自动化完成海产品捕捞、鱼类养殖、生物监控等任务[3,4]。然而水下的场景的多样性和复杂性，给目标检测技术提出了新的挑战。第一，不同水质下的水下环境非常不同，不同江、河、海拥有着不同的水质成分，导致水对不同波长光有不同的衰减率，其水质的清晰度和颜色均不一样，图 1.3（a）展现了 4 种不同的水质环境。第二，水下环境的光照条件也是多变的，水下的光源主要来自于水面的太阳光，一天不同时刻的光照条件及其不同，而水下的各种地形环境（礁石等）会导致视觉系统采集到的图像呈现出不均匀光照的现象，如图 1.3（b）；第三，水下场景下需要检测的目标的尺度多样，且因为生物聚居的原因，常常会产生遮挡现象，如图 1.3（c）；第四，水下生物为了躲避捕食者的袭击，往往存在着生物拟态效应，即水下生物偏向于在颜色和形态上接近所处环境，如图 1.3（d），生活在海底的扇贝和周围的石头长相非常接近。第五，水下机器人往往处于运动状态，而在水下的光照不充足的情况下，相机在高速运动之下没有办法充分感知光线，因此所捕获的图像会出现因运动而导致的模糊。因此，复杂多变的水下场景下的目标检测技术有着不可替代的学术意义与应用意义。

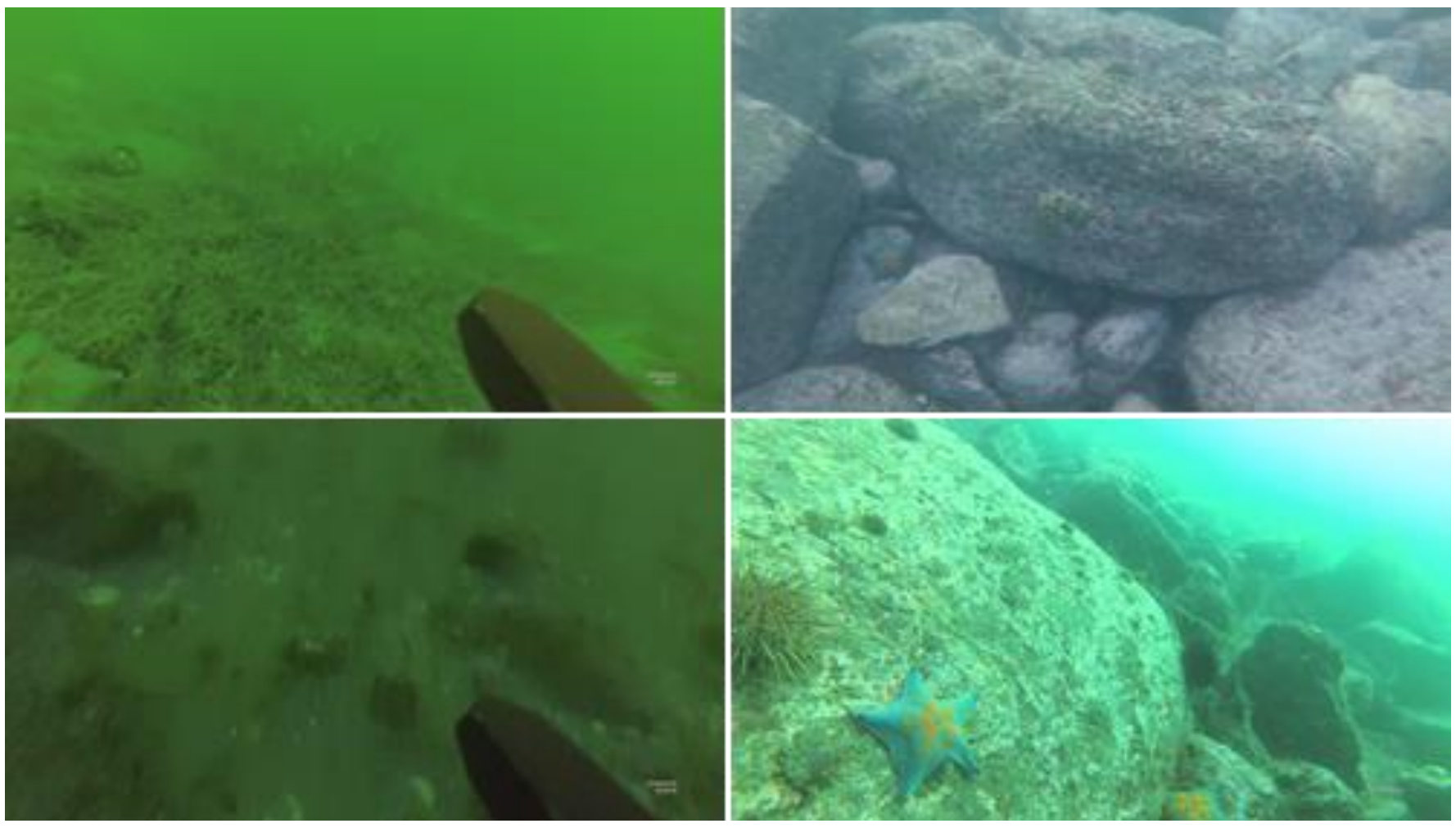

（a）水质变化

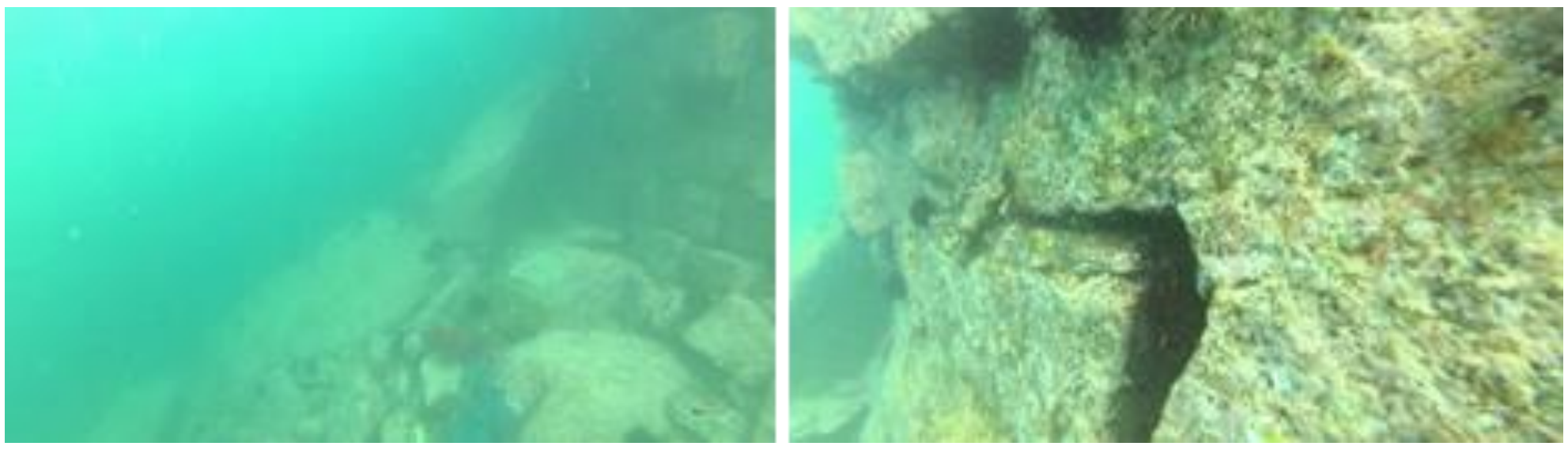

（b）不均匀光照条件

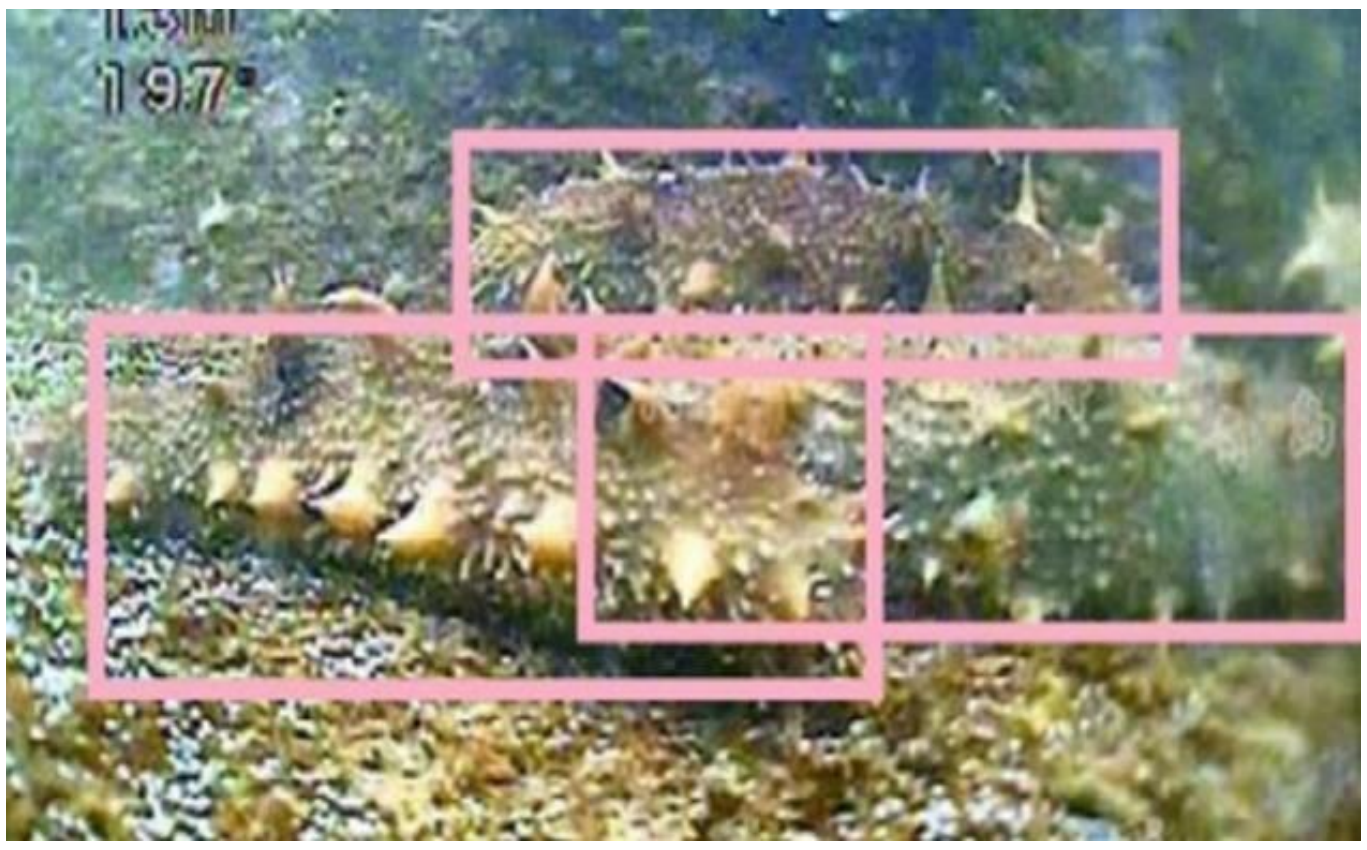

（c）遮挡问题

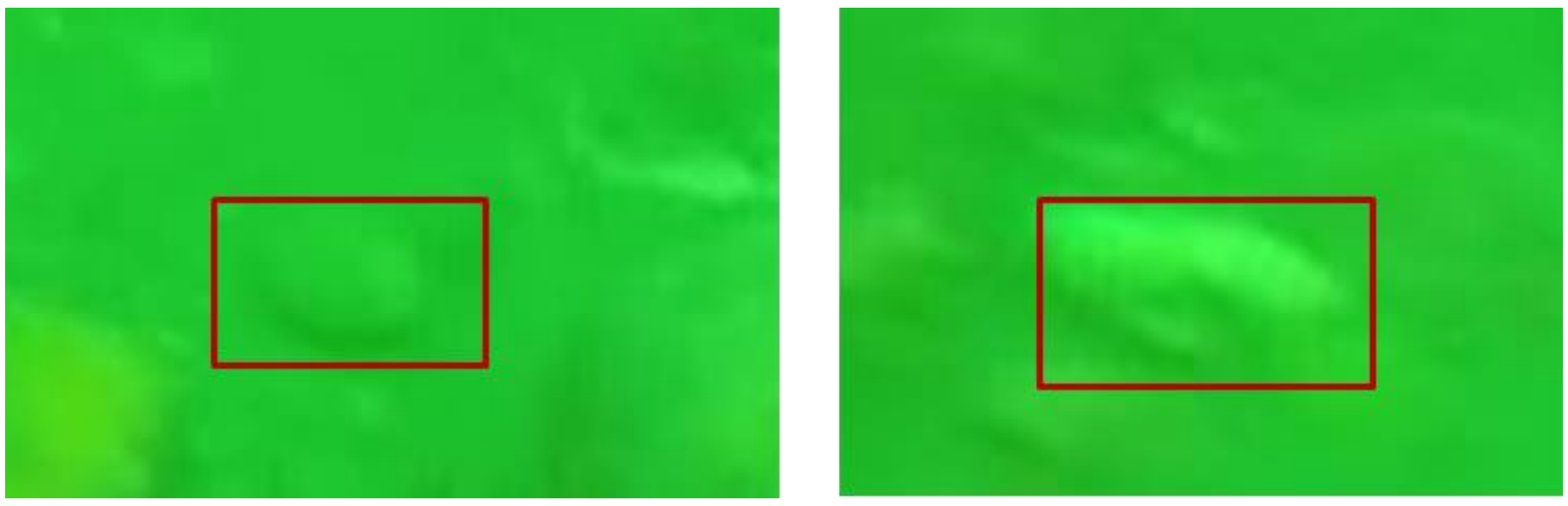

（d）生物拟态，左侧为扇贝，右侧为石头

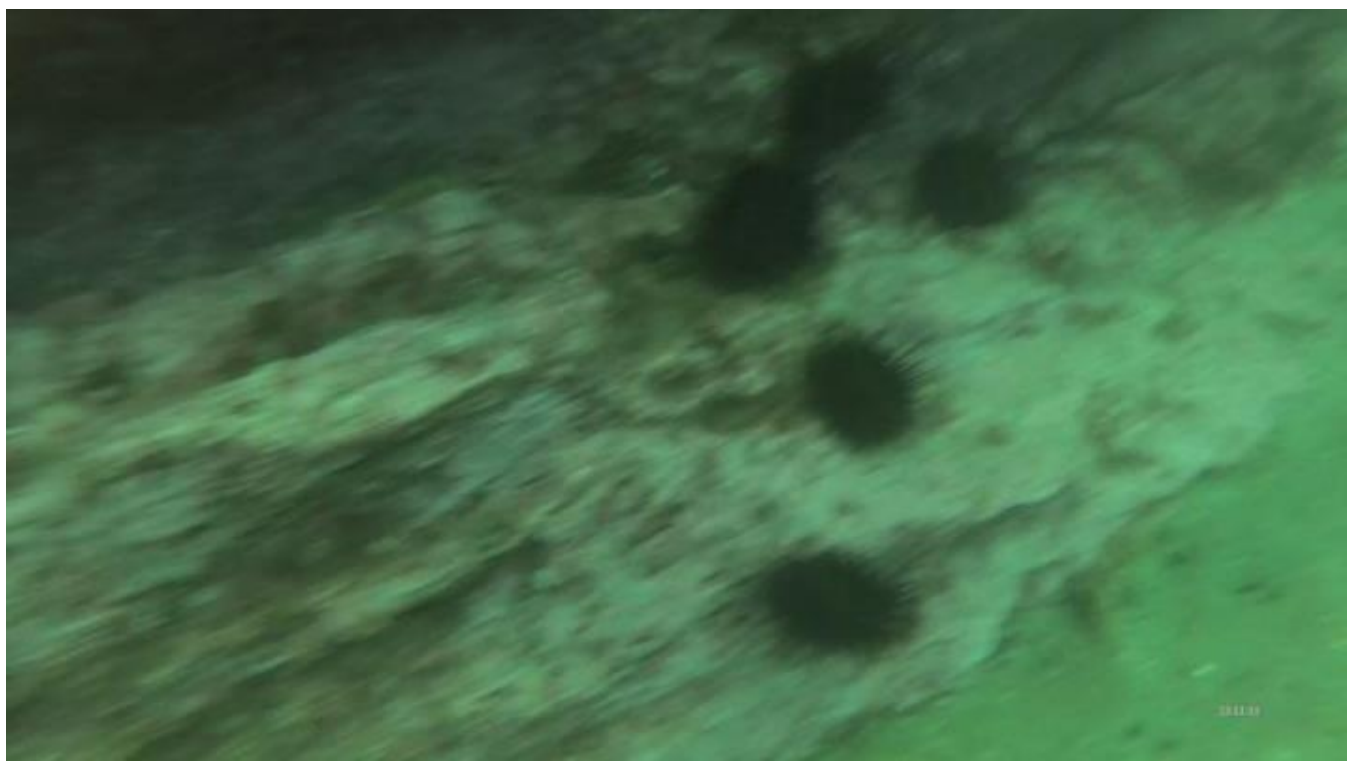

（e）运动模糊

图 1.3　水下场景的难点问题

Figure 1.3 The challenges in underwater scene

(a) Various water quality; (b) Unbalanced light condition; (c) Occlusion;

(d) Biological mimicry; (e) Motion blur

### 1.1.4　水下域泛化技术的研究背景与意义

机器学习和深度学习在各个领域之中都取得了卓越的成就，深刻地改变了工业生产和人们的日常生活。机器学习的目标是建立一个模型，并希望在给定的训练数据中提取通用的、总结性的知识，并可以将这些知识应用在新的数据下。在传统的机器学习下会遵守一个基本假设，即训练数据和测试场景的数据均是独立同分布进行采样的。然而，出于采样条件的限制，这个基本假设在实际应用上并不一定满足。当训练数据和测

试数据的分布有差异时，模型的性能就会产生极大的下降。训练数据与测试数据上共同拥有的信息被称作语义（Semantic Content），而训练数据与测试数据上不同的信息被称为域（Domain），而训练数据和测试数据分布上的差异被称作域间隙（Domain Gap）。如图 1.4，真实场景下的猫和动画中的猫虽然都有着“猫”这一共同的语义，却有着极大的差异，而不同表现风格即为域信息。

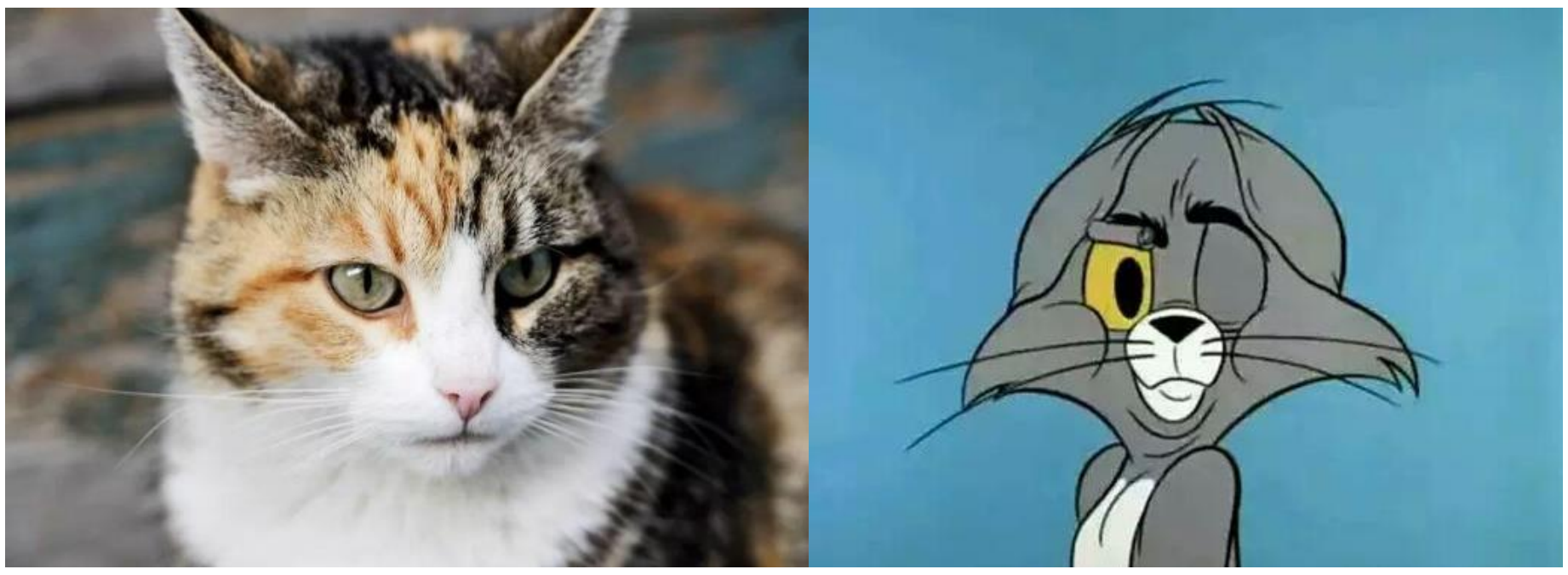

（a）真实场景下的猫　　（b）动画中的猫

图 1.4 域间隙在“猫”样本上的示例

Figure 1.4 Illustration of domain gap of cat

在机器学习的实际运用中，因为采样到所有的域是不可能的一件事，所以域间隙基本上是无法避免的。例如，采集到所有画风或者所有品种的猫是不可能的。但是人类是有能力去辨认不同风格的猫的，即使人类也从来未看过所有风格下的猫的图片。因此也希望对机器学习模型提出同样的要求：超越域去捕捉到语义信息，即域泛化（Domain Generalization）的能力。域泛化的定义为从一个或多个不同但相关的域中去训练模型，使得模型能在从未见过的测试域之下也能获得很好的性能（即泛化到未知域）。如图 1.5，给三种不同域的数据（草图、卡通、艺术画），希望模型能够在这三种域的数据中训练，却可以在真实场景（图片）的域上以低错误率进行准确的识别。

与之类似，域自适应（Domain Adaptation，DA）也意在解决训练域和测试域不一致的场景。通常为无监督域自适应（Unsupervised Domain Adaptation，UDA），其定义为：模型在给定有标签的源域数据和无标签的指定目标域数据上进行训练，能够在指定目标域上获得很好的性能（即泛化到目标域）。与域泛化不同，在无监督域自适应范式中，目标域的数据是可得的（无标签），而最终的性能测试标准也是目标域上的性能。因为其与域泛化的相似性，两主题下的工作也可以相互借鉴相互研究。

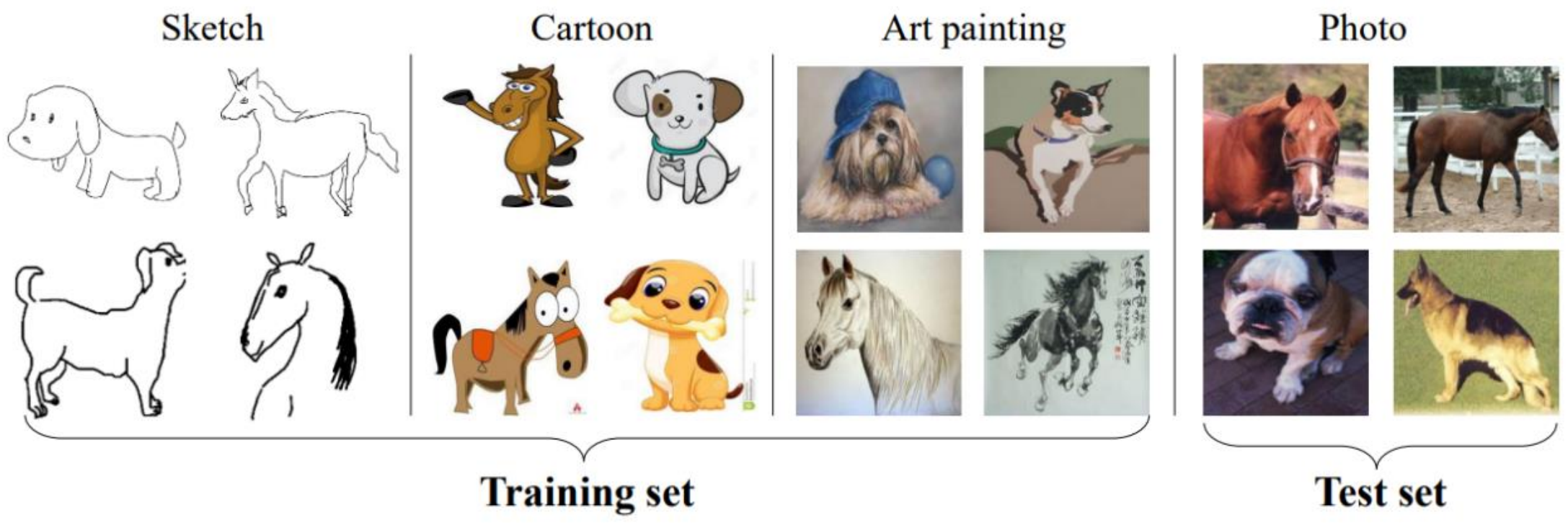

图 1.5 域泛化数据集 PACS 的示例
Figure 1.5 Illustration of domain generalization dataset PACS

一个完备的人工智能系统应该要有足够好的泛化性能，能够处理不同复杂场景的任务。因此对于人工智能系统，尤其是在视觉任务上，域间隙是无法避免的，高泛化性能在实际应用上是必须的。于是，这些年来域泛化问题越来越受到研究者的关注，对于域泛化的研究层出不穷。图 1.6 展示的是自 2011 年至今，Domain Generalization 这一关键词的谷歌趋势指数，可以发现在 2019 年前后这一词条的热度有明显地上涨。图 1.7 展示的是域泛化主题的论文在视觉顶会（CCF-A 会议）和顶刊（TPAMI、TIP、PR、TMM）中的论文数量，可以很明显地发现自 2019 年之后，域泛化的论文数量有着井喷式的爆发。

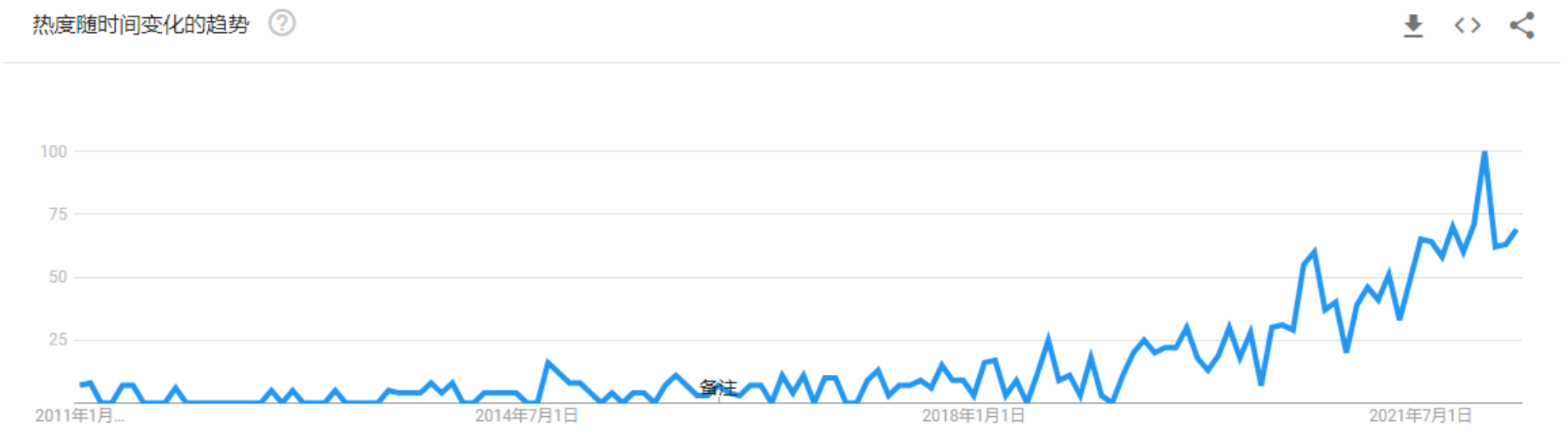

图 1.6 域泛化（Domain Generalization）2011 年至今的谷歌趋势统计
Figure 1.6 Google trend of domain generalization from 2011

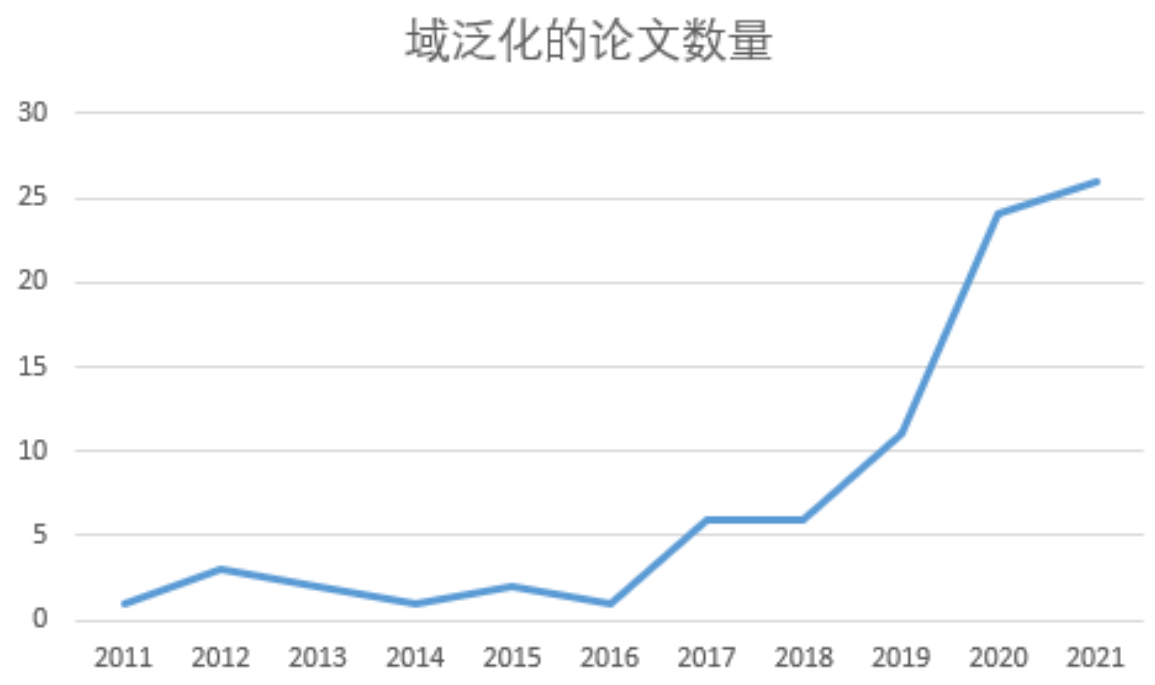

图 1.7 2011 年至 2021 年域泛化主题在视觉顶会和顶刊中的论文数量统计
Figure 1.7 The statistic of domain generalization papers in top tier conference

域泛化的早期大部分的工作主要是集中在视觉识别任务（Recognition）上。近年来，各个视觉领域对于域泛化的关注也越来越高，例如语义分割（Semantic Segmentation）、行为识别（Action Recognition）、人脸防欺诈（Face Anti-spoofing）、行人重识别（Personal Re-ID）和视频理解（Video Understanding）。

水下环境的多样性和复杂性，使得在封闭的、单一域的水下数据集训练的目标检测模型面对极端的、多样的光照条件的水质环境下必然会产生域间隙，导致模型的性能大大下降。早期的研究者认为水下目标检测可以借助图像还原（Image Restoration）和图像增强（Image Enhancement）的工作对水下图像进行复原，还原出图像在没有水的样子，再借助强大的通用目标检测器进行检测。早期进行水下图像复原和水下图像增强的工作均有提到他们的工作对于水下场景的目标检测有着性能的提升，但是并未做大规模的对比实验。2020 年，中科院自动化所的 Xingyu Chen 等人[5]详细研究了水下图像复原对于目标检测的性能增益的帮助。实验结论显示，（1）水下图像复原技术对于模型的训练和测试上均无明显增益；（2）在有损图像上训练模型对于模型的鲁棒性有较强的增益。因此，对于水下目标检测，本文认为传统意义上的先做图像复原再进行检测的思路并非一个合理的思路，更重要的是要从根本上提高模型本身的鲁棒性。

水下场景的多样性和复杂性（光照、水质等），恰好符合域泛化问题的研究主题：（1）水下数据难以采集，给水下生物进行标注往往需要更专业的知识，因此囊括多种水质场景的大规模水下数据集在工程上难以实现；（2）不同水质下的水下环境非常不同，如江、河、海拥有着不同的水质成分，导致水对不同波长光有不同的衰减率，其水质的清晰度和颜色均不一样；（3）一天不同时刻的光照条件极其不同，而水下的各种地形环境（礁石等）会导致视觉系统采集到的图像呈现出不均匀光照的现象。而在工业应用中，例如水下机器人的海产捕捞，人们希望设计的算法能够不加调整或者尽可能减少调整立即投入使用。若模型的鲁棒性不能达到要求，机器人的产品化便难以实现。因此，水下场景的域泛化问题具有非常高的研究价值。

## 1.2 研究现状

### 1.2.1 基于深度学习的目标检测技术的研究现状

目标检测是计算机视觉的基本任务之一，在如今仍然是一项极具挑战性的研究工作。自 AlexNet 在 ImageNet 任务上取得了卓越的性能之后[6]，目标检测领域焕发了新的生机，获得了长足的发展。现今的目标检测模型倾向于利用一个强大的骨干网络提取图像特征，并在特征层面上对图像中的潜在目标进行检测。由于其任务的复杂性，使得各个研究团队能够在各种切入点上进行，大大推动了该领域的发展。现如今基于深度学习的目标检测技术占据了绝对的主流，取得了传统检测方法无法达到的性能，因此本

文的研究现状只关注深度学习系列的目标检测算法，对过往的传统目标检测算法不做介绍。

（1）二阶段目标检测算法

目标检测包括两个子任务：识别和定位。在目标检测之前，识别任务已经有了充分的研究，若想要实现从识别到检测的飞跃，最简单的思路是将检测降级为识别任务：预设一群候选框（Proposal），对每一个候选框进行识别。2014 年，R. Girshich 等人提出的 R-CNN 便是采用此思路[7]。如图 1.8（a），该工作提出选择性搜索（Selective Search）这一技术，用于得到原图像中的感兴趣区域（Region of Interest，RoI）。这些 RoI 转换成固定大小的图像，并分别送到 CNN 之中提取特征（原论文采用了 AlexNet）。对 CNN 提取的特征利用 SVM 进行分类和利用线性回归进行边界框校正，最终得到检测结果。2015 年，R. Girshich 等人继续提出了 Fast R-CNN[8]。如图 1.8（b），该工作直接在骨干网络中的特征图上截取 RoI，使用 RoI Pooling 将不同大小的 RoI 缩放成同一大小再进行分类和识别，相比于 R-CNN，Fast R-CNN 能达到更高的检测速度和更高的检测精度。2015 年，S. Ren 等人提出了 Faster R-CNN[9]。如图 1.8（c），该工作提出了区域提议网络（Region Proposal Network，RPN）代替选择性搜索，大大提高了检测效率和检测性能。至此 Faster R-CNN 奠定了二阶段检测器的范式：第一阶段，先采用 RPN 提出一系列候选框，在选取高质量的候选框并在特征图上截取下来作为 RoI；第二阶段，进行进一步的回归和识别。

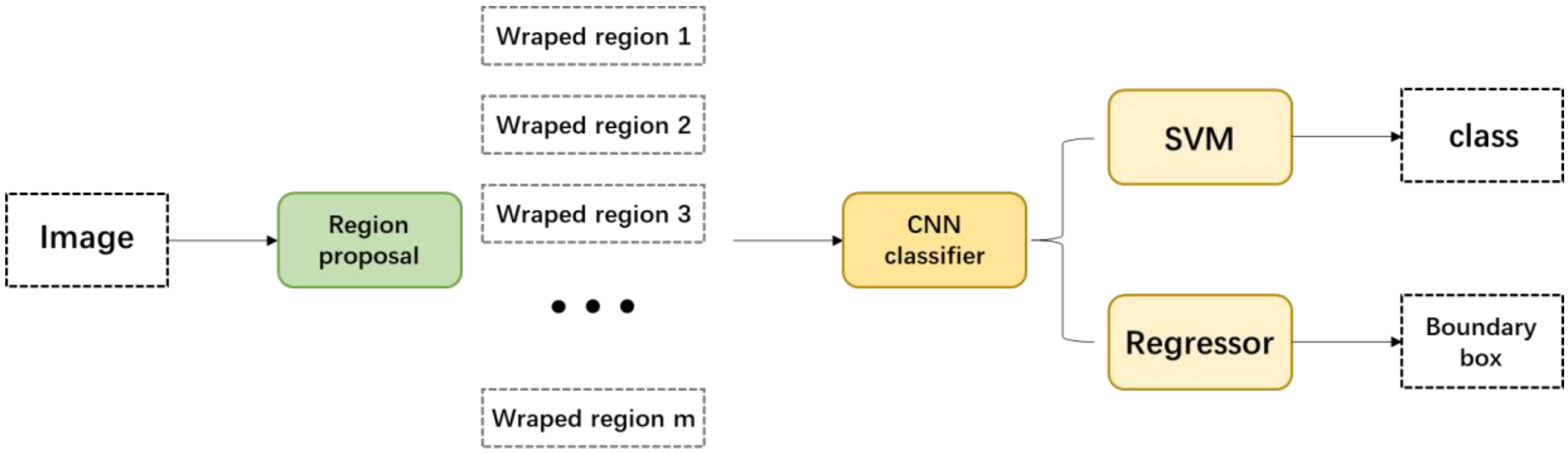

（a）R-CNN 的流程图

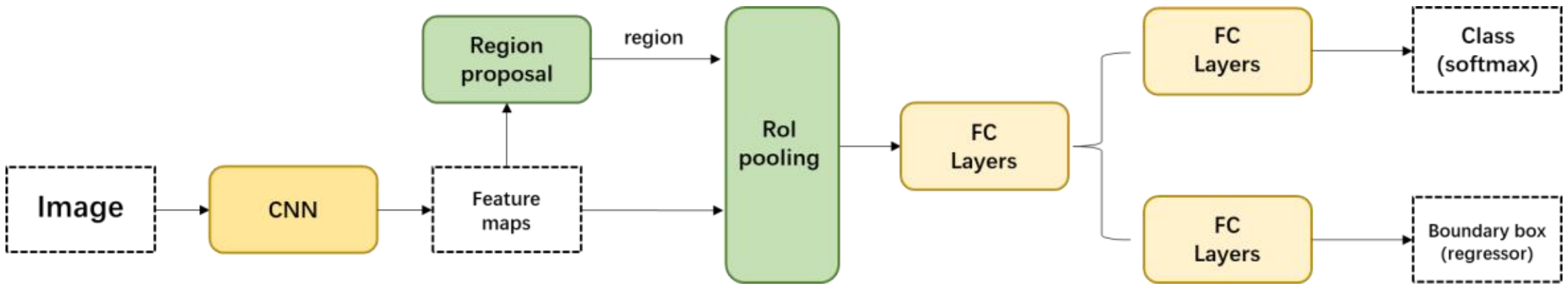

（b）Fast R-CNN 的流程图

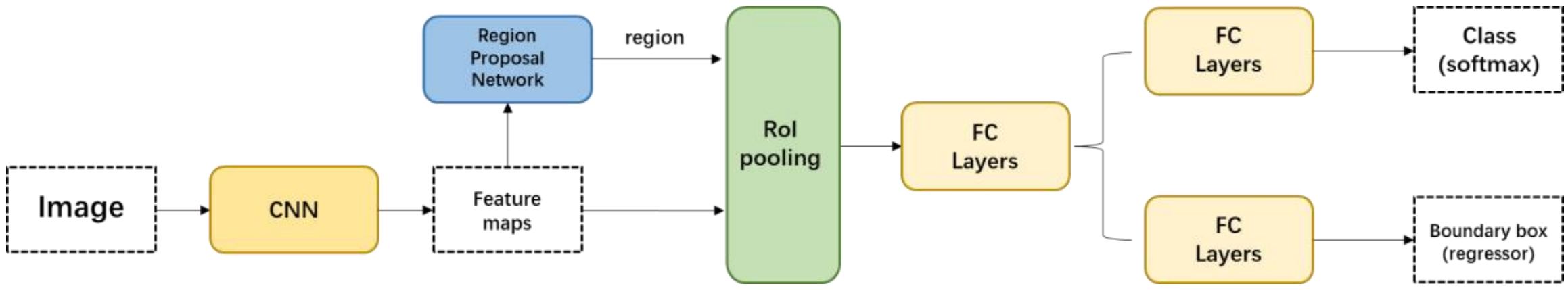

（c）Faster R-CNN 的流程图

图 1.8 二阶段目标检测器的流程图

Figure 1.8 Flow chart of two-stage detectors

（2）一阶段目标检测算法

2015 年，R. Joseph 等人提出了 YOLO（You Only Look Once）[10]，这是第一个真正意义上的一阶段检测算法。相比于二阶段检测器 Faster R-CNN，YOLO 是一个全卷积网络，不需要 RPN 提取候选框，也不需要 RoI Pooling 进行候选框的缩放，直接得到目标的坐标和识别结果。因此，YOLO 大幅度提升了检测速度，这也是一阶段检测器的普遍特征。2015 年，W. Liu 等人提出了另一个一阶段检测器 SSD（Single Shot MultiBox Detector）[11]，区别于之前的检测算法，其利用 VGG[12]的多层特征同时进行检测，提升模型在多种尺度目标上的检测性能。2017 年，T.-Y. Lin 等人研究认为，一阶段与二阶段性能上的差异来自于其正负样本的极度不平衡，因此该工作提出了一个新的损失函数：Focal Loss[13]，并以此设计出一阶段检测器 RetinaNet，该检测器能够达到和二阶段检测器不相上下的性能。

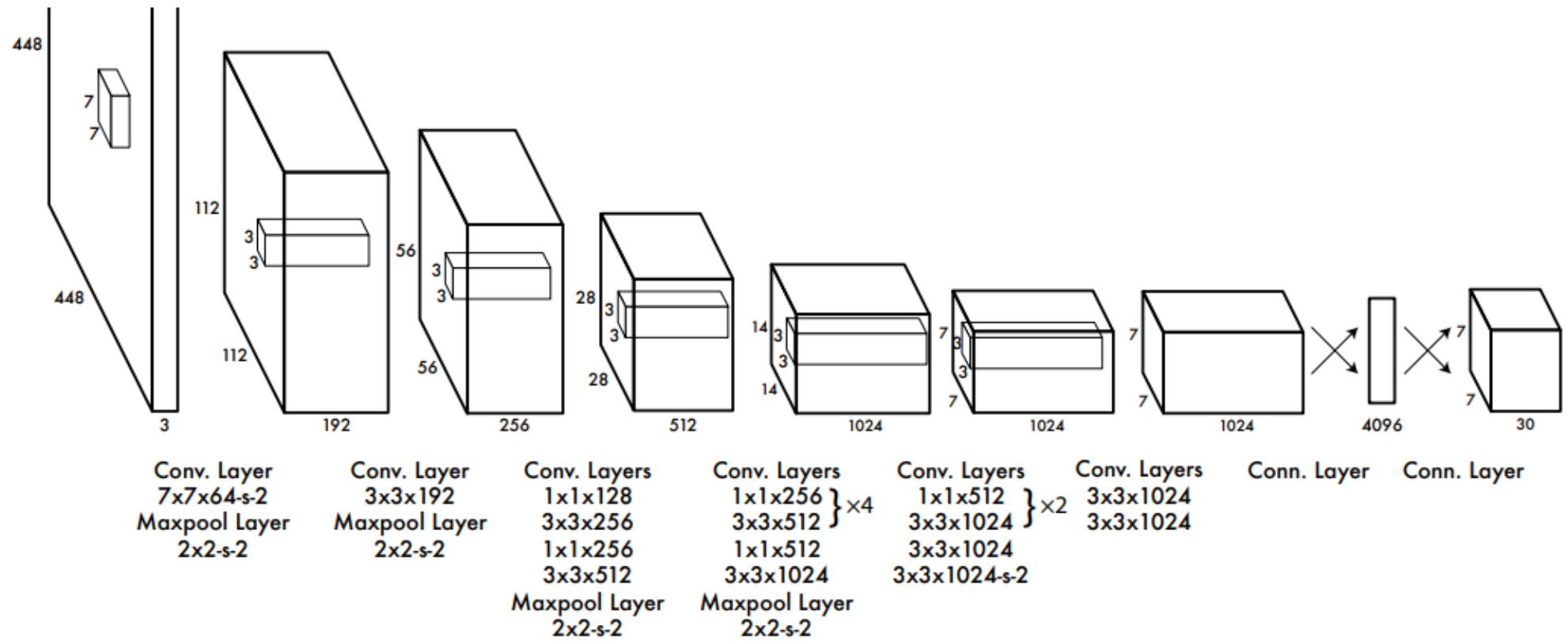

（a）YOLO[10]

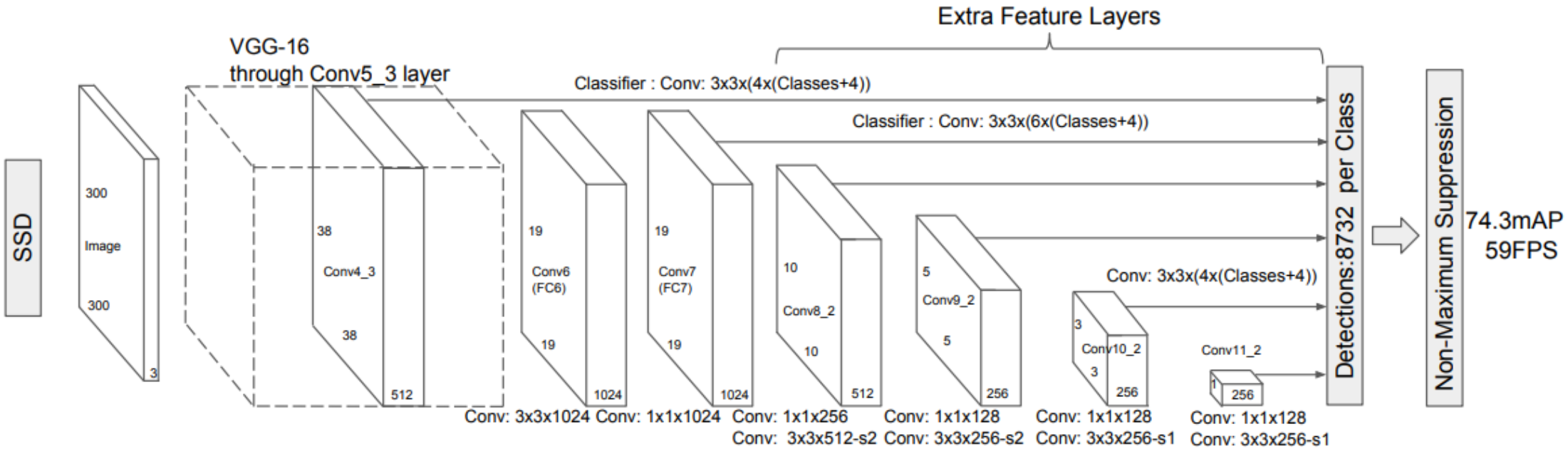

（b）SSD[11]

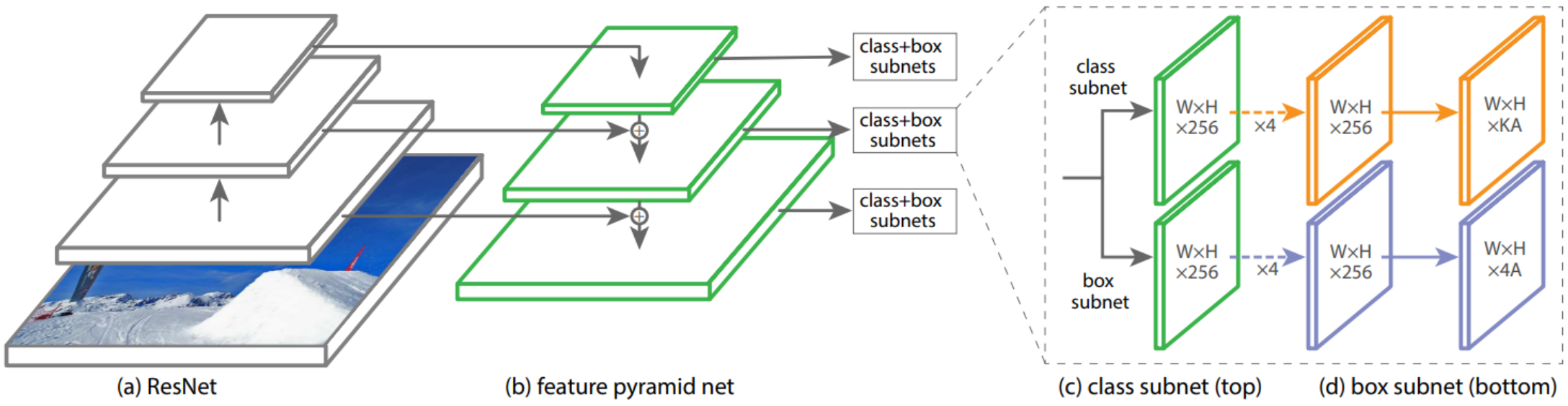

（c）RetinaNet[13]

图 1.9 一阶段目标检测器的流程图

Figure 1.9 Flow chart of one-stage detectors

（3）多尺度的特征融合

早期的目标检测器利用的特征均为骨干网络的最后一层。虽然最后一层特征有利于目标的识别，但其不利于目标的回归任务。2017 年，T.-Y. Lin 等人[14]设计了一个自顶向下的金字塔结构，把骨干网络中的浅层信息和深层信息进行进一步结合，提供给检测网络更丰富的特征，从而提升检测任务的性能。2018 年，S. Liu 提出了 PANet[15]，该工作是基于 FPN 的改进，将 FPN 得到的特征继续上采样融合得到新的特征金字塔，通过自适应池化融合多层金字塔的信息进行处理。2019 年，K. Sun 等人提出了 HRNet[16]，该网络使用四条支流并联的方式，保有多种分辨率的信息，并且使用重复的多尺度融合以提高高分辨率表征，增强图像的识别和检测能力。2019 年，G. Ghaisi 等人提出运用网络架构搜索（Network Architecture Search， NAS）[17]去搜索出最合适的特征金字塔结构，得到最优的特征金字塔结构 NAS-FPN。2020 年，M. Tan 等人提出了 EfficientDet[18]，该工作中提出的 BiFPN 将 PANet 进行简化，加入了更多的跳层连接以实现更多尺度的融合，大大简化了参数量，同时提高了性能。

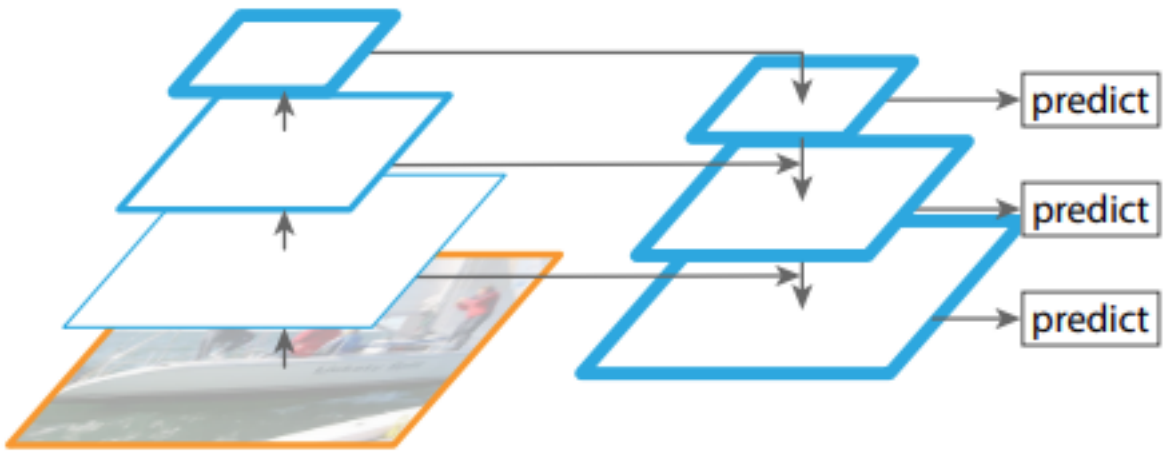

（a）FPN[14]

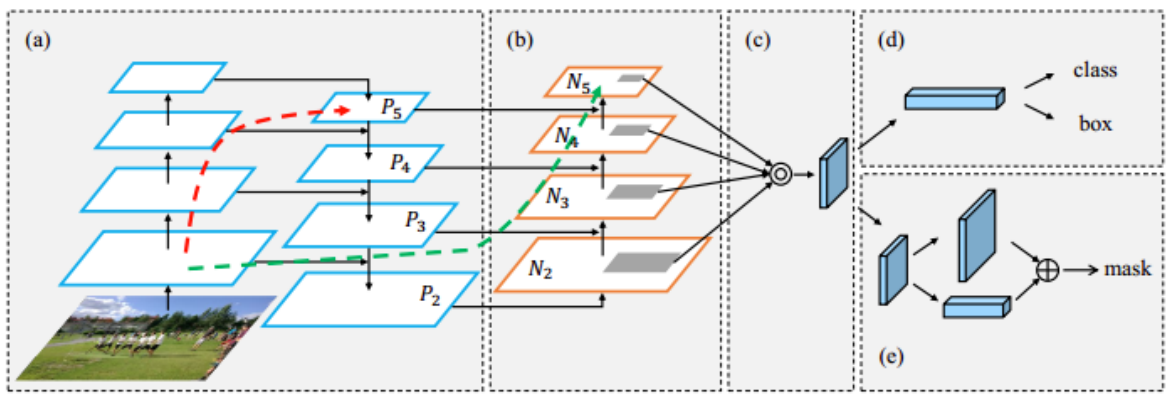

（b）PANet[15]

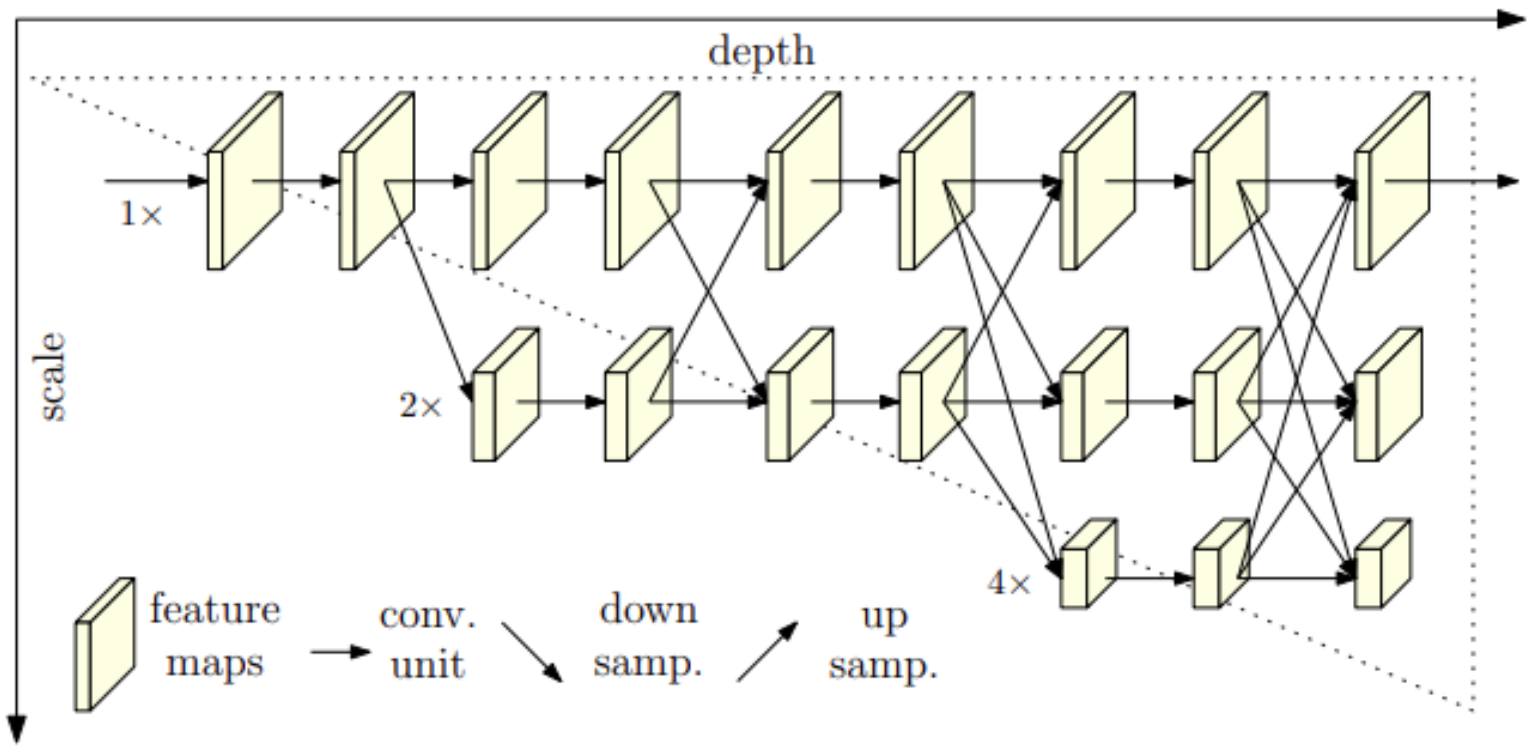

（c）HRNet[16]

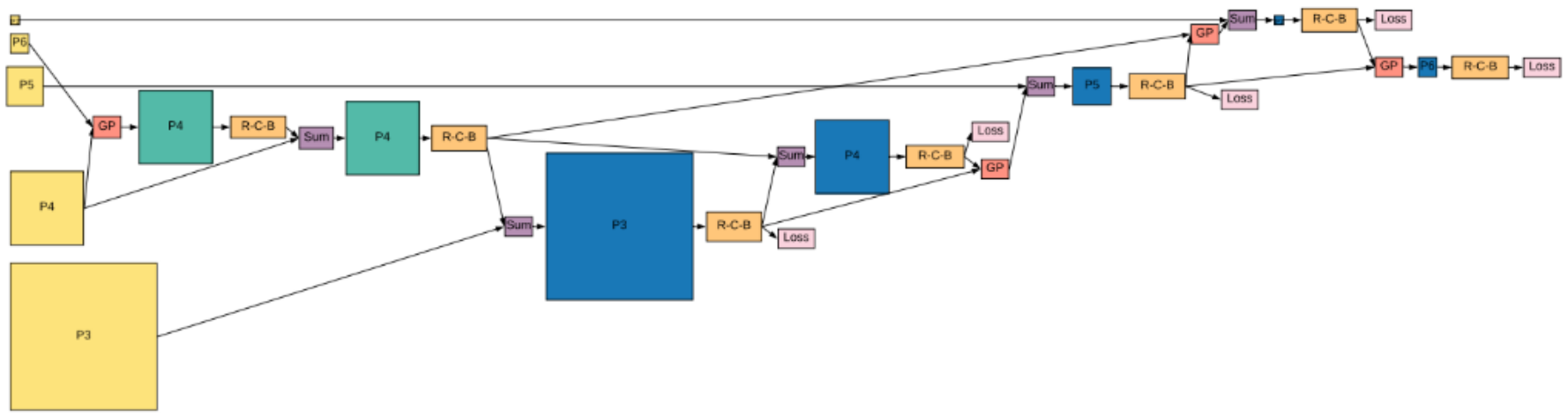

（d）NAS-FPN[17]

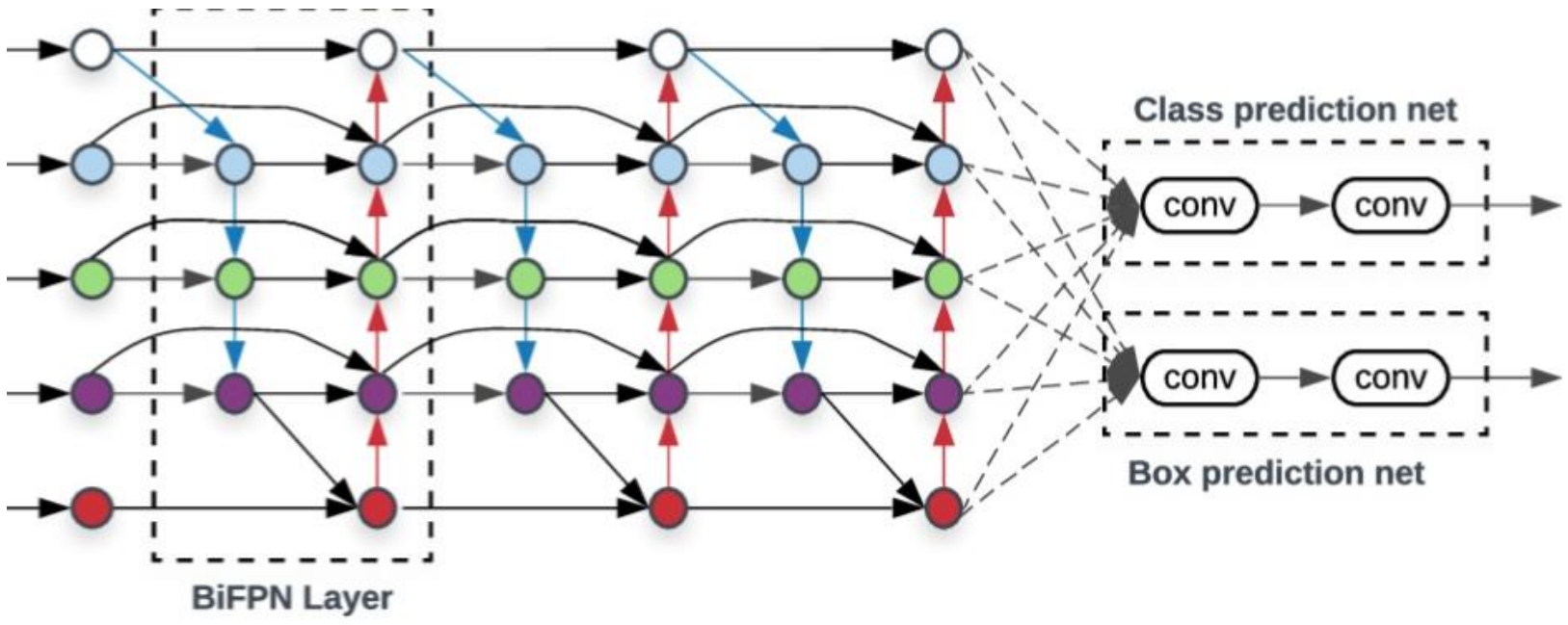

（e）BiFPN[18]

图 1.10  多尺度的特征融合

Figure 1.10 Multi-scale feature fusion

（4）Anchor-free 系列算法

Anchor 的原型为候选框，在 Faster R-CNN[9]的 RPN 模块中正式引入这一概念，作为进行检测前预设的一个默认检测框。在 RPN 中（如图 1.11）特征图上的每一个像素点对应三种不同形状的 anchor。在进行回归时，模型输出的是 anchor 和真实框的相对差值，早期研究者认为，相比起直接回归绝对空间坐标，基于 anchor 的检测能让回归任务更容易收敛。而随着对于 anchor 的研究的深入，研究者开始质疑 anchor 的必要性。YOLOv1[10]提出时采用的是 anchor-free 的设计，其回归输出为[x,y,w,h]，[x,y]为输出框到物体真实框的中心位置坐标的偏移值，并被归一化到[0,1]，[w,h]是物体的宽度和高度。在 YOLO 之后的系列中[19, 20]又引进了 anchor 这一概念。2018 年，H. Law 提出了 CornerNet[21]，将目标检测看作关键点检测，检测出一系列的左上角点和右下角点进行配对，舍弃了 anchor 的设计。2019 年，X. Zhou 等人提出了 CenterNet[22]，延续了 CornerNet

关键点检测的思路，其检测物体中心，回归中心偏移和物体长宽，本质上是将像素点视作 anchor；Z. Tian 等人提出 FCOS[23]，设计出了一种新型的无 anchor 检测器（如图 1.12），进行像素点中心到真实框边界值偏移的回归；Z. Yang 等人提出了 RepPoints[24]，基于像素点中心预测出 9 个点去捕捉物体的关键点，而这 9 个点可以组成一个预测框。

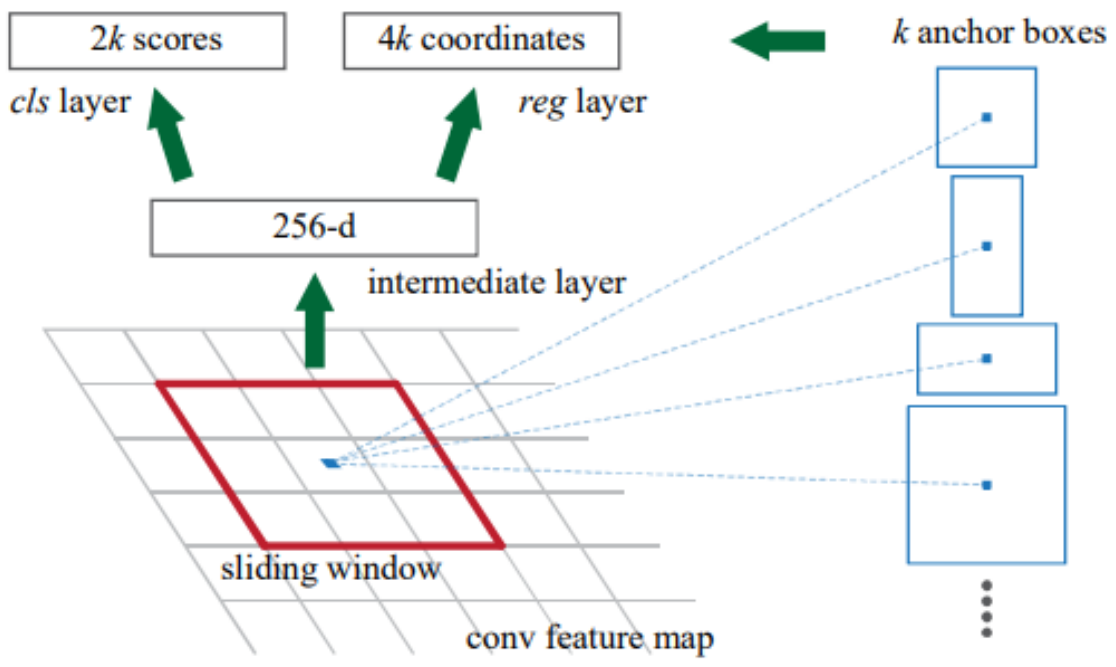

图 1.11 区域提议网络的 anchor 设计[9]
Figure 1.11 Anchor design of Region Proposal Network

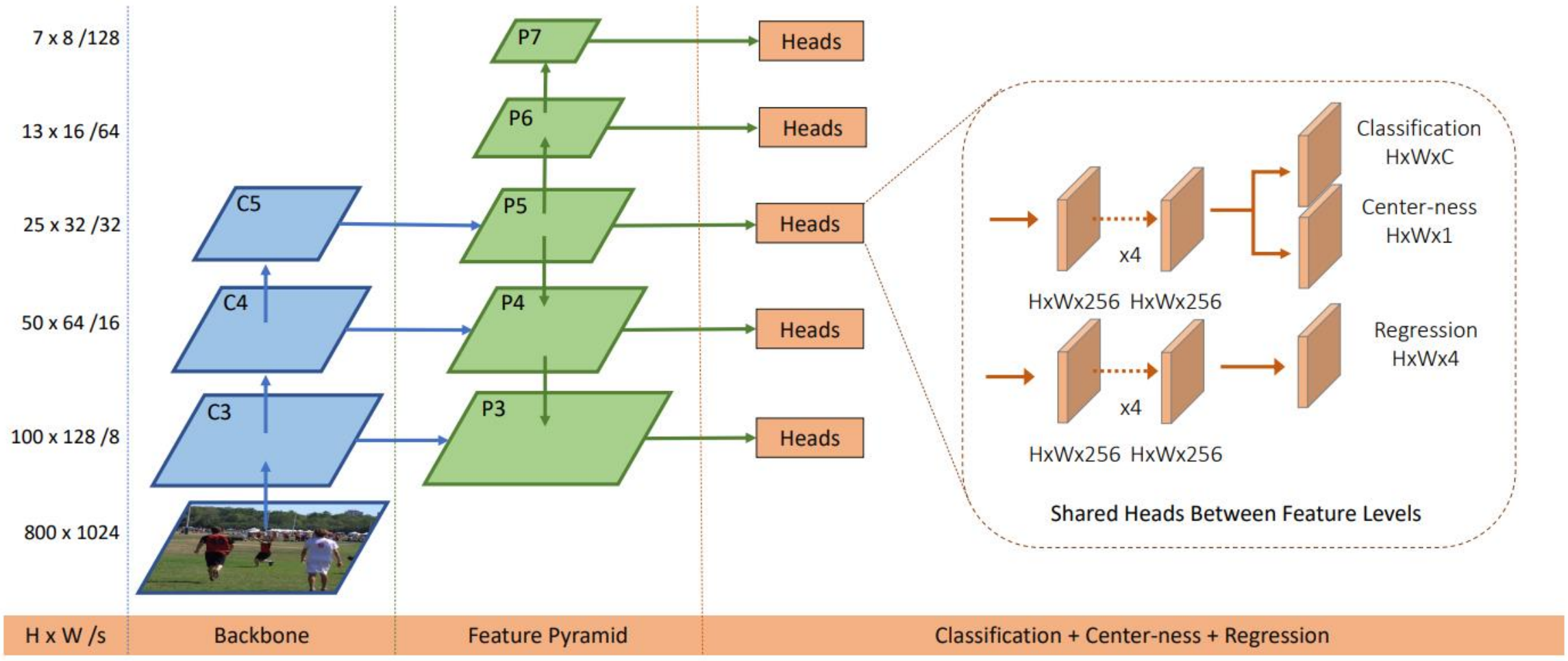

图 1.12 Anchor-free 检测器 FCOS 的模型结构[23]
Figure 1.12 The pipeline of anchor-free detector FCOS

（5）正负样本分配

当训练一个目标检测器时，首先需要定义正样本和负样本，然后用正样本去做回归。在 Faster R-CNN 中，一般规定预测框和真实框交并比（Intersection over Union，IoU）超过 0.5 以上的为正样本，低于 0.5 的为负样本。这种硬性的正负样本选择的方法并非是最优的，在模型训练前期，大多数的预测框的 IoU 都没办法超越阈值，导致模型前期训练困难；而在模型训练后期，大多数的预测框 IoU 都远远超越阈值，导致最后的预测结果引入了大量质量低下的预测框。2018 年，Z. Cai 等人[25]研究发现简单提升 IoU 阈值并不能直接提升检测性能，因此采用级联的方式逐步得到高质量检测框。2019 年，

X. Zhang 等人提出 FreeAnchor[26]，将目标检测器的分类和回归过程用最大似然估计进行训练，使 anchor 的分配往实现最大似然估计方向的目标去接近。2020 年，S. Zhang 等人提出 ATSS[27]，其核心思想根据训练状态动态调整 IoU 阈值：将所有与 Ground Truth 存在重叠的预测框的 IoU 拟合到一个高斯分布之中，计算其均值 m 和标准差 v，得到设置阈值为 t=m+v，根据模型的训练动态调整 IoU 阈值；K. Kim 等人提出 PAA[28]，将 IoU 和目标置信度去代表 Anchor 的质量，计算出 Anchor Score，采用具有双成分的高斯混合模型拟合所有 Anchor Score，通过判断其归属于哪一个成分来区分正负样本；2020 年，H. Zhang 等人提出 Dynamic R-CNN[29]，通过在训练过程中线性提高 IoU 阈值来指导模型训练，提高了模型的性能；旷视的 B. Zhu 等人指出，一个样本是同时可以具有正样本和负样本两种属性，其提出 AutoAssign[30]，通过回归精度和模型输出的置信度去判断当前样本的正负样本属性权重。2021 年，Z. Ge 等人[31]使用最优传输理论（Optimal Transport）设计出 OTA（Optimal Transport Assignment），能够高效地对正负样本进行分配，提升检测性能。

（6）水下目标检测算法

近年来，水下目标检测越来越收到研究者的关注。自水下机器人检测比赛 URPC（Underwater Robot Picking Contest）举办以来，提供给了众多研究者们丰富的数据资源，为了研究者进行水下目标检测算法研究提供支持。2019 年，H. Huang 等人提出了三种为水下目标检测设计的数据增强方法：仿射变换、湍流模拟、不平衡光照模拟[32]；2020 年，W. Lin 等人针对水下生物群居导致严重遮挡的问题，提出了 RoIMIX[33]，在二阶段 Faster R-CNN 中的 RoI 之间进行 Mixup 操作，进行了数据增强，提升了模型在水下目标检测的鲁棒性；L. Chen 等人提出 SWIPENET[34]，借助高分辨率和语义丰富的特征图提高小目标检测的性能，同时提出了一种基于 Adaboost 集成学习的方法，增强模型的鲁棒性；B. Fan 等人提出了 FERNet[35]（如图 1.13），包含三个模块：组合连接骨干网络、感受野增强模块和精细回归模块，大大提升了模型在水下目标检测的性能；Z. Wang 等人[36]提出了 POISSON GAN，用泊松融合的方式将目标贴在背景上，然后用 CNN 修正得到完好的图像，同时该工作提出一阶段水下目标检测器 UnderwaterNet，在提升检测速度的同时得到高性能识别结果。2021 年，Z. Zhao 等人提出了一种视频水下目标检测器 Composited FishNet[37]，利用背景信息和前景信息的差异进行前后景分离特征提取，同时提出了增强路径集成网络提升模型多尺度检测能力。

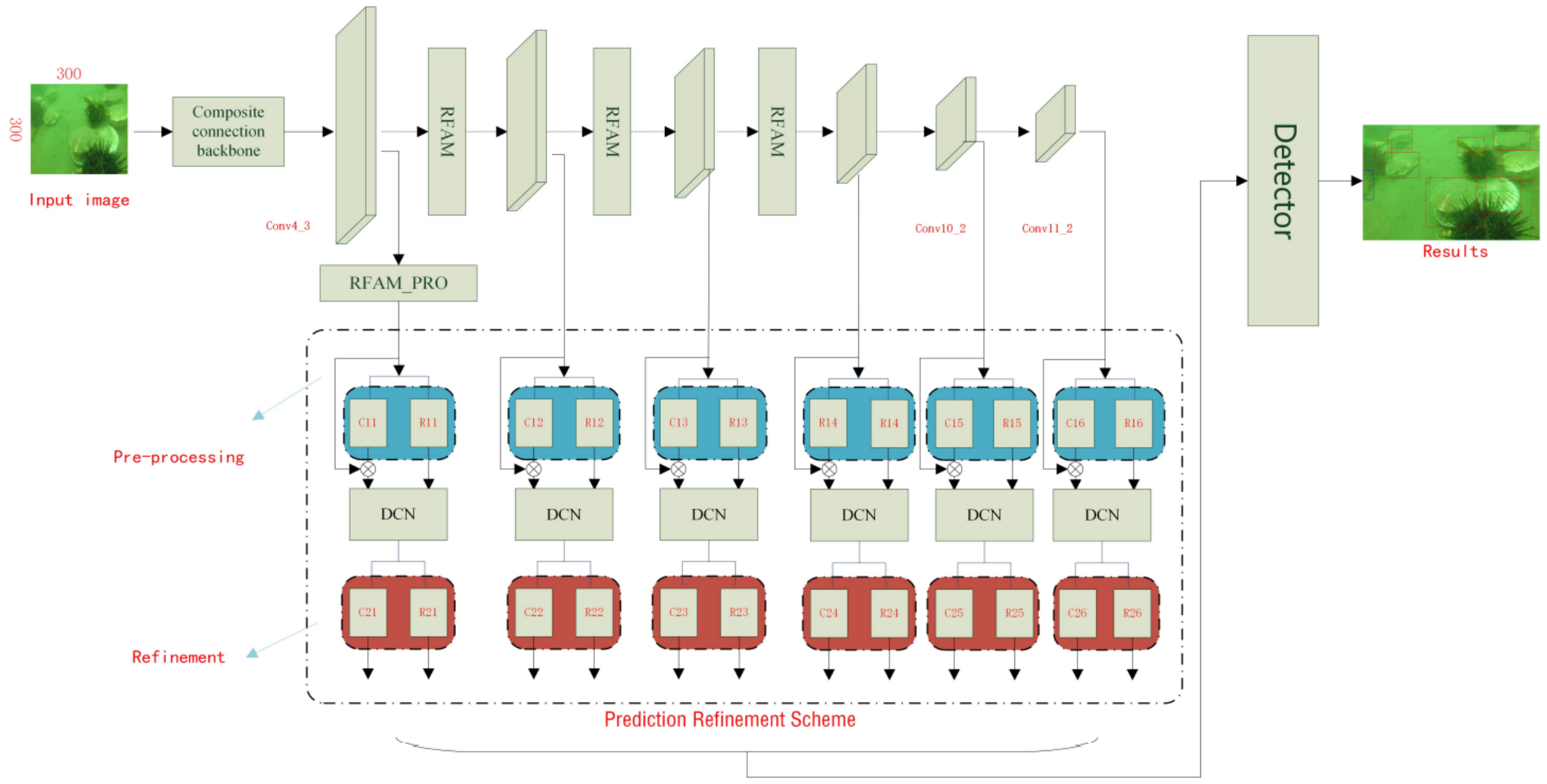

图 1.13 水下检测器 FERNet 的模型结构[35]
Figure 1.13 The pipeline of underwater detector FERNet

经过了研究者们多年的努力，可以总结出多个工作中水下目标检测中有效的设计思路：（1）数据增强：目前的水下目标检测数据集小，数据增强能够给现有检测算法极大的性能提升；（2）增大感受野：由于水下目标对比度低、模糊，局部的特征不足以对目标进行判断；（3）更强的特征提取器（骨干网络）：更加细粒且具有表征性的特征对于水下模糊目标的检测有着重要作用。

### 1.2.2 域泛化技术的研究现状

域泛化任务是指给定一系列数据来自多个源域，模型需要在源域数据上训练，希望在一个未曾见过但是和源域相关的目标域上测试仍然得到好的性能。深度学习模型在真实场景下进行应用的时候，往往会遇到数据集中没有见过的场景。如果人工智能系统需要走向真实应用，模型便需要对域迁移有强鲁棒性，这便使得域泛化技术的研究具有很强的必要性。

（1）基于数据增强的域泛化技术

深度学习是一个数据驱动的技术，在缺乏数据的情况，提升数据多样性是最为直接且简单提升模型泛化性能的方式。利用现有数据和对预想的应用场景的理解，生成多样的训练数据，可以大大提升模型的域泛化能力。例如，智能系统的快速训练使用的是合成数据，却需要应用到现实场景，因此域间隙不可避免。2017 年，J. Tobin 等人[38]对灯光、桌子的纹理等特性进行调整合成更多的数据训练，能够有效提升模型的泛化能力，相似的技术也在其他工作中都有采用。

对抗攻击也可以用于数据增强之中以提高域泛化性能。2018 年，S. Shankar 等人[39]

提出利用在域分类器上进行对抗攻击得到对抗样本，以此生成未知域的数据。2019 年，R. Volpi 等人[40]设计了一种迭代式的方法，基于源域数据生成一些虚拟的困难目标域数据。2020 年，K. Zhou 等人[41]提出一个对抗式学习的变换网络以用于数据增强；Z. Huang 等人提出 RSC[42]，通过遮掩掉高梯度像素值，让模型在更少信息的基础上给出更准确的结果，提升模型的泛化性能。

生成性的模型常被用于数据增强之中，例如风格迁移（Style Transfer）、对抗性生成网络（Generative adversarial Network，GAN）和变分自编码器（Variational AutoEncoder，VAE）。2019 年，M. M. Rahman 等人借助了 ComboGAN[43]去生成新数据；2020 年， K. Zhou 等人提出 L2A-OT[44]，借助 StarGAN 和最优传输理论生成新域数据；N.Somavarapu 等人[45]仅仅是利用风格迁移模型 AdaIN 进行数据增强，便可以在多数数据集上取得了 SOTA 效果。

根据对风格迁移和域泛化的多年探索，研究者发现风格信息和域信息是非常接近的两个概念。在风格迁移 AdaIN 的工作中，揭示了特征统计值（均值、标准差）和风格之间的相关性。基于此思路，2021 年，K. Zhou 等人提出 MixStyle[46]，在特征统计值层面上进行扰动进行数据增强，获得可观的性能提升。

（2）基于表征学习的域泛化技术

域泛化问题可视作让深度网络提取鲁棒表征以应对下游任务，研究者们会通过增强正则项去规范模型的训练，使得模型对于域间隙的影响降低。基于表征学习的域泛化常分为两类：域无关的表征学习和特征解缠。

**域无关的表征学习。**如果提取出来的特征和域无关，则模型的性能便不受域的变化的影响，即可以泛化到各式的场景中。对于域无关的表征学习主要存在三类：基于核方法、域对抗训练、显式特征对齐。

核方法是最经典的机器学习方法之一，其依赖于核函数将特征投射到一个高维空间上，不需要计算数据的特定坐标，只需要成对计算样本之间的内积。2011 年，G. Blanchard 等人[47]第一次提出域泛化的概念，并将核方法用于解决域泛化问题。2013 年，K. Muandet 等人[48]提出 DICA，最小化特征空间中核转换后的分布差异。2015 年，T. Grubinger 等人[49]使用迁移成分学习（Transfer Component Analysis，TCA）去缩短不同域之间的差距。2016 年，C. Gan 等人[50]基于 DICA，加入了归因正则化，进一步提升性能。2018 年，相比于处理边缘分布的 DICA，Y. Li 等人[51]让模型学习一种类条件分布的域无关特征。2019 年，S. Hu 等人提出多域辨别分析，以实现类别上的核方法。2020 年，D. Mahajan 等人[52]提出利用因果匹配的思路去学习解缠表征。

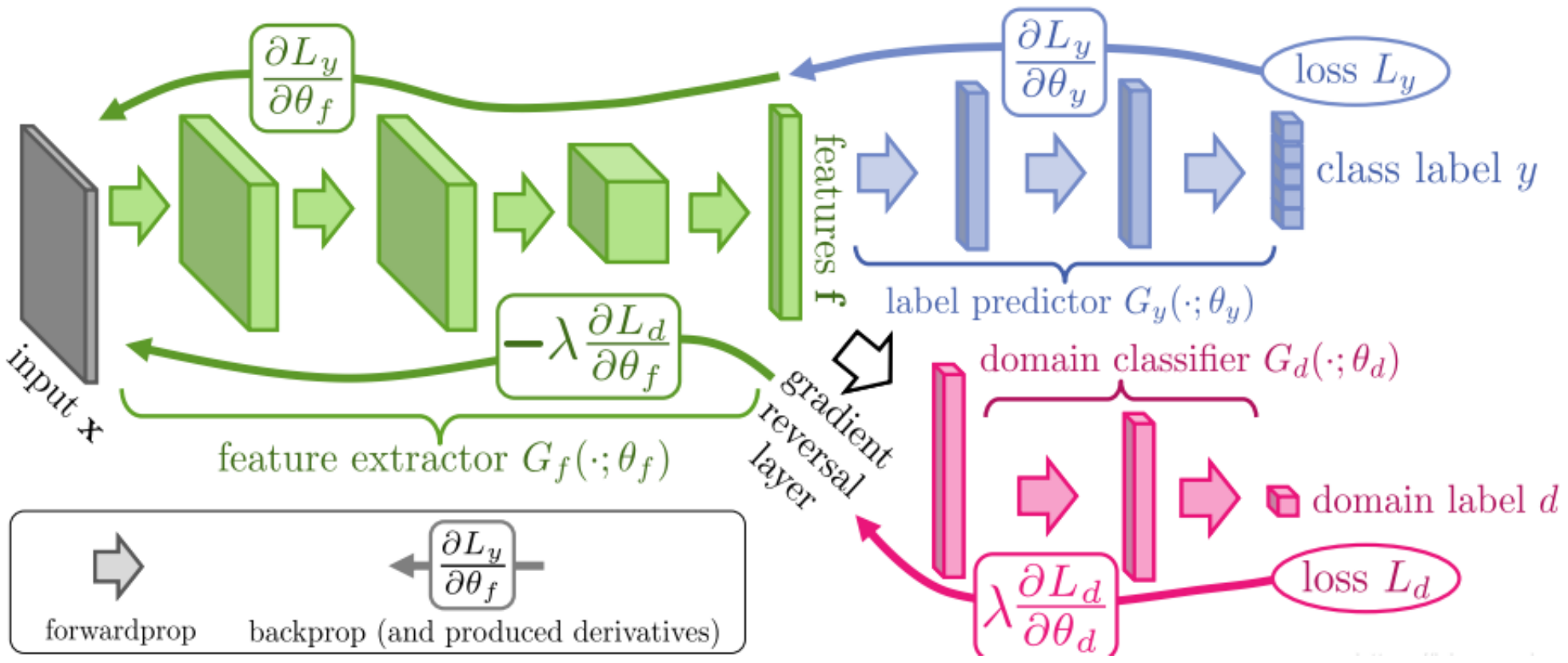

图 1.14　DANN 的模型结构[53]

Figure 1.14 Model structure of DANN

域对抗训练在有关域的任务中被广泛应用，其用于提取域无关特征。2016 年，Y. Ganin 等人提出了一种域自适应的对抗性训练方法 DANN[53]（如图 1.14），域辨别器作用于中间特征，目的是为了区分来自不同域的数据，在域辨别器之前接上一个梯度反转层，在端到端的训练过程，域辨别器的能力会越来越强，然而由于梯度反转层的效果，骨干网络倾向于欺骗域辨别器，因此会在提取特征的过程中去掉域的信息，以达到提取域无关特征的效果。该方法是为域自适应问题而提出的，但是其思想广泛用于了域泛化问题中。2018 年，H. Li 便将对抗训练的思路和自编码器结合提出 MMD-AAE[54]；Y. Li 扩展了 DANN，对于每一个类都引入一个域辨别器进行对抗训练[55]。2020 年，T. Matsuura 等人[56]研究认为，域的概念是模糊的，当如果没有一个显性域的区分（没有域标签），普通的对抗训练便无法进行，因此提出了一种域空间聚类对抗的方式，可以在没有域标签的情况下进行对抗训练；S. Zhao 等人[57]在对抗训练之外引入了熵正则化的概念，减小域条件分布之间的 KL 散度，以追使模型得到域无关特征。

显式特征对齐是一个相对直接的域泛化技术，通常希望捕捉数据的相似之处，在主损失函数上加入某种正则化项以拉近各个域之间的特征，缩减特征差异。2017 年，S. Motiian 等人[58]引入一个跨域对比损失，保证语义上的对齐。除此之外，一些方法引入一些度量表示特征分布上的差异，并使其最小化，例如最大均值差异（Maximum Mean Discrepancy，MMD）、二阶相关性、特征均值和方差、wasserstein 距离等。除了引入正则化项，特征归一化方法也用于特征对齐以缩减域间隙。2019 年，X. Pan 等人结合 IN（Instance Normalization）和 BN（Batch Normalization）提出了 IBN-Net[59]（如图 1.15），在域自适应任务上有较好的表现。2020 年，S. Seo 等人[60]借助 CIN 的思路（Conditional Instance Normalization），提出域特定的归一化模块，并在推断过程中进行参数融合提升泛化性能。2021 年，C. Sungha 等人提出 Instance Selective Whitening[61]，通过聚类搜索

IN 之后特征空间的敏感部分，然后进行特征对齐。

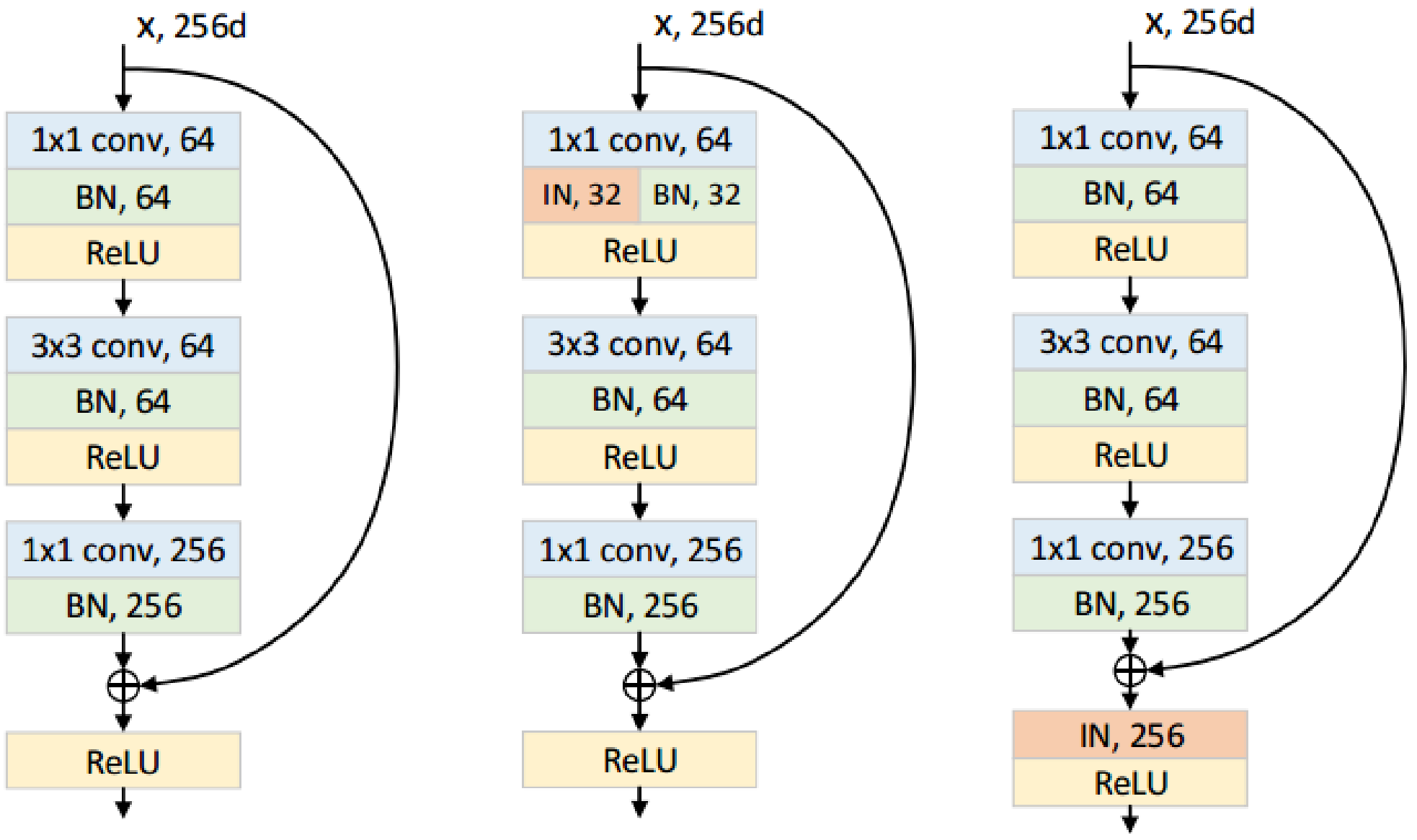

(a) ResNet (b) IBN-a (b) IBN-b

图 1.15 IBN-Net 的模型结构[59]

Figure 1.15 Model structure of IBN-Net

**特征解缠。**特征解缠旨在得到一个函数可以将样本映射到一个特征向量，该特征向量包含各种不同因素的变动，每一个维度（或者维度的子集）都仅仅对应一种或几种因素。基于特征解缠的域泛化技术通常会将特征分解为可理解的次级特征：域无关的特征和域特定的特征。2012 年，A. Khosla 等人提出 UndoBias[62]，此方法基于支持向量机（Support Vector Machine, SVM），他们将 SVM 的参数表示为$w_i = w_0 + \Delta_i$，i 为域索引，因此多个 SVM 有域间共有的参数，也有域特定的参数。2017 年，D. Li 等人将 UndoBias 的思路迁移到了 CNN 上[63]; Z. Ding 等人[64]为每一个域设计出域特定的网络，以及一个域无关的网络以处理所有域，以得到解缠特征，其中，低秩重建用于对齐两个网络的低阶特征。2020 年，X. Jin 等人提出 FAR[65]，通过类 Squeeze-and-Excitement 模块，得到注意力图，从而分离域无关信息和域特定信息；M. Ilse 等人提出 DIVA[66]，借助变分自编码器，分离域信息、类别信息以及其他信息；F. Qiao 等人[67]也采用了 VAE 的架构，提出了 UFDN，将域和感兴趣的图像属性作为隐层因素来解缠。

（3）基于训练策略改进的域泛化技术

除了在数据层面和特征层面上进行研究，研究者们也开始研究如何改变训练策略以提高域泛化能力。主要存在四种策略用于域泛化问题：集成学习、元学习、自监督学习

**基于集成学习的域泛化技术。**集成学习（Ensemble Learning）旨在通过构建多个学习器，并对多个学习器进行决策融合来完成任务[68]。对于域泛化来说，集成学习可以通过特定的模型结构和训练策略，挖掘多个源域的关系，实现泛化性能的提升。此类方

法假定任何样本都可以看作存在着多个源域成分，总预测结果可以看作是多个域网络预测结果的叠加。2018 年，A. D'Innocente 等人提出 D-SAM[69]，针对不同的源域学习不同的分类器，在目标域上会综合多个不同的分类器进行决策。2020 年，M. Segu 等人[70]为每一个源域都设计了一个域依赖的 BN，在推断过程中的 BN 参数是由训练的多个域依赖 BN 的参数线性组合而成；K. Zhou 等人[71]提出了 DAEL，此模型结合了一个在多个源域参数共享的 CNN 和多个域特定的分类器，每一个分类器都可以看作是一个域的专家模型，DAEL 旨在以这些专家模型相互训练彼此，以提高这些分类器在未见域的泛化性能；P. Chattopadhyay 等人[72]提出 DMG，采用的是 dropout 的思路，针对不同的源域在特征图上学习不同的Mask，在决策过程中将在不同Mask下的结果做决策融合。

**基于元学习的域泛化技术。**元学习（Meta-learning）旨在从多个任务上得到一个通用性的模型，这个思路自然也可以移植到域泛化问题当中。2017 年，C. Finn 等人[73]提出 MAML，此工作是小样本学习的工作。2018 年，MAML 的思路被 D. Li 等人运用到了域泛化问题之中，提出了 MLDG[74]。如图 1.16，MLDG 将源域分成元训练和元测试以在训练集中模拟域间隙，减小元测试上的错误率这一优化目标指导梯度回传，得到泛化性能更强的表征；Y. Balaji 等人提出学习一个元正则化项（MetaReg）[75]。2019 年，Y. Li 等人提出特征批评学习，设计出了一个元优化器；Q. Dou 等人[76]使用和 MLDG 类似的思想，额外引入了两个补充损失函数显式地对语义特征进行正则化。2020 年，Y. Du 等人[77]引入了信息瓶颈这个概念，提出了 MetaVIB，其对同类不同域样本隐式分布之间的 KL 散度进行正则化。

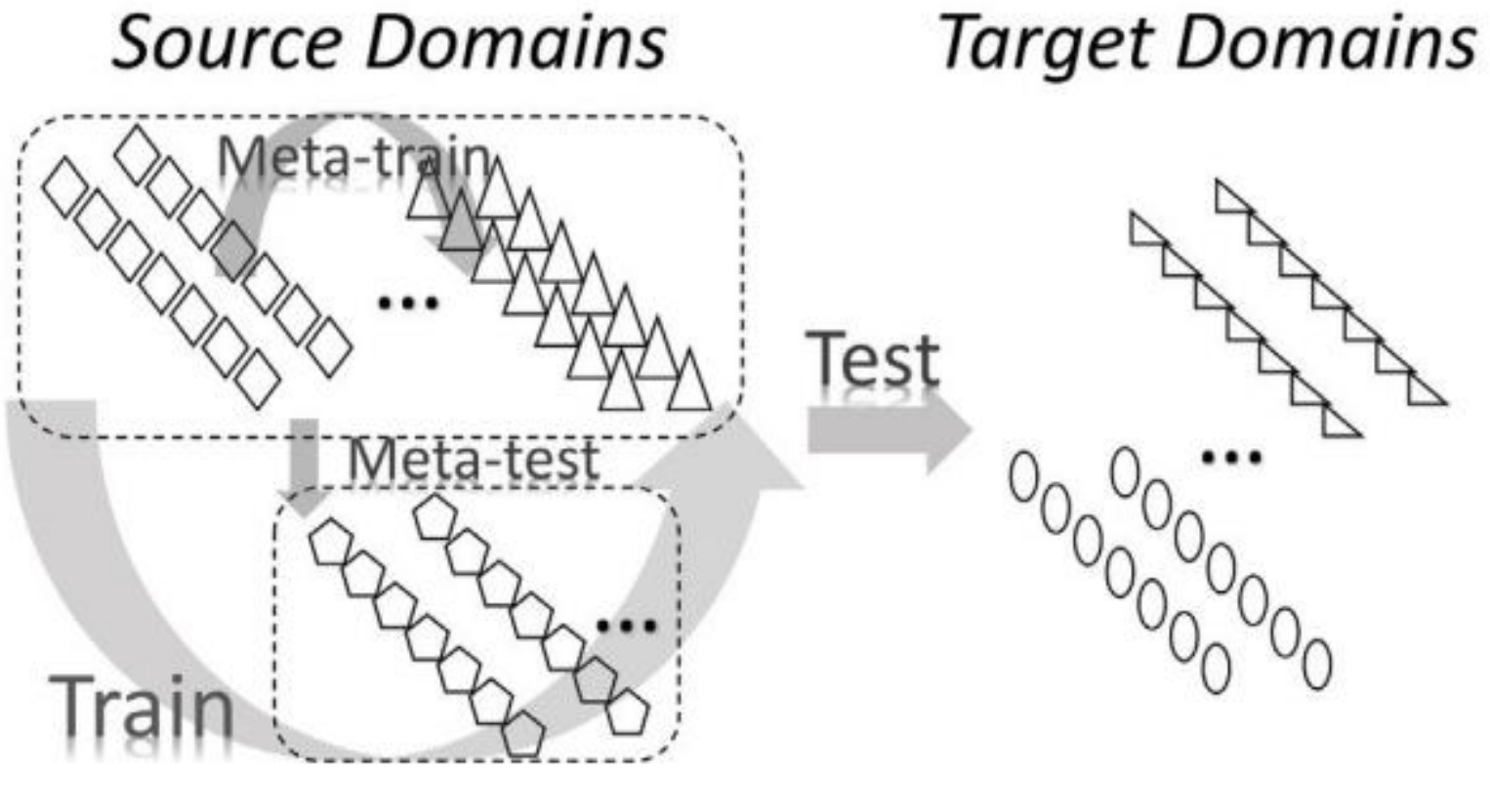

图 1.16　MLDG 的示意图[74]
Figure 1.16 Illustration of MLDG method

**基于自监督学习的域泛化技术。**自监督学习（Self-supervised Learning）在近期收到了研究者的火热关注，其旨在利用数据内在的不变性自我提供标签，让模型可以免去真标签进行大规模的无监督学习。由此启发，2018 年，F. M. Carlucci 等人提出 JiGEN[78]（如图 1.17），将拼图任务（Jigsaw Puzzles）这一自监督任务引入域泛化的监督训练范

式之中，提升了域泛化能力。2020 年，S. Wang 等人[79]融合了两个自监督技术：拼图学习和对比学习 MoCo[80]，在 JiGEN 之上进一步提升了性能。

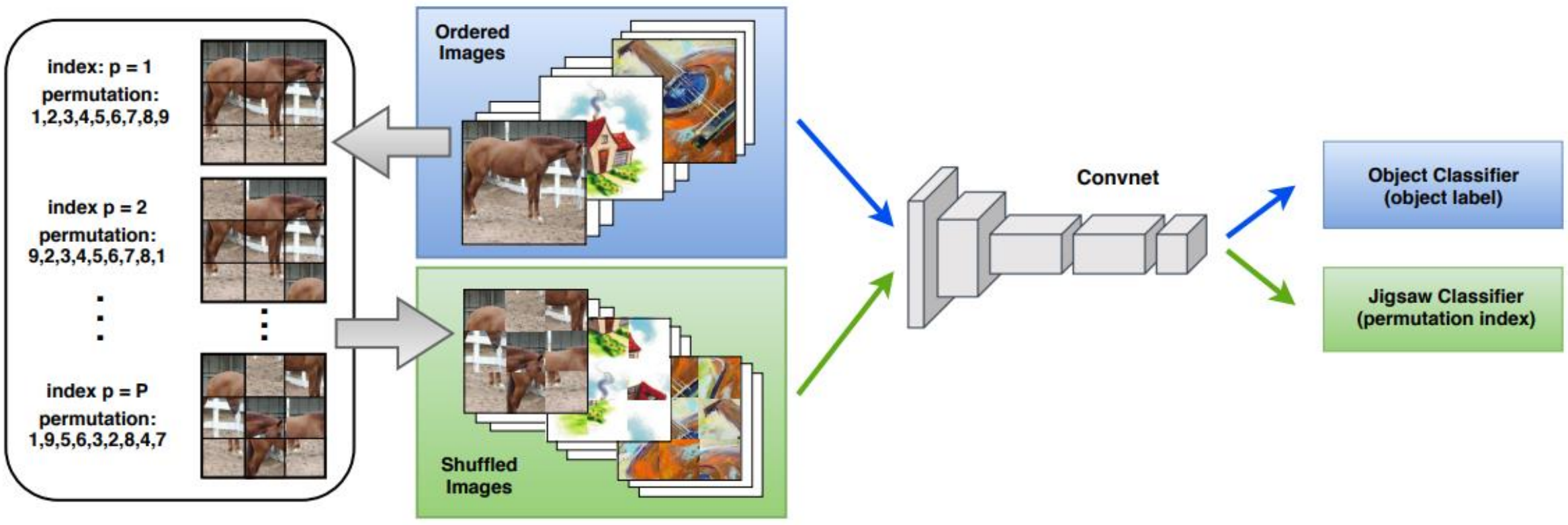

图 1.17　自监督学习 JiGEN 的训练流程图[78]

Figure 1.17 The pipeline of self-supervised learning method JiGEN

此外，还有一些没有被分类的基于训练策略改进的域泛化技术，如 Epi-FCR[81]等。

总结：现如今研究者主要从数据增强、表征学习、训练策略三个角度对域泛化技术进行研究和突破，有了不少的成果，其涉及的领域包括目标识别、语义分割、强化学习、行为识别、人脸防欺诈、行人重识别和视频理解。然而，研究在目标检测上的域泛化技术的工作却少之又少。本文所关注的正好是目标检测的域泛化技术，希望通过本文的研究能够填补上这一研究空白。

## 1.3　相关理论知识

### 1.3.1　水下图像成像原理

水下图像往往有严重的颜色模糊、光照复杂且不均匀、可见度低的问题。因此，清澈的海中水下的能见度大概为 20 米，而在浑浊的水域之中能见度只能达到 5 米。造成一系列问题的原因是因为水对于光的吸收和散射严重限制了水下成像系统的表现。前向散射造成图像细节的模糊，后向散射降低图像对比度，水中溶解的不同物质和浮游物更加重了图像的模糊和失真。此外，光在水中的衰减有着波长的相关性，并随距离增大颜色丢失和失真会越严重。较短的波长的光，如蓝光和绿光，衰减程度较小；而较长的波长的光，如红光，衰减程度较大，大部分在传播的过程中被水所吸收，因此水下的环境大多是蓝色和绿色的而少有红色的水下图像。

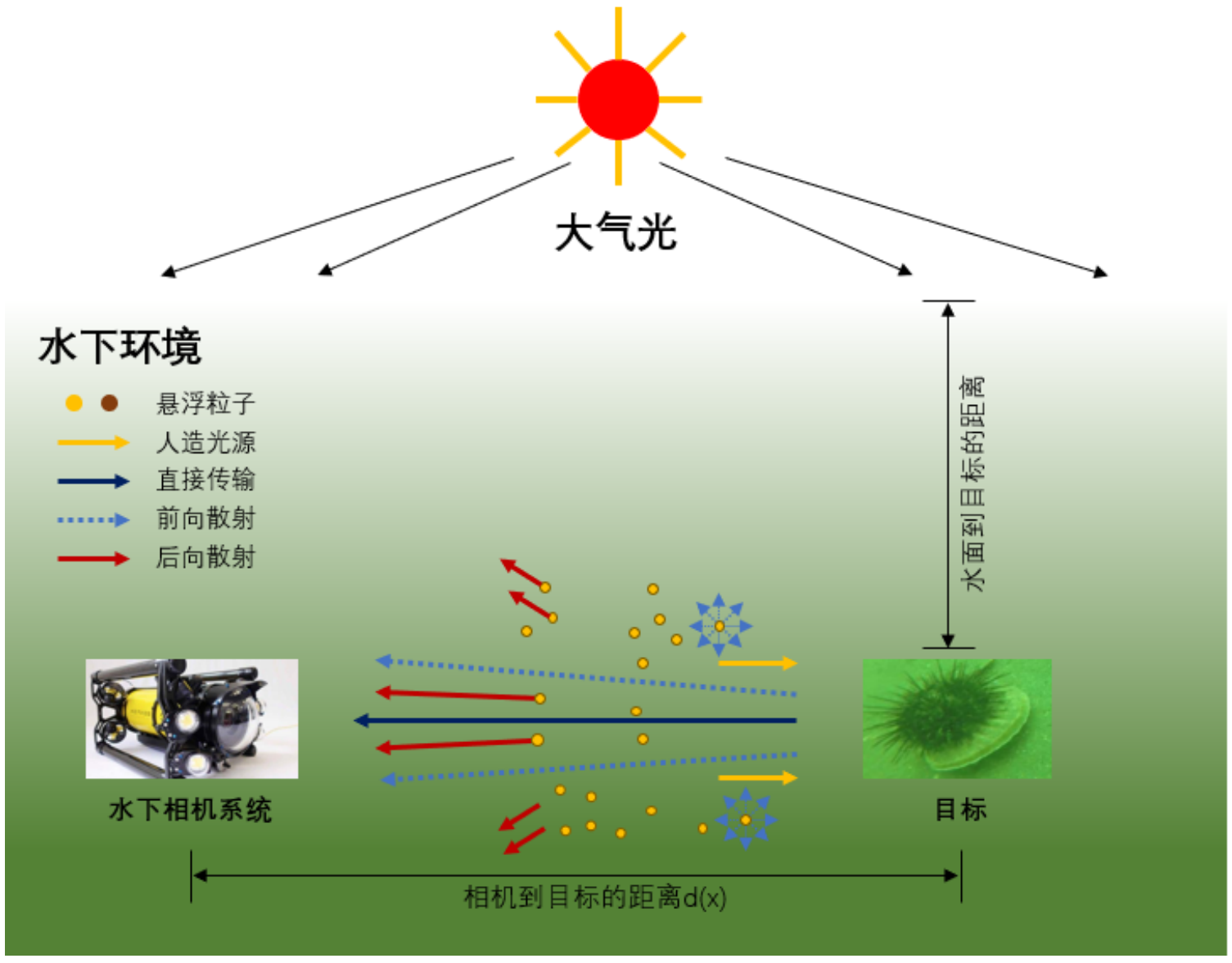

图 1.18　水下图像成像模型
Figure 1.18 Underwater Image Formation Model

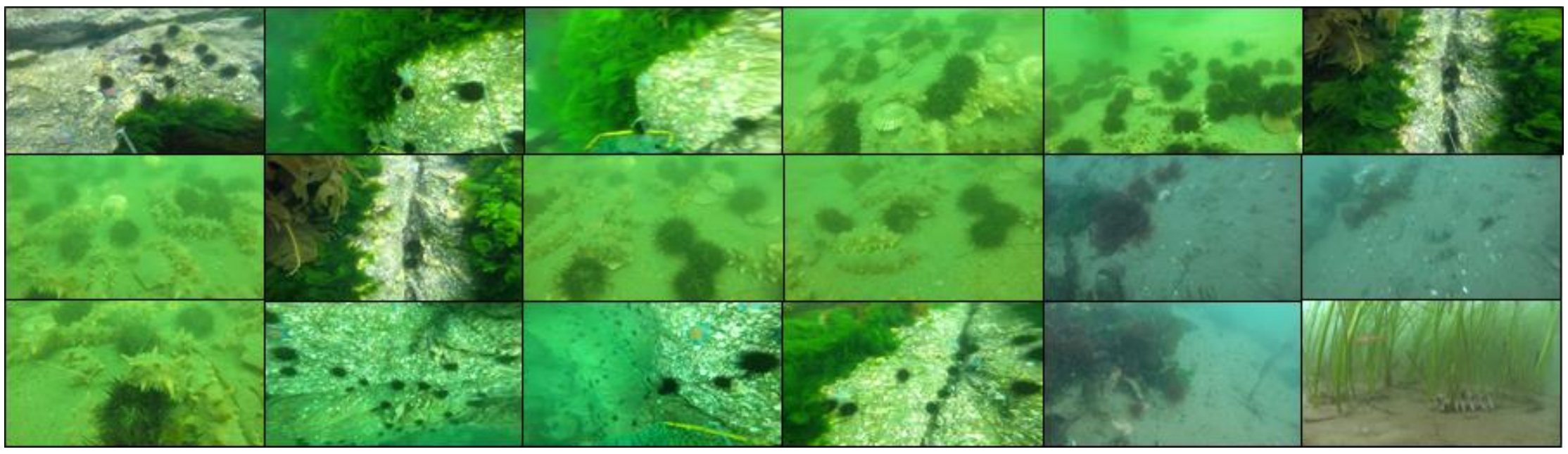

图 1.19　水下图像示例
Figure 1.19 Illustration of underwater images

为了获得更好的水下成像质量，早期工作主要关注如何得到更高质量的成像结果。经典的水下图像成像物理模型是由 Jaffe-McGalmery[82]提出的 IFM（Image Formation Model）模型，水下相机的图像由三部分组成：直接传输（Direct Component）、后向散射（Backward Scattering Component）、前向散射（Forward Scattering Component），如图 1.18。具体的，水下图像成像模型可以表示为：

$$\mathrm{I}_\lambda(x) = J_\lambda(x)t_\lambda(x) + B_\lambda\big(1 - t_\lambda(x)\big), \lambda \in \{R, G, B\}, \tag{1.1}$$

其中，$x$ 为水下图像像素的坐标，$\lambda$为红绿蓝三通道，$\mathrm{I}_\lambda(\mathrm{x})$为水下相机捕获到的各通道图像，$J_\lambda(x)$为各通道潜在清晰图像，$B_\lambda$为各通道背景光，$\mathrm{t}_\lambda(\mathrm{x})$为各通道对应的传输率，其描述的是场景物体反射光中没有被散射或者吸收到达相机的场景光，具体为

$$\mathrm{t}_\lambda(x) = e^{-\beta(\lambda)d(x)} = Nrer(\lambda)^{d(x)}, \tag{1.2}$$

其中，$\beta(\lambda)$为光谱体积衰减系数，$d(x)$为水下目标与相机的距离，$Nrer(\lambda)$为标准化残余能量比，表示光在水中传输单位距离后光能量的衰减值，其值至于光的波长。

如图 1.19 所示是自然光照射下的水下图像，图像整体偏绿或者偏蓝，且水中浮游物对自然光的吸收和散射导致图像出现雾化，其可见程度视当前水域情况而定。

### 1.3.2 域泛化的相关知识

（1）域的定义：假定$\mathcal{X}$表示非空输入空间，$\mathcal{Y}$为输出空间。一个域由从某一分布中采样的数据组成。令

$$\mathrm{S} = \{(\mathbf{x_i}, \mathrm{y_i})\}_{i=1}^{\mathrm{n}} \sim P_{XY}, \tag{1.3}$$

其中$\boldsymbol{x} \in \mathcal{X} \subset \mathbb{R}^d$，$y \in \mathcal{Y} \subset \mathbb{R}$表示标签，$P_{XY}$表示输入样本和输出标签的联合分布。$X$，$Y$表示输入和输出的随机变量。

（2）域泛化的定义：在域泛化的问题设置中，给定 M 个训练域（源域）

$$\mathrm{S_{train}} = \left\{S^i \middle| i = 1, \ldots, M\right\}, \tag{1.4}$$

其中，$\mathrm{S^i} = \left\{\left(\mathrm{x_j^i}, \mathrm{y_j^i}\right)\right\}_{\mathrm{j=1}}^{\mathrm{n_i}}$ 表示第 $i$ 个域。每一对域所代表的联合分布均不相同，

$$\mathrm{P}_{XY}^{\mathrm{i}} \neq \mathrm{P}_{XY}^{\mathrm{j}}, 1 \leq \mathrm{i} \neq j \leq \mathrm{M}. \tag{1.5}$$

域泛化的目标是从 $M$ 个训练域中学习一个鲁棒的模型$h: \mathcal{X} \to \mathcal{Y}$，以在未见过的测试域$\mathrm{S_{test}}$中实现最小的预测错误率（测试域不可以在训练阶段中被获取，且$\mathrm{P_{XY}^{test}} \neq \mathrm{P_{XY}^{I}}$，$\mathrm{i} \in \{1, \ldots, \mathrm{M}\}$，即

$$\min_{\mathrm{h}} \mathbb{E}_{x,y \in S_{test}}[\ell(h(x), y)], \tag{1.6}$$

其中$\mathbb{E}$是期望，$\ell(.,.)$是损失函数。

（3）域自适应错误下界：域自适应是一个与域泛化非常相似的任务，对于域自适应的分析同时也有利于域泛化技术的开发。首先回顾域自适应的定义，对于一个二元分类任务，将在源域上的真实函数定义为 $h^{*S}: \mathcal{X} \to [0,1]$，目标域上的真实函数定义为$h^{*t}$。$h, h': \mathcal{X} \to [0,1]$为假设空间$\mathcal{H}$上的两个假设（分类器），两个分类器在源域的错误率可以表示为：

$$\epsilon^{\mathrm{s}}(\mathrm{h}, \mathrm{h'}) = \mathbb{E}_{\mathrm{X \sim P_X^S}}[h(x) \neq h'(x)] = \mathbb{E}_{\mathrm{x \sim P_X^S}}[|h(x) - h'(x)|]. \tag{1.7}$$

相似地，也可以定义其在目标域的错误率，只需要将上式中的 $x$ 从$P_X^t$采样即可。因此可以定义：

$$\epsilon^{s}(h) = \epsilon^{s}(h, h^{*s}), \tag{1.8}$$

$$\epsilon^{t}(h) = \epsilon^{t}(h, h^{*t}), \tag{1.9}$$

$\epsilon^{s}(h)$和$\epsilon^{t}(h)$分别为分类器 $h$ 在源域和目标域上的风险，而域自适应和域泛化的目标就是最小化$\epsilon^{t}(h)$，既然没办法获取$h^{*t}$，便希望用$\epsilon^{s}(h)$去表示$\epsilon^{t}(h)$的下界[83]：

$$\epsilon^{t}(h) \leq \epsilon^{s}(h) + 2d_1(P_X^s, P_X^t) + \min_{P_X \in \{P_X^s, P_X^t\}} \mathbb{E}_{x \sim P_X}[|h^{*s}(x) - h^{*t}(x)|], \tag{1.10}$$

其中，

$$d_1(P_X^s, P_X^t) \coloneqq \sup_{A \in \mathcal{X}} |P_X^s(A) - P_X^t(A)|, \tag{1.11}$$

是两分布的全变分，$\mathcal{X}$是 $X$ 的 sigma 域。第二项描述了两分布之间的距离，第三项描述了标注函数的差异。然而，全变分是一个很大的距离，会让这个下界不够紧凑，Ben-David 等人[83]提出了另外一个下界：

$$\epsilon^{t}(h) \leq \epsilon^{s}(h) + 2d_{\mathcal{H}\Delta\mathcal{H}}(P_X^s, P_X^t) + \lambda_{\mathcal{H}}, \tag{1.12}$$

其中，$\mathcal{H}\Delta\mathcal{H}$散度的定义为：

$$d_{\mathcal{H}\Delta\mathcal{H}}(P_X^s, P_X^t) \coloneqq \sup_{h, h' \in \mathcal{H}} |\epsilon^{s}(h, h') - \epsilon^{t}(h, h')|, \tag{1.13}$$

理想联合风险$\lambda_{\mathcal{H}} \coloneqq \inf_{h \in \mathcal{H}} [\epsilon^{s}(h) - \epsilon^{t}(h)]$衡量的是$\mathcal{H}$在预测任务上的复杂度。

（4）平均风险估计误差界：当目标域数据完全不可知，衡量所有可能域的平均风险成为了一个关键问题。假定所有可能的目标域服从一个潜在的超分布$\mathcal{P}$，$P_{XY}^t \sim \mathcal{P}$，源域分布也来自于同样的超分布，即$P_{XY}^1, \ldots, P_{XY}^M \sim \mathcal{P}$。对于一个分类器 h，它的在所有目标域上的平均风险为：

$$\mathcal{E}(h) \coloneqq \mathbb{E}_{P_{XY} \sim \mathcal{P}} \mathbb{E}_{(\mathbf{x}, y) \sim P_{XY}} [\ell(h(P_X, \boldsymbol{x}), y], \tag{1.14}$$

其中，$\ell$是损失函数。想要完好估计出这个期望是不可能的，但可以通过服从于$\mathcal{P}$的有限源域分布，以及每个源域分布上有限的样本去估计这一期望，有：

$$\hat{\mathcal{E}}(h) \coloneqq \frac{1}{M} \sum_{i=1}^{M} \frac{1}{n^i} \sum_{j=1}^{n^i} \ell\left(h\left(\mathcal{U}^i, \boldsymbol{x}_j^i\right), y_j^i\right), \tag{1.15}$$

其中，$\mathcal{U}^i \coloneqq \{x_j^i | \left(x_j^i, y_j^i\right) \in S^i\}$是来自第 i 个源域的训练数据。

（5）域泛化错误下界：令$\epsilon^1, \ldots, \epsilon^M$为源域风险，$\epsilon^t$为目标域风险，且假定域之间存在协变量偏移（域间隙）。I. Albuquerque[84]等人用源域分布构成的凸包去模拟目标分布：$\Lambda \coloneqq \{\sum_{i=1}^{M} \pi_i P_X^i | \pi \in \Delta_M\}$，$\Delta_M$是 M-1 维的单纯型，每一个$\pi$都代表这一个正则后的混合

权重。和域自适应的情况相似，分布差异可以用$\mathcal{H}$去衡量。令 $\gamma := \min_{\pi\in\Delta_M} d_{\mathcal{H}}(P_X^t, \sum_{i=1}^{M}\pi_i P_X^i)$，其最小值点$\pi^*$是$P_X^t$和凸包$\Lambda$的距离，$P_X^* := \sum_{i=1}^{M}\pi_i^* P_X^i$ 是凸包$\Lambda$的最佳近似，$\rho := \sup_{P_X', P_X''\in\Lambda} d_{\mathcal{H}}(P_X', P_X'')$是凸包$\Lambda$的直径。于是有：

$$\epsilon^{t}(h) \leq \sum_{i=1}^{M}\pi_i^*\epsilon^i(h) + \frac{\gamma+\rho}{2} + \lambda_{\mathcal{H},(P_X^t,P_X^*)}, \tag{1.16}$$

其中$\lambda_{\mathcal{H},(P_X^t,P_X^*)}$是目标域和其他源域构建的最佳近似$P_X^*$的理想联合风险。这一下界驱动研究者研发基于域无关的表征学习的域泛化技术，即同时最小化每一个源域的风险的同时（式子第一项），也减少了源域和目标域之间的分布距离。

## 1.4 本文内容与结构安排

上文介绍了目标检测技术、域泛化技术的研究背景和研究难点，以及介绍了域泛化水下目标检测的研究意义。据上文所分析，目标检测技术的工作主要围绕通用场景和通用目标进行，对于水下环境的目标检测研究甚少；同时，域泛化技术的研究主要集中在分类任务上，少有研究域泛化目标检测的工作，其在分类上研究成熟的技术并不能很好的适应在检测上的任务，且其性能也未能保证。因此本文从实际问题出发，基于水下场景提出了域泛化的目标检测这一任务，研究得到一个高性能的、鲁棒性的水下目标检测器，能够在不同水质下得到良好的检测性能，本文的总体技术框架如图 1.20 所示。

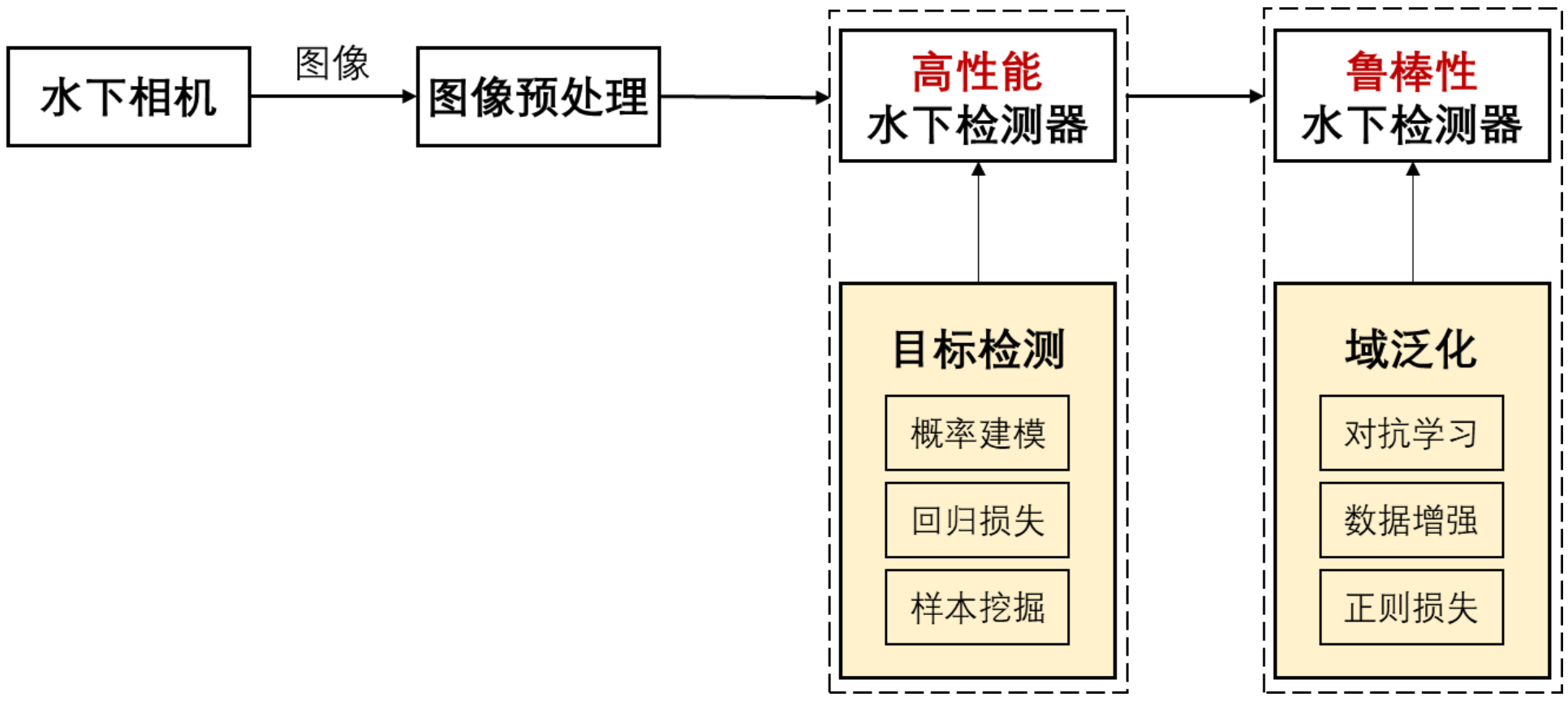

图 1.20 水下鲁棒目标检测技术总体框架
Figure 1.20 Illustration of underwater images

本文一共五章。第一章综述了目标检测、域泛化技术、水下目标检测的研究背景、研究现状，并基于此提出域泛化的水下目标检测这一任务。在研究现状中，对于目标检

测，本文从二阶段检测算法、一阶段检测算法、多尺度特征融合、Anchor-free 系列算法、正负样本分配、水下目标检测六个板块进行介绍；对于域泛化技术，本文从数据增强、表征学习、训练策略三大板块进行介绍。最后一小节补充了水下成像模型和域泛化问题的相关知识。

第二章提出了一个高性能的水下目标检测器。该方法对现有的目标检测器的一阶段算法和二阶段算法进行融合，用一阶段的技术增强二阶段的检测性能，同时增强二阶段的检测速度。同时，本章引进一个概率分布建模来表示二阶段目标检测，不同于现有的使用条件概率分布进行预测的二阶段方法，本方法结合一阶段和二阶段的信息，组成完整的边缘分布进行检测。另外，本方法还使用了有着更快收敛速率的损失函数。最后，本章提出一种新的困难样本挖掘的方法，可以不增加训练时间的同时提高性能，并且于本章提出的概率建模有着非常良好的兼容性。

第三章将会探究水下域间隙问题，以证明水下域间隙的普遍存在以及水下域泛化技术的必要性，并提出了 DG-YOLO 模型，从表征学习的角度出发，加入了域对抗训练和无关风险最小化惩罚项，迫使骨干网络捕获域无关信息，以提高模型在跨水质下的鲁棒性。

第四章将会深入研究水下域泛化问题。建立正式的水下域泛化数据集 S-UODAC2020，并从数据增强和对比学习的思路，在数据端上利用风格迁移进行数据增强，在特征端上做线性插值获得采样域空间，对特征进行选择性正则化，让模型学得域无关的特征，进一步提高检测模型在跨水质下的检测鲁棒性。

第五章对论文的内容和硕士期间的工作进行总结，以及对本文提出的研究方向、研究方法进行未来展望。

# 第二章　基于困难样本挖掘的概率型二阶段检测器

水下场景有着极端的光照、低对比度、遮挡以及生物拟态等问题，水下图像往往包含着多样的、大量的困难样本，通用的目标检测技术经常会在水下困难样本上失效。为了提升目标检测器在困难样本上的检测效果，本章基于困难样本挖掘的思路提出了一种概率型二阶段的水下检测器 Boosting R-CNN。首先，Boosting R-CNN 利用了现今一阶段检测器积累的技术经验，将此应用到二阶段的区域提议网络（Region Proposal Network）上，增强第一阶段的检测能力，并且本章提出了一个新的回归损失函数快速交并比损失（Fast IoU Loss），能显著提高模型的回归能力。其次，本章以贝叶斯的角度出发解析二阶段检测器，一阶段的输出为先验概率，二阶段的输出结果为似然概率，通过分析发现大部分的二阶段检测器仅仅使用不完备的似然概率进行结果预测，而本章对此进行修正，使用完备的边缘分布输出检测结果，能够有效提升检测效果。最后，本章提出了一个根据区域提议网络的错误的软采样方法提升再权重模块（Boosting Reweighting，BR），当区域提议网络错误地估计了某一个样本的先验概率，提升再权重模块就会在训练阶段放大这一个样本的损失，同时减少简单样本的损失，提升模型对困难样本的鲁棒性。本章提出的方法在水下数据集 UTDAC2020 和 Brackish 上取得了比当前方法更高的性能，同时在通用目标检测数据集 Pascal VOC 和 COCO 上也有不错的性能表现。

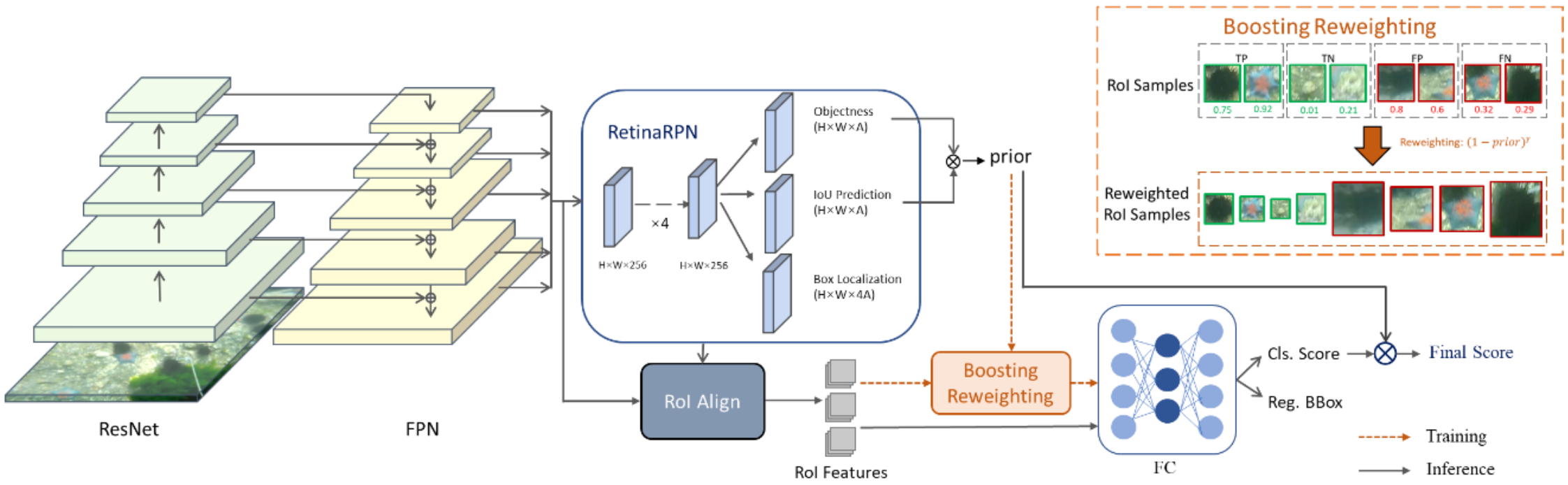

图 2.1 Boosting R-CNN 的整体结构

Figure 2.1 The overall architecture of Boosting R-CNN

图 2.1 展示了 Boosting R-CNN 的网络结构，其主要包括骨干网络（Backbone）、特征金字塔（Feature Pyramid Network）、视网膜区域提议网络（Retina Region Proposal Network，RetinaRPN）、RoI Align、提升再权重模块、检测头。具体而言，输入的水下图像通过骨干网络抽取特征，其特征经过特征金字塔之后得到多层特征图。对于每一层特征图，RetinaRPN 会提供高质量的提议框，并提供相应的先验概率。RoI Align 把

提议框从特征图上截取得到感兴趣区域（Region of Interest，RoI）。在训练阶段，提升再权重模块将先验概率被错误估计的困难样本的分类损失放大，将容易样本的分类损失减小，用于检测头的训练。在推断阶段，模型跳过推荐再权重模块，将先验概率和检测头得到的后验概率相乘得到最终的分类分数。

## 2.1　视网膜区域提议网络

近年来，一阶段检测器的工作有了丰富的成果。这些成果也可以为二阶段检测器的研究所用，因为二阶段检测器的第一个阶段，即区域提议网络，可被视为一个一阶段检测器。本小节旨在使用常见的一阶段检测器技术指导设计一个新的区域提议网络，因为其与视网膜网络（RetinaNet）有则相似之处，因此本章称之为视网膜区域提议网络（RetinaRPN）。

### 2.1.1　特征金字塔层次

特征金字塔（FPN）利用了骨干网络中每一个阶段的最后一个特征图，通过自上而下路径和中间链接进行信息重组，以获得多尺度信息，增强了特征的空间信息和语义信息。通常在二阶段检测器中，FPN 输出的 5 层特征图 P2-P6 的大小分别为原图的{1/4，1/8，1/16，1/32，1/128}。而常用的一阶段检测器中，会在最后的一层特征图上再进行一次下采样，然后选择为原图的{1/8，1/16，1/32，1/128，1/256}大小的特征图 P3-P7 进行检测。本章选择后者作为特征图的输入，而这一改动能大幅减少计算量和 GPU 显存的使用。

### 2.1.2　网络结构设计

在 Faster R-CNN 中，RPN 只有一个共享的卷积层以及分别用于回归和分类的两个卷积层。流行的一阶段检测器，比如 RetinaNet、FCOS、ATSS 等，均使用 4 个卷积层，并且将原本的批归一化（Batch Normalization，BN）换成了组归一化（Group Normalization，GN），组数为 32。本章也对 RetinaRPN 做同样的改动。更多的卷积层有更强的特征提取能力，能够提取到更准确的提议框。

### 2.1.3　Anchor 设计及其正负样本分配

一阶段检测器有两种：基于 Anchor 的检测器和无 Anchor 检测器。根据实验发现，Anchor 在水下目标检测相比于无 Anchor 能够取得更高的性能。因此 RetinaRPN 是基于 Anchor 的检测器。RetinaRPN 基本上沿用了 RetinaNet 的 Anchor 设计，即使用三种长宽比{1:2，1:1，2:1}和{$2^0$, $2^{1/3}$, $2^{2/3}$}倍于当前特征图缩放步长的尺寸。因此，对于特征图上的每一个像素，都能够拥有九个 Anchor。

关于 Anchor 的正负样本分配策略，在 Faster R-CNN 的 RPN 中，与 Ground Truth 的 IoU 大于 0.7 的被视为正样本，小于 0.3 的被视作负样本，IoU 在[0.3-0.7]的区间之中的不考虑回归；而在 RetinaNet 中，IoU 大于 0.5 的被视为正样本，小于 0.4 的被视作负样本，IoU 在[0.4-0.5]的区间之中的不考虑回归。近期，对动态 IoU 阈值的研究取得了卓越的成效，Dynamic R-CNN 使用采样+滑动平均的方式确定 IoU 阈值，ATSS 采用高斯分布动态调整 IoU 阈值，而 PAA 则使用高斯混合模型进行正负样本分配。而本方法采用了最简单的方法：以 0.5 为阈值，IoU 大于 0.5 的为正样本，IoU 小于 0.5 的为负样本。实验证明这种简单的分配策略能获得最好的性能。

### 2.1.4 监督任务和损失函数设计

不同于常见的 RPN，RetinaRPN 有三个子任务：回归、分类和 IoU 预测。

对于分类任务，RetinaRPN 负责判断当前 Anchor 是否为正样本，并且输出相应的目标似然概率（Objectness Likelihood）。常见的分类损失有两种：交叉熵损失和 Focal loss。二分类交叉熵损失的表达式为：

$$L_{ce} = \sum_i -[y_i \log(p_i) + (1 - y_i)\log(1 - p_i)], \tag{2.1}$$

其中，$y_i \in \{0,1\}$表示样本 i 的标签（Label），$p_i \in [0,1]$为模型输出，其表示样本 i 为正类（即 1）的概率。

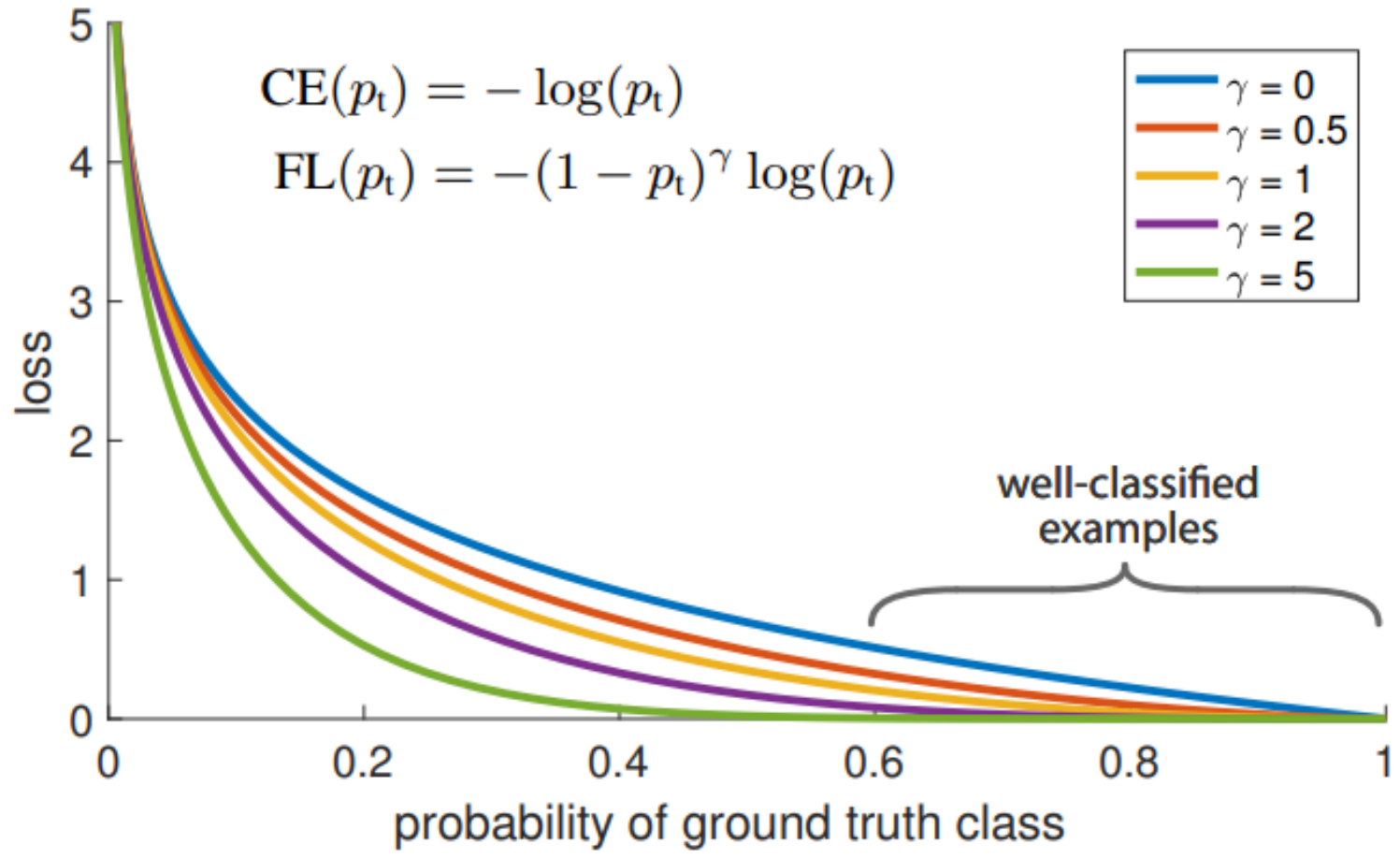

图 2.2 交叉熵损失和 Focal loss 的函数曲线[13]
Figure 2.2 The function curve of cross entropy (CE) loss and Focal loss (FL)

Lin 等人[13]认为，一阶段检测器和二阶段检测器性能差异的主要原因是在检测任务中的正负样本不平衡问题，大量简单负样本占据了梯度的主要部分。因此其提出了 Focal loss，其减少了简单负样本的权重。 Focal loss 的表达式为：

$$\mathrm{L_{fl}} = \begin{cases} -\alpha(1-p_i)^{\gamma} log p_i, & y=1, \\ -(1-\alpha)p_i^{\gamma}\log(1-p_i), & y=0, \end{cases} \tag{2.2}$$

其中，γ为专注参数，$(1-p_i)^{\gamma}$和$p_i^{\gamma}$为调节因子，α为平衡参数。当样本被错误分类时，调节因子便会放大，loss 便不受影响；当样本被正确分类时，调节因子会趋近于 0，因此该样本的权重就会变小。RetinaRPN 选择 Focal loss 而非交叉熵损失作为分类损失函数，并且将 RPN 原有的采样模块去掉。

对于回归任务，模型会基于当前的 anchor 输出一个预测框，并希望它与真实框尽可能接近。几种常见的回归损失的介绍如下：

**平均绝对值损失（Mean Absolute Error，MAE/L1 loss）**：其公式如下所示：

$$\mathrm{L_1}(e) = |e|, \tag{2.3}$$

$$\frac{\partial \mathrm{L_1}(e)}{\partial e} = \begin{cases} 1, & if\ x \geq 0 \\ -1, & otherwise' \end{cases} \tag{2.4}$$

其中$\mathrm{f(x_i)}$为预测值，$\mathrm{y_i}$为真实值，e 为误差损失值。其中 e 的计算公式为：

$$\mathrm{e} = \mathrm{t_i} - t_i^*, \qquad \mathrm{i} \in \{\mathrm{x,y,w,h}\}$$

$\mathrm{t_i}$和$\mathrm{t_i^*}$表示框的 4 坐标的编码值，具体而言，编码方式为[9]：

$$\mathrm{t_x} = \frac{x-x_a}{\mathrm{w_a}}, \mathrm{t_y} = \frac{y-y_a}{h_a}, t_w = \log\left(\frac{w}{w_a}\right), t_h = \log\left(\frac{h}{h_a}\right), \tag{2.5}$$

$$\mathrm{t_x^*} = \frac{x^*-x_a}{\mathrm{w_a}}, \mathrm{t_y^*} = \frac{y^*-y_a}{h_a}, t_w^* = \log\left(\frac{w^*}{w_a}\right), t_h^* = \log\left(\frac{h^*}{h_a}\right), \tag{2.6}$$

其中 x，y，w 和 h 表示框的中心点的横纵坐标以及长和宽。变量x，$\mathrm{x_a}$，$\mathrm{x^*}$分别为预测框，anchor，真实框的横坐标值（y，w，h 同理）。

**均方差损失（Mean Square Error，MSE/L2 loss）**：其公式如下所示：

$$\mathrm{L_2}(e) = |e|^2, \tag{2.7}$$

$$\frac{\partial \mathrm{L_2}(e)}{\partial e} = 2e. \tag{2.8}$$

L1 Loss 相比较于 L2 Loss 收敛速度慢，0 点处无梯度且其他位置梯度为常数，因此在极值点附近容易产生震荡，但其对离群点的处理上要好于 L2 Loss。

**平滑 L1 损失（Smooth L1 loss）**：R. Girshich 等人[8]结合了 L1 loss 和 L2 loss 各自的特点，设计出了 Smooth L1 loss，其公式如下：

$$\mathrm{Smooth\ L_1}(e) = \begin{cases} 0.5e^2, & if\ |e| < 1 \\ |e| - 0.5, & othjerwise' \end{cases} \tag{2.9}$$

Smooth L1 Loss 在 e 较大时，使用的 L1 Loss 的形式，以防止离群点的影响；在 e 较小时使用的 L2 Loss 的形式，以获得较小的梯度（收敛速度），防止在极值点附近震

荡。

**平衡 L1 损失（Balanced L1 loss）[85]**：J. Pang 等人认为，训练样本的 inliers（损失小于 0.1）对于模型训练有着更好的效果，而其只有 outliers（损失大于等于 0.1）产生的损失的 0.3 倍，希望能够在 Smooth L1 loss 的基础上放大 inliers 的梯度。其公式为：

$$\mathrm{L_b}(e) = \begin{cases} \frac{\alpha}{b}(b|e|+1)\ln(b|e|+1) - \alpha|x|, & if\ |x| < 1 \\ \gamma|x| + C, & otherwise \end{cases}, \tag{2.10}$$

$$\frac{\partial \mathrm{L_b}}{\partial e} = \begin{cases} \alpha(b|e|+1), & if\ |x| < 1 \\ \gamma, & otherwise \end{cases}, \tag{2.11}$$

其中$\alpha, b, \gamma$为超参数，且满足$\alpha\ln(b+1) = \gamma$。

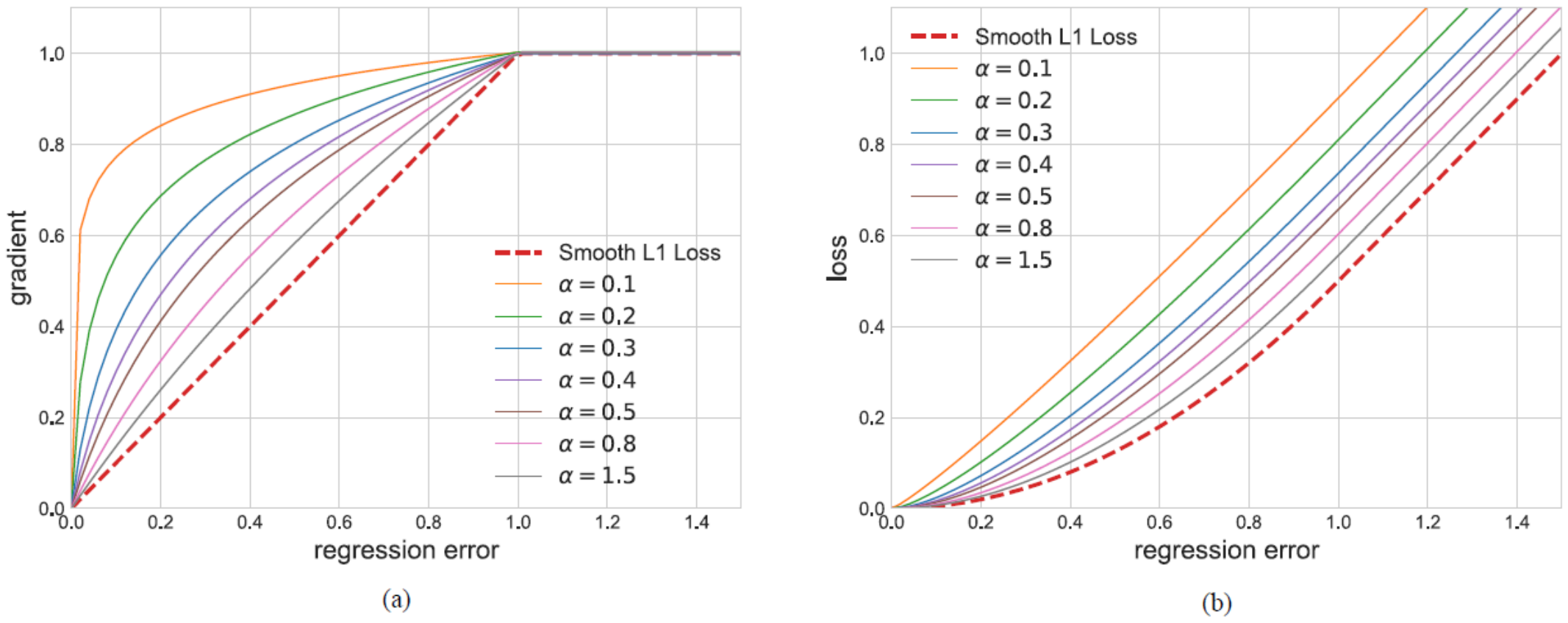

图 2.3 平衡 L1 损失和平滑 L1 损失的函数曲线对比[85]

Figure 2.3 The function curve of balanced L1 loss and Smooth L1 loss

**交并比损失（IoU Loss）[86]**：以上 4 种用于预测框回归的损失要求独立计算两个点的差异值，然后进行相加得到最终的 loss，其前提假设是两个点的横纵坐标（4 个值）是相互独立的，而实际上，这四个值是有一定相关性的。另一方面，实际预测框的评价指标使用的是 IoU，这与以上几种损失都是不等价的，存在损失相同但是 IoU 差异很大的情况，为了解决这一问题，J. Yu 等人提出直接使用 IoU 作为损失函数，使得损失和最终评价指标等同，公式如下：

$$\mathrm{L_{IoU}} = 1 - \frac{Intersection(\boldsymbol{b},\boldsymbol{b}^*)}{Union(\boldsymbol{b},\boldsymbol{b}^*)}, \tag{2.12}$$

其中$\boldsymbol{b}$和$\boldsymbol{b}^*$表示预测框及其对应的真实框。交并比损失的缺点在于，它的收敛速度较慢，而且没办法处理两框没有交集的情况。

**广义交并比损失（Generalized IoU loss，GIoU loss）[87]**：IoU loss 在预测框和真实框没有交集时损失恒为 1，梯度恒为 0。为了解决这一缺点，H. Rezatofighi 等人提出 GIoU loss，其表达式为：

$$\mathrm{L_{GIoU}}(A,B)=1-IoU(A,B)+\frac{|C-A\cup B|}{|C|},\tag{2.13}$$

其中，$A,B\subseteq\mathbb{R}^n$ 是两个框，C 是能包围住 A 和 B 的最小凸包。

**完整交并比损失（Complete IoU loss，CIoU loss）[88]**：CIoU loss 考虑了三个重要的几何指标：重叠面积，中心点距离，长宽比。给定一个预测框$\boldsymbol{b}$和一个目标框$\boldsymbol{b}^*$，CIoU loss 可以表示为：

$$\mathrm{L_{CIoU}}=1-IoU(\boldsymbol{b},\boldsymbol{b}^*)+\frac{\rho^2(\boldsymbol{b},\boldsymbol{b}^*)}{c^2}+\alpha v,\tag{2.14}$$

$$v=\frac{4}{\pi^2}\left(\arctan\frac{w^*}{h^*}-\arctan\frac{w}{h}\right)^2,\ \alpha=\frac{v}{1-IoU}+v,\tag{2.15}$$

其中$\rho(\boldsymbol{b},\boldsymbol{b}^*)$为预测框和目标框的中心点欧式距离，CIoU loss 相比于 IoU loss 收敛速度和检测精度上都有了较大的提升。

**专注高效交并比损失（Focal Efficient IoU loss，F-EIoU）[89]**：Y. Zhang 等人重新回归了 CIoU loss，提出了一种更为高效的 IoU loss，称其为 EIoU loss：

$$\begin{aligned}\mathrm{L_{EIoU}}&=\mathrm{L_{IoU}}+L_{dis}+L_{asp}\\&=1-\mathrm{IoU}+\frac{\rho^2(\boldsymbol{b},\boldsymbol{b}^{gt})}{c^2}+\frac{\rho^2(w,w^{gt})}{C_w^2}+\frac{\rho^2(h,h^{gt})}{C_h^2},\end{aligned}\tag{2.16}$$

其中，$\mathrm{C_w}$和$\mathrm{C_h}$是两个框的最小包围框的长和宽。从公式可得，EIoU loss 分为三块：交并比损失$\mathrm{L_{IoU}}$、距离损失$\mathrm{L_{dis}}$和外观损失$\mathrm{L_{asp}}$。从这样看，EIoU loss 考虑的点和 CIoU loss 是一致的，因此也保留了 CIoU 所有的特点。为了解决上文提到的抑制 outliers 的问题，作者进一步提出了 F-EIoU loss：

$$\mathrm{L_{F-EIoU}}=IoU^{\gamma}L_{EIoU},\tag{2.17}$$

其中，其中$\gamma$是用于控制异常值的抑制参数。

本节提出了一个**快速交并比损失（Fast IoU loss，FIoU loss）**，其公式可表达为：

$$\mathrm{L_{FIoU}}=IoU(\boldsymbol{b},\boldsymbol{b}^*)^{\eta}\left(1-IoU(\boldsymbol{b},\boldsymbol{b}^*)+\sum\nolimits_{i\in\{x,y,w,h\}}\left|\left|t_i-t_i^*\right|\right|_2^2\right),\tag{2.18}$$

其中$\eta$是用于控制异常值的抑制参数。FIoU loss 希望加入一个 L2 loss（FIoU loss 的最后一项）去弥补 IoU loss 收敛速度慢的问题。然而正如上文所提及的，L2 loss 非常容易受到 outliers 的影响，对 outliers 的回归会导致回归精度的下降。因此希望加入一项抑制项，即$IoU(\boldsymbol{b},\boldsymbol{b}^*)^{\eta}$，去减轻 outliers 带来的问题，那些有着高回归损失的低质量样本会被过滤掉，而模型会关注有着中等回归损失的主要样本。因此，FIoU loss 维持了在保持对 outliers 高鲁棒性的同时，实现了快速收敛。

对于 IoU 预测任务，本方法使用交叉熵作为损失函数。最终的目标先验概率为目标

似然概率和 IoU 预测值的相乘的平方根，即：

$$\text{prior} = \sqrt{IoU_{pred} * objectness}, \tag{2.19}$$

在水下环境中，模糊目标频繁出现，加入 IoU 预测项可以引入一部分不确定性到目标先验概率之中，增强了模型的检测能力和鲁棒性。

## 2.2 概率型二阶段检测器

目标检测器旨在对每一个目标 i 输出预测框的四个坐标$b_i \in \mathbb{R}^4$以及类似然分数$s_i \in \mathbb{R}^{|C|}$。一阶段和二阶段检测器的主要区别在于它们表示类似然概率的方式。

### 2.2.1 一阶段检测器的贝叶斯解释

一阶段检测器同时预测目标的位置和其类似然分数。令$L_{i,c} = 1$表示一个候选框 i 分类为 c 为正样本，$L_{i,c} = 0$表示为背景。大多数的一阶段检测器使用 Sigmoid 函数将每一个类似然分数变成独立的伯努利分布，即：

$$s_i(c) = P\left(L_{i,c} = 1\right) = \sigma\left(w_c^T \vec{f_i}\right), \tag{2.20}$$

其中，$f_i \in \mathbb{R}^C$是骨干网络得到的特征，$w_c$是特定类的权重向量，σ是 sigmoid 函数。训练过程中，一阶段检测器会最大化类对数似然概率$\log\left(P\left(L_{i,c}\right)\right)$。一阶段检测器之间的不同在于如何去定义正负样本，但所有的一阶段检测器都会优化对数似然概率。

### 2.2.1 二阶段检测器的贝叶斯解释

在第一个阶段，RPN 会提取可能的目标的位置，然后给出目标似然$P(O_i)$，RPN 通过一个对数似然优化的二分类任务将 Anchor 分类为前景和背景，最高分数的前景预测将会成为目标提议（Object Proposals），但 RPN 做决策时会相对保守，而这会提高召回率，且每一个提议框的目标似然都非常高。在第二个阶段，一个 softmax 分类器会将每一个提议框进行多分类，它们从这些目标提议中提取特征，将他们分类为 C 个类别或者背景，这个分类器仍然使用对数似然作为训练目标。然而在训练过程中，这一类的分布实际上已经隐式地条件于一阶段的正样本检测，即$P(C_i|O_i = 1)$。在最终的输出中，二阶段检测器的类别分数为二阶段多分类器的结果，即：

$$s_i(c) = P(C_i|O_i = 1). \tag{2.21}$$

### 2.2.1 概率型二阶段检测器的贝叶斯解释

在常见的二阶段检测器中仅使用第二个阶段的条件概率进行最终的预测，而本文认

为这并非是完备的推断，其忽视的第一个阶段的先验概率。在概率型二阶段检测器中，将使用完备的边缘分布进行概率预测，即：

$$\mathrm{s_k}(c) = P(C_k = c),\ c \in \mathcal{C} \cup \{bg\}, \tag{2.22}$$

其中$\mathrm{s_k}(c)$为提议 k 类别 c 上的最终检测分数，$\mathcal{C}$为类别集，bg 为背景类别。如果令$\mathrm{P(C_k = bg|O_k = 0) = 1}$,边缘分布$\mathrm{P(C_k)}$可以表示为：

$$\begin{aligned}\mathrm{P(C_k} = c) &= \textstyle\sum_\mu P(C_k = c|O_k)P(O_k = \mu) \\ &= \mathrm{P(C_k} = c|O_k = 1)P(O_k = 1) + \mathrm{P(C_k} = c|O_k = 0)P(O_k = 0) \\ &= \mathrm{P(C_k} = c|O_k = 1)P(O_k = 1).\end{aligned} \tag{2.23}$$

在实现上，最终的检测分数将是 RetinaRPN 输出的先验概率和 R-CNN 的类别分数的结合，即：

$$\mathrm{s_k}(c) = \sqrt{prior(k) * cls_k(c)}, \tag{2.24}$$

其中$\mathrm{cls_k}(c)$为类别分数。使用概率型二阶段，检测器可以将一阶段的先验不确定性加入到最终检测结果上，在面对水下环境的低对比度、遮挡、生物拟态等问题时，这种不确定性的建模有利于模型更好地得到合理的检测结果。

## 2.3 提升再权重模块

前一小节提及的概率型二阶段检测器存在一个显著的问题。在原来的二阶段检测器中，第二个阶段会独立于第一个阶段做预测。因此，对于一些高质量（和真实框重合度高）的提议框，只要它被选中为提议框，即使其没有获得足够高的先验概率分数，也不会影响最终的检测。然而，在概率型二阶段检测器中，当 RPN 错误地生成了一个带有低先验概率的提议框，它是非常难被重新视作为一个高置信度的预测的，因为最终的检测分数是一阶段先验概率和二阶段条件概率的相乘。在水下环境中，模糊目标常常作为困难样本出现，RPN 的初次判断容易出现失误，从而影响最终的检测结果。

为了解决这一个问题，希望当 RPN 错误地估计了提议框的先验概率时，R-CNN（第二个阶段）可以将这一个错误修正回来。因此，本节提出了一个软采样策略提升再权重（以下缩写 BR），其设计受到 boosting 算法思想的影响，并且可以很好地适应现有的二阶段检测框架。普通的 Faster R-CNN 会将所有提议框的分类权重都设置为 1，不同于此，BR 倾向于赋予困难样本更大的权重，这里的困难样本定义为先验概率被错误估计的提议框。对于一个样本 k，其分类权重为：

$$\mathrm{w_k} = \begin{cases}(1 - prior(k))^\gamma,\ k \in \mathcal{F} \\ prior(k)^\gamma,\ k \in \mathcal{B}\end{cases}, \tag{2.25}$$

其中，$\gamma \geq 0$为专注参数，$\mathcal{F}$表示前景样本的集合，$\mathcal{B}$表示背景样本的集合。当检测器遇到了困难的正/负样本时，其先验概率会非常小/大。因此，权重项会增加，损失值也会相应放大。

BR 可被视作一种困难样本挖掘。有几种相似的困难样本挖掘的工作：在线困难样本挖掘（Online Hard Example Mining，OHEM）[90]和 Focal loss。

OHEM 是自举法（bootstraping）的一种，其最初是为 Fast R-CNN 设计的。它先为所有的 RoI 进行一次前向传播并进行非极大值抑制操作（Non-Maximum Suppression），计算每一个 RoI 的分类损失，并分别采样正样本和负样本中损失最大的样本，然后进行第二次前向传播和反向传播完成一次训练。OHEM 的缺点在于其需要两次前向传播，计算量和储存消耗都比较大。

Focal loss 在前文已经提及，其为 RetinaNet 模型设计，旨在解决前景和背景的极度不平衡，实际是困难样本挖掘的一种方式。通过模型自身输出的置信度与真实值的差异调节权重大小，当前输出背离真实值时（困难样本），便放大此样本损失权重；当前输出接近真实值时（简单样本），便缩小此样本的损失权重。Focal loss 极少被用于二阶段模型，尤其是第二个阶段的 R-CNN 模块，原因是 RPN 的提议框经过 NMS 之后再经过一轮采样，正负样本的问题就已经缓解了，这和 Focal loss 的机制本身存在重合。

表 2.1 困难样本挖掘之间的异同

Table 2.1 The differences between hard example mining methods

| | OHEM | Focal loss | BR |
|---|---|---|---|
| 困难样本定义 | R-CNN 上损失大的样本 | R-CNN 上似然分数被错误估计的样本 | RPN 上先验概率被错误估计的样本 |
| 采样方式 | 选取最困难的样本进行采样 | 随机采样 | 随机采样 |
| 正样本权重 | 1 | $(1-p_i)^\gamma$ | $(1-prior(k))^\gamma$ |
| 前向传播次数 | 2 | 1 | 1 |

本节提出的 BR 与前两者既有相似之处，也有不同之处（如表 2.1 所示）。在困难样本定义上，OHEM 和 Focal loss 都将 R-CNN 误判程度高的样本作为困难样本，而 BR 将 RPN 犯错程度高的样本作为困难样本，因此相比于 OHEM，BR 无需进行两次前向传播，大大减少了计算量。其次在采样方式上，OHEM 只选取最困难的样本进行采样，而 Focal loss 和 BR 都是维持原本的随机采样。OHEM 对所有的采样后的样本一视同仁，Focal loss 和 BR 会倾向于将更大的权重分配到更难的样本上。后续的实验会揭示

BR 在概率型二阶段检测器上的兼容性。

## 2.4　实验与讨论

### 2.4.1　数据集介绍

本节对本章提出的 Boosting R-CNN 的性能进行测试，并测试本章提出的各个模块的性能。本节采用了两个水下目标检测数据集：UTDAC2020 和 Brackish，以及两个通用目标检测数据集 Pascal VOC 和 COCO。以下是对这三个数据集的介绍：

**UTDAC2020 数据集**[91]是最新的水下数据集（如图 2.4），其来自于 2020 年水下目标检测算法竞赛（Underwater Target Detection Algorithm Competition 2020）。数据集有 4 个类别：海胆（Echinus）、海参（Holothurian）、海星（Starfish）和扇贝（Scallop）。训练集中有 6461 张图像，但测试集的标注并不公开，因此手动将原来的训练集分割为 5168 张图片的训练数据和 1293 张图片的验证数据。数据的图片有四种分辨率：3840×2160、1920×1028、720×405 和 586×480。

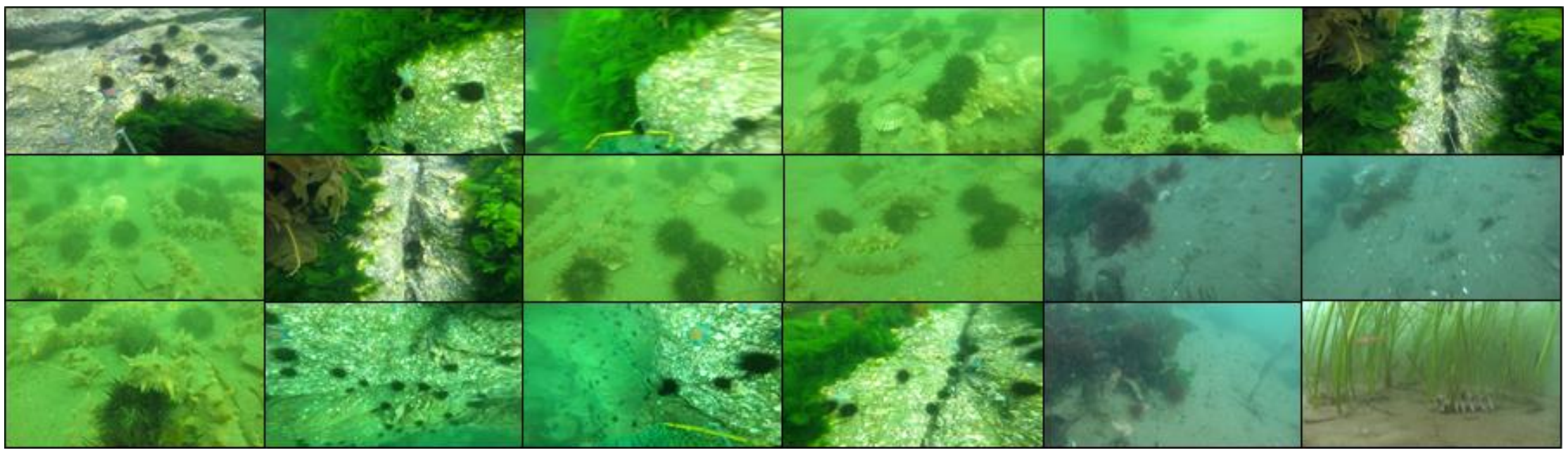

图 2.4 UTDAC2020 数据集的介绍
Figure 2.4 The overview of UTDAC2020 dataset

**Brackish 数据集**[92]是一个采集于温带咸水域的数据集（如图 2.5），有 6 个类别：大鱼（Big fish）、螃蟹（Crab）、水母（Jellyfish）、海虾（Shrimp）、小鱼（Small fish）和海星（Starfish）。其有 14518 张图像和 25613 个标注，训练集、验证集、测试集的分别为 9967、1467 和 1468 张图像。数据集图像的分辨率为 960×540。

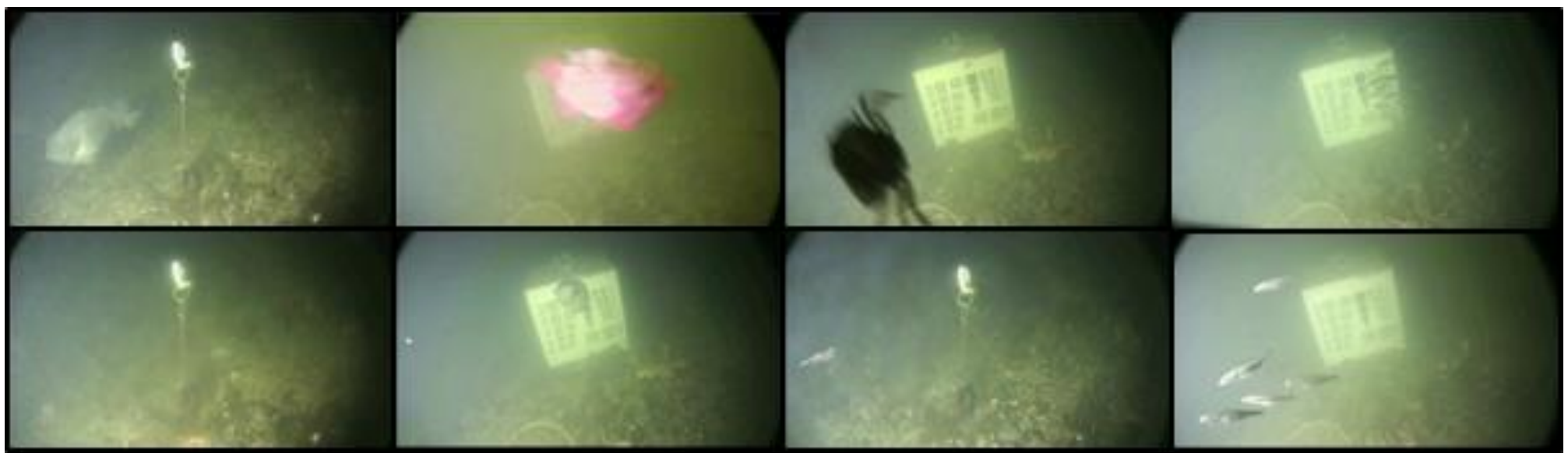

图 2.5 Brackish 数据集的介绍
Figure 2.5 The overview of Brackish dataset

**Pascal VOC 数据集**[93]是一个通用目标检测数据集（如图 2.6），其数据集有 20 个常用的类别。Pascal VOC2007 包含 5000 张训练验证集图像和 5000 张测试图像。Pascal VOC2012 包含 11540 张训练验证集图像。本方法将在 Pascal VOC2007 和 Pascal VOC2012 的训练验证集的并集中训练，在 Pascal VOC2007 的测试集上进行测试。数据集的图像分辨率为 1000×600。

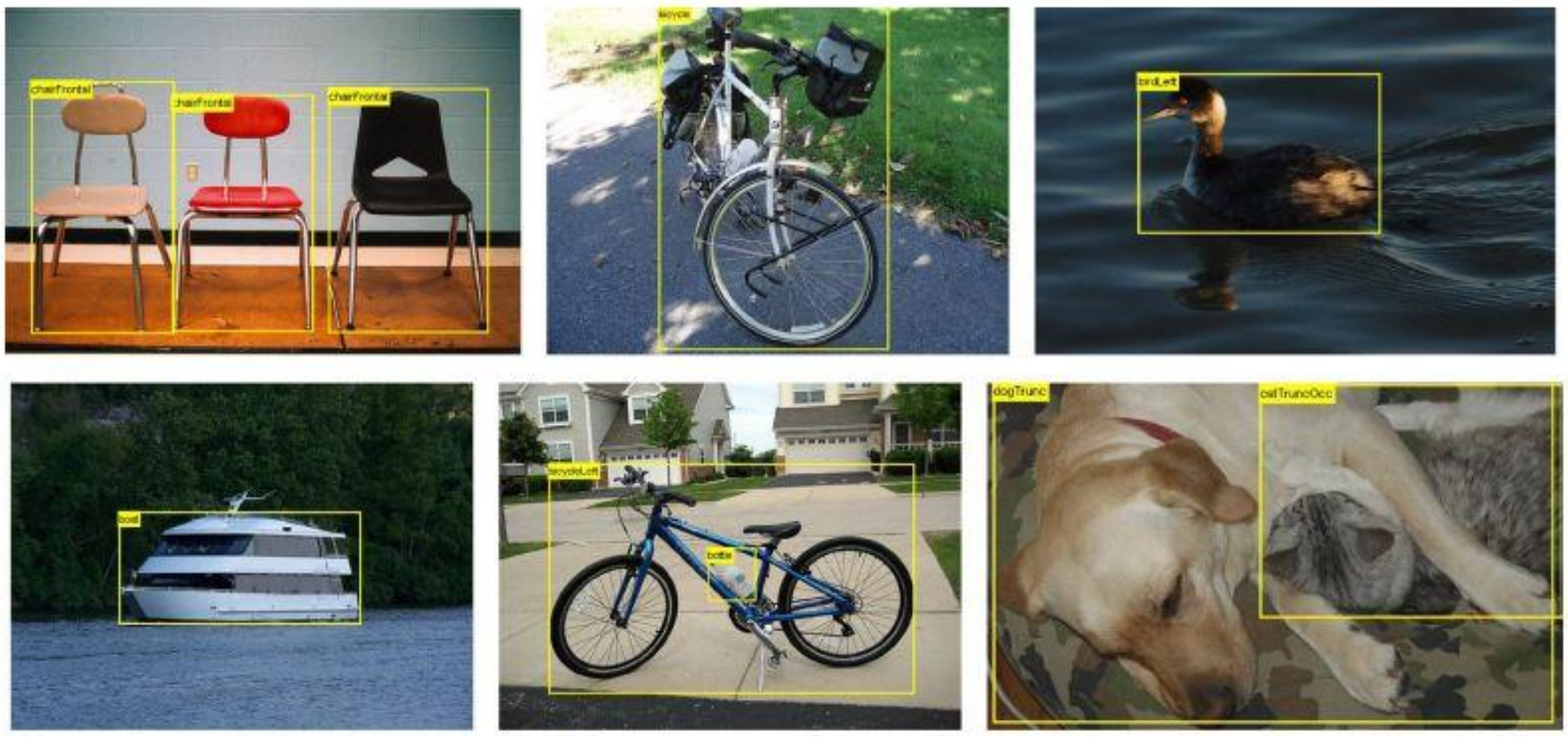

图 2.6 Pascal Voc 数据集的介绍
Figure 2.6 The overview of Pascal Voc dataset

**COCO 数据集**[94]是一个通用的目标检测数据集（如图 2.7），包含 80 个常用的检测类别。在 COCO2017 年数据集中有 16 万张图像，其中 12 万张图像有标签。在目标检测中，常用 trainval 数据集（123287 张图像）进行训练，在 val 数据集（5000 张图像）上验证，在最终的测试集 test-dev 上进行测试。

### 2.4.2 实验设置

本方法将在 PyTorch 1.9 上运行，并基于 MMDetection 平台实现。模型将在一张 NVIDIA GTX 1080Ti GPU 上训练 12 个 epochs，批大小（Batch Size）为 4。优化器为 SGD，初始学习率为 0.005，并在第 9 和第 12 个 epoch 时都缩放 0.1 倍。权重衰减设为 0.0001，动量设为 0.9。在推断过程中，只使用最多 256 个提议框输送到第二个阶段，这将大大提升模型推断速度。在实验中，不使用除了水平翻转以外的数据增强。

### 2.4.3 评估指标

本节实验采用两种评估方式：Pascal Voc2012 数据集的 mAP50 评测方式和 COCO 的 AP[0.5:0.05:0.95]评测方式。以下详细阐述评测指标的计算方法。

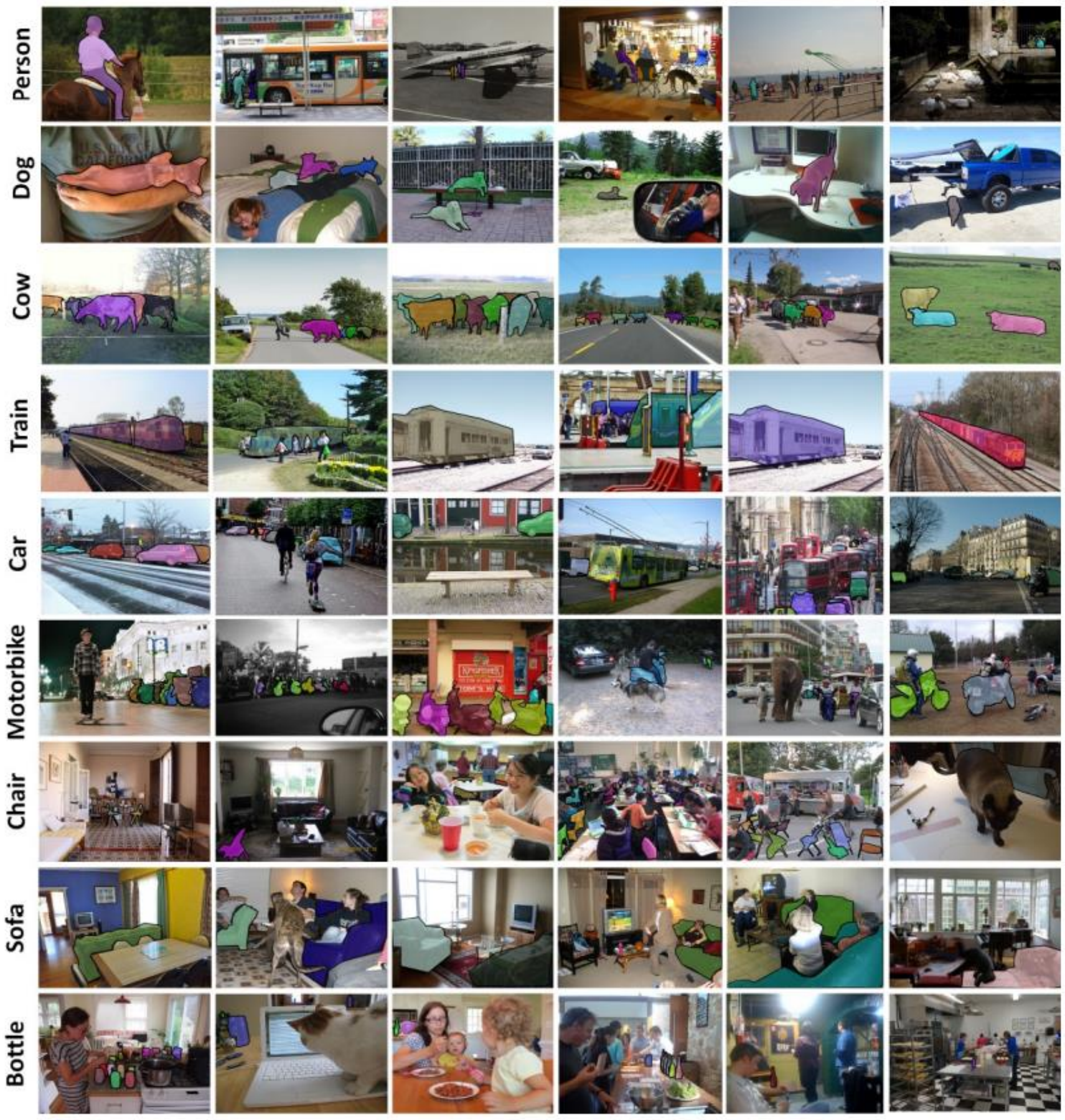

图 2.7 COCO 数据集的介绍[94]
Figure 2.7 The overview of COCO dataset

### 2.4.3.1 交并比

交并比（Intersection over Union，IoU）指的是检测框和 Ground Truth 的交集和并集之间的比，当 IoU=1 时，检测框完全等于 Ground Truth，当 IoU=0 时，检测框和 Ground Truth 没有重叠。

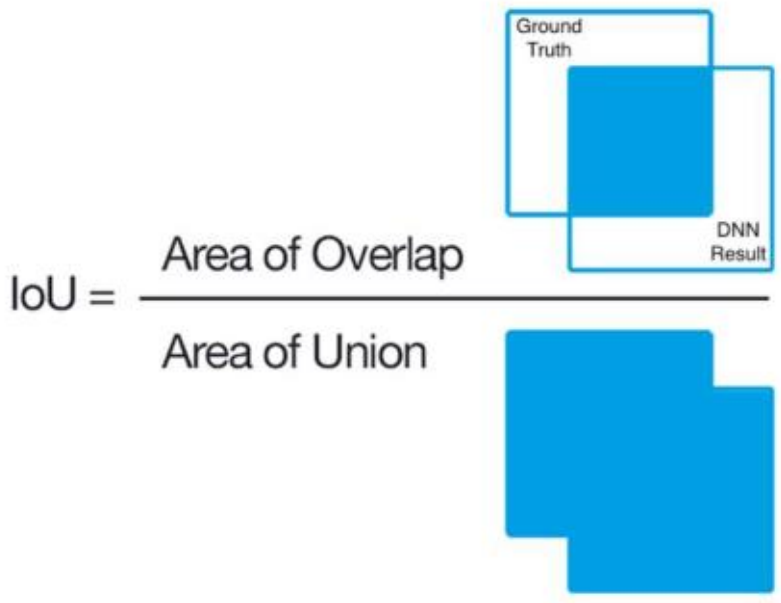

图 2.8 IoU 的计算方法
Figure 2.8 The calculation of IoU

#### 2.4.3.2 准确率和召回率

当检测框和有着最大 IoU 的 Ground Truth 的 IoU 超过一定阈值，且分类正确时，当前检测框便是一个真正样本（True Positive，TP）。当检测框将一个背景判断为前景时，当前检测框便是一个假正样本（False Positive，FP）当检测框和任何一个 Ground Truth 的 IoU 都没有超过阈值，或者检测框和 Ground Truth 的 IoU 超过阈值，但是分类错误，当前检测框便是一个假负样本（False Positive，FN）。

准确率的计算公式为：

$$\mathrm{P} = \frac{\mathrm{TP}}{TP+FP}. \tag{2.26}$$

召回率的计算公式为：

$$\mathrm{R} = \frac{\mathrm{TP}}{TP+FN}. \tag{2.27}$$

#### 2.4.3.3 平均精度均值

把图像输入到深度学习模型之中，会得到一系列的检测框，及其分类结果和置信度。根据置信度进行排序，设置置信度阈值，每一个置信度阈值能得到一组准确率和召回率。当将置信度阈值从 1 到 0 逐渐减少，每把一个检测框的当作正样本时，记下当前准确率和召回率。将每一组准确率和召回率相乘，求均值，则得到平均精度（Average Precision，AP）。每一个检测类别都会得到一个 AP，而平均精度均值 mAP（Mean Average Precision，mAP）的计算公式为：

$$\mathrm{mAP} = \frac{1}{\mathrm{N_c}}\sum_{\mathrm{i}}^{\mathrm{N_c}} AP_i, \tag{2.28}$$

其中，Nc 为数据集的类别数。mAP@50 便是将计算正样本时的 IoU 阈值设置为 0.5，mAP@[0.5:0.05:0.95]（以下将缩写为 AP）是将 IoU 阈值从 0.5 开始每次增加 0.05 一直增加到 0.95，对每一个 IoU 阈值都计算一次 mAP，然后计算均值的结果。

### 2.4.4 实验结果和分析

#### 2.4.4.1 在 UTDAC2020 数据集上的实验结果

本节将 Boosting R-CNN 和主流的目标检测方法进行对比，其中包括二阶段检测器：Faster R-CNN[9]、OHEM[90]、Cascade R-CNN[25]、Libra R-CNN[85]、Cascade RPN[95]、PAFPN[15]、Double-Head[96]、Dynamic R-CNN[29]、FPG[97]、GRoIE[98]、SABL[99]、PISA[100]、DetectoRS[101]、CenterNet2[102]，以及一阶段检测器：SSD[11]、RetinaNet[13]、FSAF[103]、FCOS[23]、RepPoints[24]、FreeAnchor[26]、NASFPN[17]、ATSS[27]、PAA[28]、AutoAssign[30]、

GFL[104]、VFNet[105]。另外，基于 Transformer 的检测器如 Deformable DETR 也被纳入了比较之中。CenterNet2*和 Boosting R-CNN*表示模型使用了多尺度训练和三倍训练时间。除了 CenterNet2，所有的对比方法都由 MMDetection 实现，CenterNet2 的官方实现平台为 Detectron2。

如表 2.2 所示，在单精度训练的设置下，Boosting R-CNN 能够达到 48.2%AP，比 DetectoRS（47.6%AP）、PAA（47.5%AP）和 CenterNet2（47.2%AP）都要高。在多尺度训练的设置下，Boosting R-CNN 仍然超越 CenterNet2（50.3%AP vs 48.9%AP）。因此，本方法能够击败所有的目标检测器，并在 UTDAC2020 建立新的 SOTA。至于推断速度，Boosting R-CNN 实现了 13.5FPS，此速度比大多数的一阶段和二阶段检测器都要高。图 2.9 是检测可视化结果，（a）为真实值，（b）为 DetectoRS 的检测结果，（c）为 Boosting R-CNN 的检测结果，粉红色的圈表示的是 DetectoRS 失效的部分。从图中可看出 Boosting R-CNN 相比于 DetectoRS 能够检测出更多的模糊目标，而其归功于 BR 的困难样本挖掘机制。

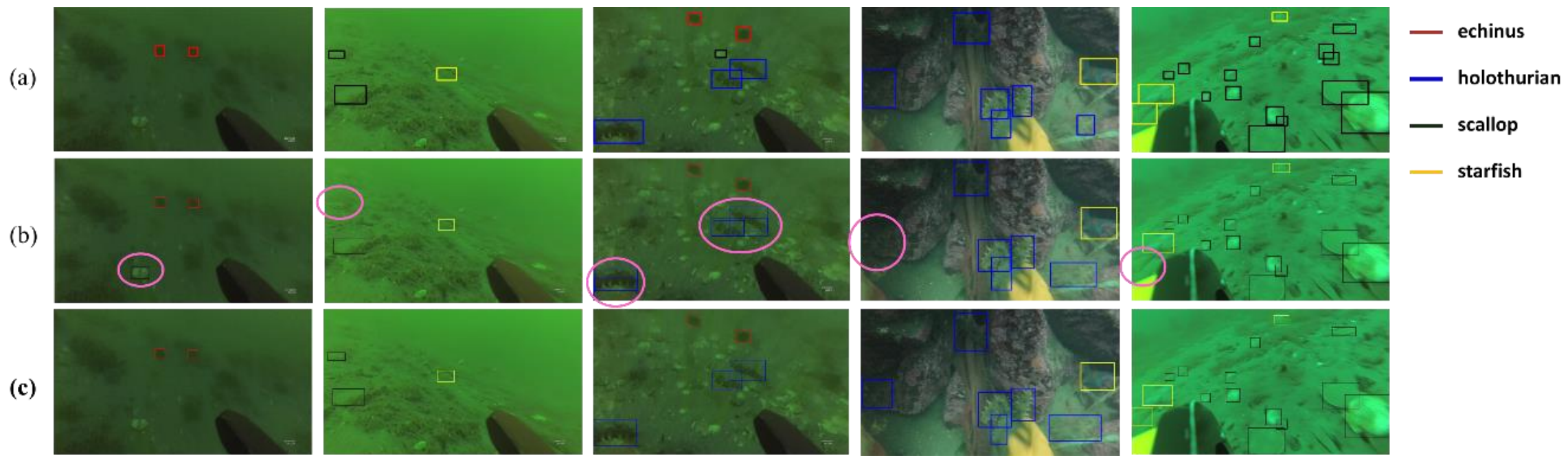

图 2.9　在 UTDAC2020 数据集上的可视化结果

Figure 2.9 The visualization of detection results in UTDAC2020

此外，在所有的一阶段检测器中，可以发现在水下目标检测任务中，基于 Anchor 的方法（SSD、RetinaNet、ATSS、FreeAnchor、PAA、GFL）相比于无 Anchor 的方法（FSAF、FCOS、RepPoints、VFNet、AutoAssign）要相对好一点。这是因为在低对比度且失真水下环境中，水下生物的边缘会很模糊，Anchor 可以辅助模型获得目标边缘的先验位置，加速了收敛速度。

表 2.2 在 UTDAC2020 数据集上的实验结果

Table 2.2 Comparisons with other object detection methods on UTDAC2020 dataset

| 方法 | 骨干网络 | AP | AP50 | AP75 | APS | APM | APL | FPS |
|---|---|---|---|---|---|---|---|---|
| Faster R-CNN | R50+FPN | 44.5 | 80.9 | 44.1 | 20.0 | 39.0 | 50.8 | 11.6 |
| Faster R-CNN +OHEM | R50+FPN | 45.1 | 82.0 | 45.1 | 21.6 | 39.1 | 51.4 | 11.6 |
| Cascade R-CNN | R50+FPN | 46.6 | 81.5 | 49.3 | 21.0 | 40.9 | 53.3 | 8.8 |
| Libra R-CNN | R50+FPN+BFP | 45.8 | 82.0 | 46.4 | 20.1 | 40.2 | 52.3 | 11.0 |
| Cascade RPN | R50+FPN | 46.5 | 79.5 | 41.2 | 20.4 | 38.6 | 47.7 | 8.3 |
| PAFPN | R50+PAFPN | 45.5 | 82.1 | 45.9 | 18.8 | 39.7 | 51.9 | 10.9 |
| Double-Head | R50+FPN | 45.3 | 81.5 | 45.7 | 20.2 | 40.0 | 51.4 | 5.7 |
| Dynamic R-CNN | R50+FPN | 45.6 | 80.1 | 47.3 | 19.0 | 39.7 | 52.1 | 12.1 |
| FPG | R50+FPG | 45.4 | 81.6 | 46.0 | 19.8 | 39.7 | 51.4 | 13.1 |
| GRoIE | R50+FPN | 45.7 | 82.4 | 45.6 | 19.9 | 40.1 | 52.0 | 6.0 |
| SABL | R50+FPN | 46.6 | 81.6 | 48.2 | 19.6 | 40.4 | 53.4 | 9.9 |
| PISA | R50+FPN | 46.3 | 82.1 | 47.4 | 20.8 | 40.8 | 52.6 | 10.3 |
| DetectoRS | R50+FPN | 47.6 | 82.8 | 49.9 | 23.1 | 41.8 | 54.2 | 4.0 |
| CenterNet2 | R50+FPN | 47.2 | 81.6 | 49.8 | 18.2 | 41.3 | 53.4 | 14.2 |
| CenterNet2* | R50+FPN | 48.9 | 83.0 | 52.6 | 21.7 | 43.5 | 55.2 | 14.2 |
| SSD512 | VGG16 | 40.0 | 77.5 | 36.5 | 14.7 | 36.1 | 45.1 | 25.0 |
| RetinaNet | R50+FPN | 43.9 | 80.4 | 42.9 | 18.1 | 38.2 | 50.1 | 11.4 |
| FSAF | R50+FPN | 43.9 | 81.0 | 42.9 | 18.5 | 38.9 | 50.9 | 12.8 |
| FCOS | R50+FPN | 43.9 | 81.1 | 43.0 | 19.9 | 38.2 | 50.4 | 12.7 |
| RepPoints | R50+FPN | 44.0 | 80.5 | 43.0 | 18.7 | 38.5 | 50.3 | 11.1 |
| FreeAnchor | R50+FPN | 46.3 | 82.3 | 46.9 | 21.0 | 40.5 | 52.6 | 11.4 |
| RetinaNet+NASFPN | R50+NASFPN | 37.4 | 70.3 | 35.8 | 12.4 | 36.4 | 40.4 | 13.8 |
| ATSS | R50+FPN | 46.2 | 82.5 | 46.9 | 19.7 | 41.4 | 52.4 | 11.8 |
| PAA | R50+FPN | 47.5 | 83.1 | 49.7 | 19.5 | 42.4 | 53.6 | 6.6 |
| AutoAssign | R50+FPN | 46.3 | 83.0 | 47.6 | 18.0 | 41.3 | 52.2 | 12.3 |
| GFL | R50+FPN | 46.4 | 81.9 | 47.8 | 19.3 | 40.9 | 52.5 | 12.7 |
| VFNet | R50+FPN | 44.0 | 79.3 | 44.1 | 18.8 | 38.1 | 50.4 | 10.5 |
| Deformable DETR | R50 | 46.6 | 84.1 | 47.0 | 24.1 | 42.4 | 51.9 | 7.6 |
| **Boosting R-CNN** | R50+FPN | 48.2 | 82.7 | 51.3 | 20.2 | 41.8 | 54.8 | 13.5 |
| **Boosting R-CNN*** | R50+FPN | **50.3** | **84.7** | **54.8** | **25.1** | **44.6** | **56.5** | **13.5** |

表 2.3 在 Brackish 数据集上的实验结果

Table 2.3 Comparisons with other object detection methods on Brackish dataset

| 方法 | AP | AP50 |
| --- | --- | --- |
| Baseline | 38.9 | 83.7 |
| Faster R-CNN w/ FPN | 79.3 | **97.4** |
| Cascade R-CNN w/ FPN | 80.7 | 96.9 |
| RetinaNet w/ FPN | 78.0 | 96.5 |
| DetectoRS | 81.6 | 97.0 |
| CenterNet2 | 79.3 | **97.4** |
| Boosting R-CNN | **82.0** | **97.4** |

表 2.4 在 Pascal VOC 数据集上的实验结果

Table 2.4 Comparisons with other object detection methods on Pascal VOC dataset

| 方法 | 骨干网络 | 输入尺寸 | mAP |
| --- | --- | --- | --- |
| Faster R-CNN | VGG16 | 1000×600 | 73.2 |
| Faster R-CNN | ResNet101 | 1000×600 | 76.4 |
| Faster R-CNN | ResNet50+FPN | 1000×600 | 79.5 |
| MR-CNN | VGG16 | 1000×600 | 78.2 |
| R-FCN | ResNet101 | 1000×600 | 80.5 |
| RON384++ | VGG16 | 384×384 | 77.6 |
| Cascade R-CNN | ResNet50+FPN | 1000×600 | 80.0 |
| CenterNet2 | ResNet50+FPN | 1000×600 | 76.8 |
| SSD300 | VGG16 | 300×300 | 74.3 |
| SSD512 | VGG16 | 512×512 | 76.8 |
| YOLO | GoogleNet | 448×448 | 63.4 |
| YOLOv2 | DarkNet-19 | 544×544 | 78.6 |
| RefineDet320 | VGG16 | 320×320 | 80.0 |
| RetinaNet | ResNet50+FPN | 1000×600 | 77.3 |
| DSSD321 | ResNet101 | 321×321 | 78.6 |
| DSSD513 | ResNet101 | 513×513 | 81.5 |
| FERNet | VGG16+ResNet50 | 512×512 | 81.0 |
| **Boosting R-CNN** | ResNet50+FPN | 1000×600 | **81.6** |

#### 2.4.4.2 在 Brackish 数据集上的实验结果

本小节将在 Brackish 数据集上对 Boosting R-CNN 进行测试。对比的方法包括 Faster R-CNN[9]、Cascade R-CNN[25]、RetinaNet[13]、DetectoRS[101]、CenterNet2[102]。结果如表 2.3 所示。Baseline 指的是 Brackish 数据集原论文中给出的实验结果。Boosting R-CNN 能够达到 82.0%AP 和 97.4%AP50，在所有对比的目标检测方法上实现了最高的性能，比 Cascade R-CNN 高 1.3%AP 和 0.5%AP50，比 DetectoRS 高 0.4%AP 和 0.4%AP50，比 CenterNet2 高 2.7%AP。图 2.10 为 Boosting R-CNN 在 Brackish 数据集上的检测结果可视化。

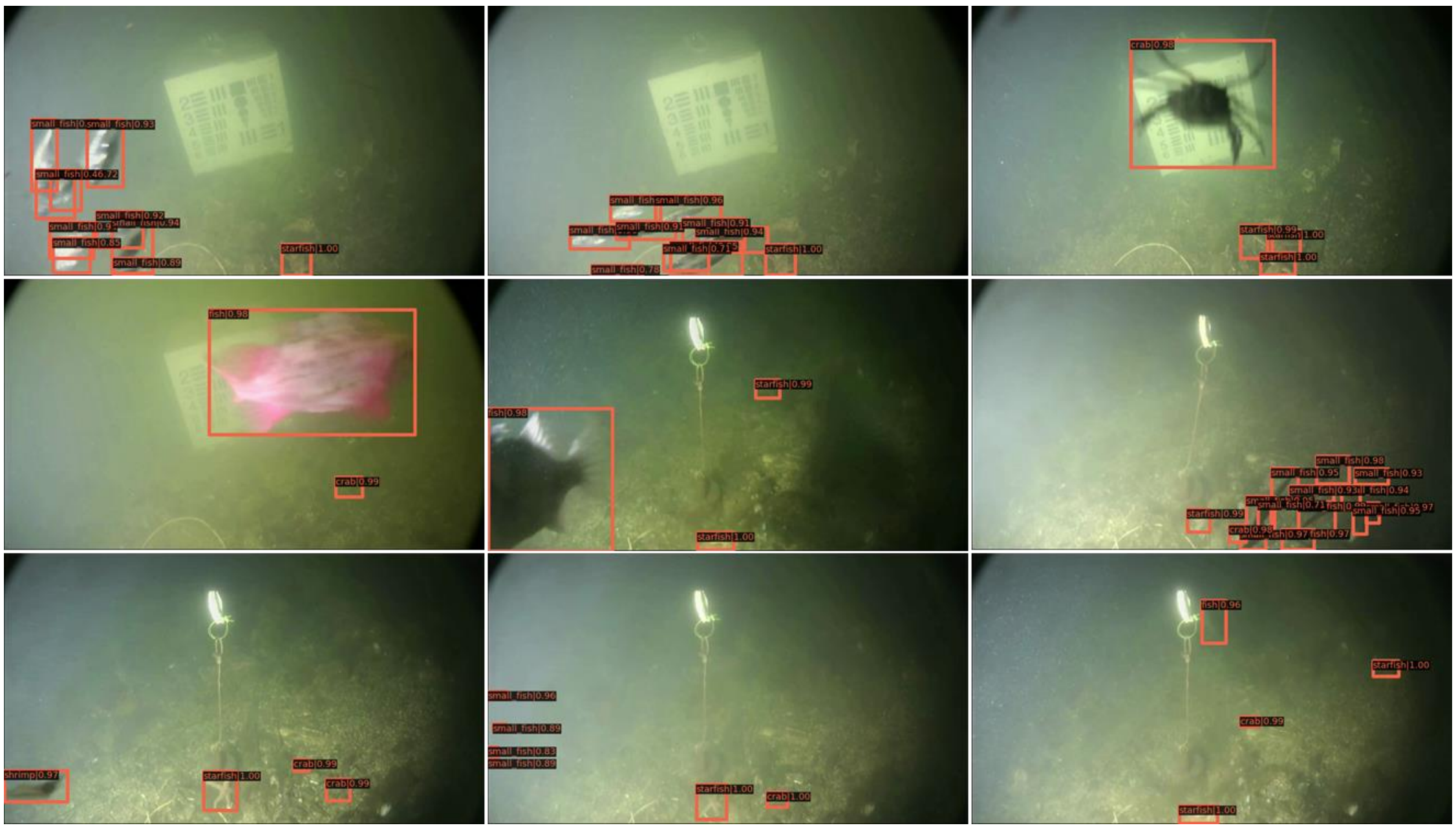

图 2.10 在 Brackish 数据集上的可视化结果

Figure 2.10 The visualization of detection results in Brackish dataset

#### 2.4.4.3 在 Pascal VOC 数据集上的实验结果

本小节将在主流的 Pascal VOC 数据集上对 Boosting R-CNN 进行测试。对比的方法包括二阶段检测器 Faster R-CNN[9]、MR-CNN[106]、R-FCN[107]、RON[108]、Cascade R-CNN[25]、CenterNet2[102]和一阶段检测器 SSD[11]、YOLO[10]、YOLOv2[19]、RefineDet[109]、RetineNet[13]、DSSD[110]、FERNet[35]。结果如表 2.4 所示。Boosting R-CNN 能够达到 81.6%mAP 的性能，比 Faster R-CNN（ResNet50+FPN）高 2.1%mAP，比 Cascade R-CNN 高 1.6%mAP，比 CenterNet2 高 4.8%mAP。

表 2.5 在 COCO 数据集上的实验结果
Table 2.5 Comparisons with other object detection methods on COCO dataset

| 方法 | 骨干网络 | AP | AP50 | AP75 | APS | APM | APL |
|---|---|---|---|---|---|---|---|
| Faster R-CNN | R101+FPN | 36.2 | 59.1 | 39.0 | 18.2 | 39.0 | 48.2 |
| Cascade R-CNN | R101 | 42.8 | 62.1 | 46.3 | 23.7 | 45.5 | 55.2 |
| Grid R-CNN | R101 | 41.5 | 60.9 | 44.5 | 23.3 | 44.9 | 53.1 |
| Libra R-CNN | X101-64x4d | 43.0 | 64.0 | 47.0 | 25.3 | 45.6 | 54.6 |
| Dynamic R-CNN | R101-DCN | 46.9 | 65.9 | 51.3 | 28.1 | 49.6 | 60.0 |
| BorderDet | X101-64x4d-DCN | 48.0 | 67.1 | 52.1 | 29.4 | 50.7 | 60.5 |
| Double-Head | R101 | 42.3 | 62.8 | 46.3 | 23.9 | 44.9 | 54.3 |
| TridentNet | R101-DCN | 46.8 | 67.6 | 51.5 | 28.0 | 51.2 | 60.5 |
| CPN | HG104 | 47.0 | 65.0 | 51.0 | 26.5 | 50.2 | 60.7 |
| FCOS | X101-64x4d-DCN | 46.6 | 65.9 | 50.8 | 28.6 | 49.1 | 58.6 |
| CornerNet | HG104 | 40.6 | 56.4 | 43.2 | 19.1 | 42.8 | 54.3 |
| CenterNet | HG104 | 42.1 | 61.1 | 45.9 | 24.1 | 45.5 | 52.8 |
| CentripetalNet | HG104 | 46.1 | 63.1 | 49.7 | 25.3 | 48.7 | 59.2 |
| RetinaNet | R101 | 39.1 | 59.1 | 42.3 | 21.8 | 42.7 | 50.2 |
| FSAF | X101-64x4d | 42.9 | 63.8 | 46.3 | 26.6 | 46.2 | 52.7 |
| RepPoint | R101-DCN | 45.0 | 66.1 | 49.0 | 26.6 | 48.6 | 57.5 |
| RepPointV2 | R101-DCN | 48.1 | 67.5 | 51.8 | 28.7 | 50.9 | 60.8 |
| FreeAnchor | X101-64x4d | 46.0 | 65.6 | 49.8 | 27.8 | 49.5 | 57.7 |
| ATSS | X101-64x4d-DCN | 47.7 | 66.5 | 51.9 | 29.7 | 50.8 | 59.4 |
| PAA | X101-64x4d-DCN | 49.0 | 67.8 | 53.3 | 30.2 | 52.8 | 62.2 |
| TSD | X101-64x4d-DCN | 49.4 | 69.6 | 54.4 | 32.7 | 52.5 | 61.0 |
| AutoAssign | X101-64x4d-DCN | 49.5 | 68.7 | 54.0 | 29.9 | 52.6 | 62.0 |
| GFL | X101-64x4d-DCN | 48.2 | 67.4 | 52.6 | 29.2 | 51.7 | 60.2 |
| YOLOv4 | CSPDarkNet-53 | 43.5 | 65.7 | 47.3 | 26.7 | 46.7 | 53.3 |
| DETR | R101 | 43.5 | 63.8 | 46.4 | 21.9 | 48.0 | 61.8 |
| Sparse R-CNN | X101-64x4d-DCN | 48.9 | 68.3 | 53.4 | 29.9 | 50.9 | 62.4 |
| Deformable DETR | X101-64x4d-DCN | 50.1 | **69.7** | 54.6 | 30.6 | 52.8 | **65.6** |
| Boosting R-CNN | R50 | 43.5 | 62.8 | 47.3 | 26.0 | 46.4 | 53.6 |
| Boosting R-CNN | R2-101-DCN | **50.6** | 68.9 | **55.7** | **31.6** | **53.9** | 63.4 |

#### 2.4.4.4 在 COCO 数据集上的实验结果

本小节将在主流的 COCO 数据集上对 Boosting R-CNN 进行测试。对比的方法包括二阶段检测器 Faster R-CNN[9]、Cascade R-CNN[25]、Grid R-CNN[111]、Libra R-CNN[85]、Dynamic R-CNN[29]、BorderDet[112]、Double-Head[96]、TridentNet[113]、CPN[114]，一阶段检测器 FCOS[23]、CornerNet[21]、CenterNet[22]、CentripetalNet[115]、RetinaNet[13]、FSAF[103]、RepPoint[24]、RepPointV2[116]、FreeAnchor[26]、ATSS[27]、PAA[117]、TSD[118]、AutoAssign[30]、GFL[104]、YOLOv4[119]，以及基于请求（Query-based）的检测器：DETR[120]、Sparse R-CNN[121]、Deformable DETR[122]。测试中所有的方法均使用单精度测试。结果如表 2.5 所示。Boosting R-CNN 在 ResNet50 的骨干网络下已经达到了 43.5%AP 的性能，在 Res2Net+DCN 骨干网络的基础上更是达到 50.6%AP 的性能，超过所有对比方法，比 GFL 高 2.4%AP，比 RepPointV2 高 2.5%AP、比 TSD 高 1.2%AP。图 2.11 显示的是 Boosting R-CNN 在 COCO 数据集的检测结果可视化，能发现本方法能够在密集遮挡的情况下更准确地识别出物体的位置与类别。

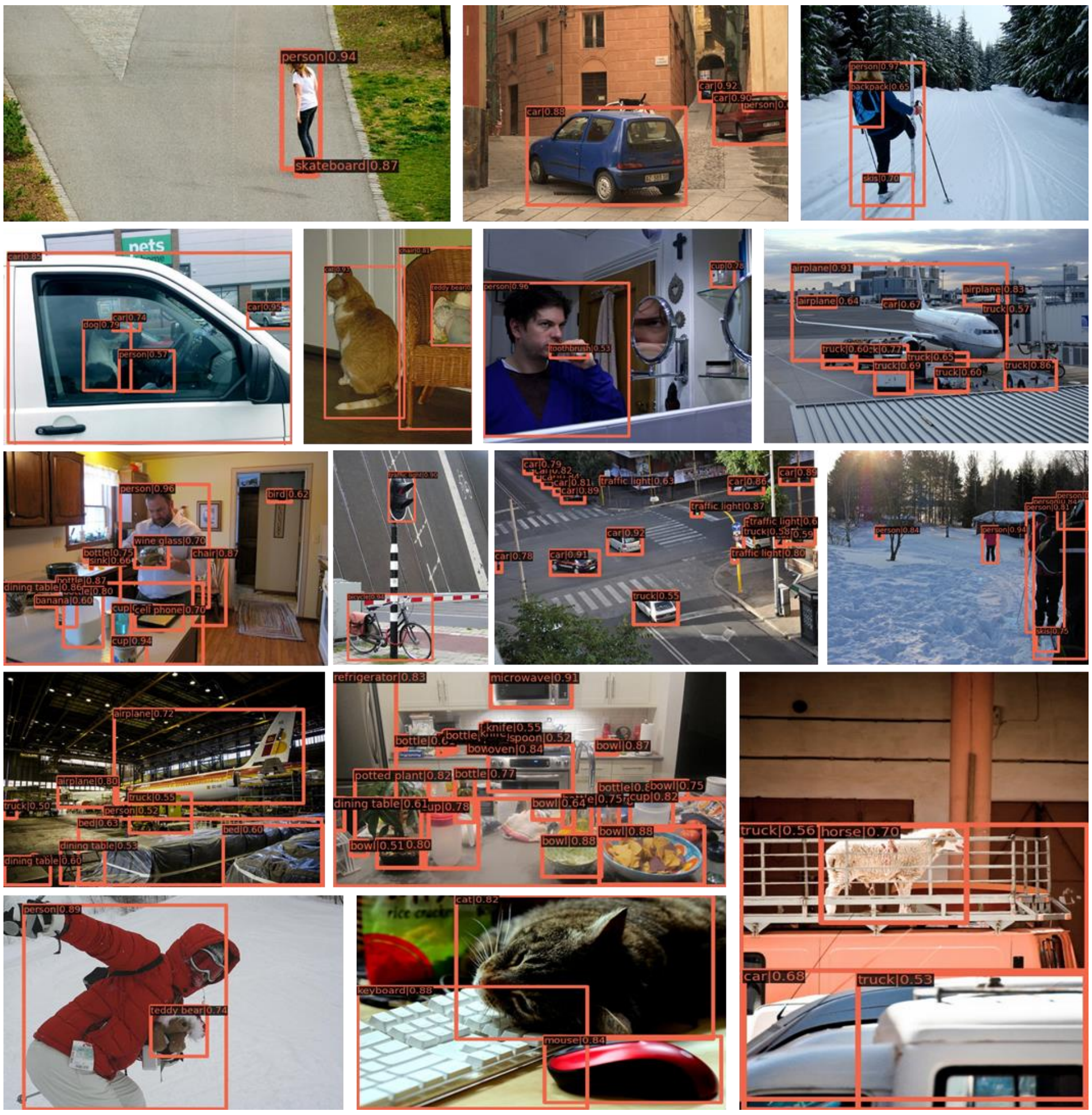

图 2.11 在 COCO 数据集上的可视化结果

Figure 2.11 The visualization of detection results in COCO dataset

表 2.6 Boosting R-CNN 的消融实验
Table 2.6 A detailed ablation study of Boosting R-CNN

| 行号 | RetinaRPN | | | | | Prob | BR | AP | AP50 | AP75 |
|---|---|---|---|---|---|---|---|---|---|---|
| | 4l. | NA. | 回归损失 | FL | IoUp | | | | | |
| 1 | | 3 | L1 | | | | | 44.5 | 80.9 | 44.1 |
| 2 | √ | 3 | L1 | | | | | 45.1 | 81.6 | 45.9 |
| 3 | √ | 3 | L1 | | | | √ | 45.3 | 81.6 | 45.8 |
| 4 | √ | 3 | L1 | √ | | | | 45.4 | 81.2 | 46.3 |
| 5 | √ | 3 | GIoU | √ | | | | 45.8 | 80.5 | 47.5 |
| 6 | √ | 9 | L1 | √ | | | | 46.7 | 80.0 | 49.0 |
| 7 | √ | 9 | GIoU | √ | | √ | | 46.8 | 82.2 | 48.8 |
| 8 | √ | 9 | L1 | √ | | √ | | 47.2 | 82.5 | 49.3 |
| 9 | √ | 9 | L1 | √ | √ | √ | | 47.5 | **83.0** | 50.3 |
| 10 | √ | 9 | GIoU | √ | √ | √ | | 47.6 | 82.7 | 50.4 |
| 11 | √ | 9 | CIoU | √ | √ | √ | | 47.6 | 82.8 | 50.1 |
| 12 | √ | 9 | F-EIoU | √ | √ | √ | | 47.7 | **83.0** | 49.8 |
| 13 | √ | 9 | FIoU | √ | √ | √ | | 47.9 | 82.8 | 50.7 |
| 14 | √ | 9 | FIoU | √ | √ | √ | √ | **48.2** | 82.7 | **51.3** |

表 2.7  和不同困难样本挖掘任务进行对比
Table 2.7 Comparisons with other hard example mining methods

| 分类损失 | 采样方法 | AP | AP50 | AP75 |
|---|---|---|---|---|
| CE | 随机采样 | 47.9 | 82.8 | 50.7 |
| CE | BR(1.1) | 47.9 | 82.5 | 51.1 |
| CE | BR(1) | **48.2** | 82.7 | **51.3** |
| CE | BR(0.5) | 47.8 | **82.9** | 50.3 |
| CE | BR(0.25) | 47.8 | 82.6 | 50.4 |
| CE | OHEM | 47.5 | 82.1 | 49.8 |
| CE | PISA | 46.9 | 82.7 | 48.2 |
| FL (0.25, 2) | 随机采样 | 44.7 | 79.7 | 45.8 |
| FL (0.5, 1) | 随机采样 | 46.5 | 81.1 | 48.8 |
| FL (0.5, 0.1) | 随机采样 | 47.0 | 82.0 | 48.9 |
| FL (0.25, 2) | 随机采样 | 46.7 | 80.2 | 49.7 |

表 2.8 不同正负样本分配方法的效果
Table 2.8 Different positive and negative assignment methods

| 分配方法 | AP | AP50 | AP75 |
|---|---|---|---|
| (0.5, 0.5) | **47.9** | **82.8** | **50.7** |
| ATSS | 46.5 | 81.9 | 48.1 |
| PAA | 47.3 | 83.1 | 49.1 |

#### 2.4.4.5 消融实验

本小节将在 UTDAC2020 数据集上对 Boosting R-CNN 进行一系列的消融实验以验证各个模块的有效性。表 2.6 展示了 Faster R-CNN 到 Boosting R-CNN 的改进过程。（1）第 1 行和第 2 行对比可知，将 FPN 的层次从 P2-P6 转到 P3-P7，并使用 4 层带 GN 的卷积能有效提高模型的特征提取能力，从而提升性能（44.5%AP 到 45.1%AP）；（2）使用 Focal loss（第 4 行）可以将所有的样本加入到考虑中，进一步提升性能（45.4%AP）（3）相比于每个像素使用 3 个 Anchor，使用 9 个 Anchor 能显著提升性能（第 6 行，46.7%AP），这也同时说明了 Anchor 对于水下目标检测任务的重要性；（4）使用 RetinaRPN 之后，对于目标提议的先验概率估计会更加准确，因此改进为概率型二阶段检测器可以将一阶段的先验不确定性考虑到最终检测上，将性能从 46.7%AP 提升至 47.2%AP（第 8 行）。（5）加入一个 IoU 预测任务能够提供更为准确的先验概率，进一步提升对于水下模糊目标的鲁棒性（第九行，性能提升至 47.5%AP）。（6）本文所提出的 FIoU loss（第 13 行，47.9%AP）比 L1 loss（第九行，47.5%AP）、GIoU loss（第 10 行，47.6%AP）、CIoU loss（第 11 行，47.6%AP）和 Focal EIoU loss（第 12 行，47.7%AP）都要优越；（7）单独使用 BR（第 3 行，45.3%AP）就能够实现较为不错的性能提升，甚至高于表 2.2 的 OHEM（45.1%AP）。而 BR 更是可以将最终性能提升至 48.2%AP（第 14 行）。

因为 BR 也是困难样本挖掘的一种，因此有必要将 BR 与其他困难样本挖掘的方法进行对比，比如 OHEM、PISA 和 Focal loss。本节将完整 Boosting R-CNN 中的 BR 替换成其他的方法以测试它们与概率型二阶段检测器的兼容性（如表 2.7 所示）。第一行（47.9%AP）相当于表 2.6 的倒数第二行。当专注参数$\gamma$被设为 1 时，BR 可以获得最佳的性能，而其他的参数设置对性能也没有太大的影响。使用 OHEM 会得到一个相对低的性能（47.5%AP），使用 PISA 更是会让性能进一步下降（46.9%AP）。Focal loss 也会造成严重的性能下降。当$\gamma$设置得很小时，这意味着 Focal loss 会退化成交叉熵损失，因此性能稍稍有回升。从这一实验可以下结论，BR 可以辅助第二阶段的 R-CNN 去修正第一阶段的 RPN 的错误，因此相比于 OHEM、PISA、Focal loss 等困难样本方法更具有兼容性。

正负样本分配也是目标检测中的一个关键技术，因此本节考虑将其他动态正负样本分配的技术应用到当前方法中，如表 2.8 所示，如 ATSS 和 PAA。表 2.8 的实验没有加入 BR。尽管 ATSS 和 PAA 在一阶段检测任务中取得了非常好的效果，但是一旦其用于二阶段检测器的第一个阶段，即 RPN 上时，会造成性能的下降。其原因是动态的正负样本分配实际上会使得模型对目标提议给出一个置信度过高的先验，这对模型的召回率产生了负面的影响。而 Boosting R-CNN 使用的最朴素的 IoU 阈值为 0.5 的方法反而能得到最好的性能

#### 2.4.4.6 鲁棒性测试

本小节将在 UTDAC2020 数据集上对 Boosting R-CNN 进行鲁棒性测试，测试模型在各种样式的畸变下的鲁棒性。本节采用 C. Michaelis 等人[123]提出的 15 种畸变（如图 2.12）中应用在 UTDAC2020 测试集上，利用训练好的模型在测试集进行测试，并将得到的 AP 进行记录。实验结果如表 2.9 所示，Boosting R-CNN 在各种畸变结果之后的检测性能都高于 OHEM。对比性能下降幅度，Boosting R-CNN 在 Gaussian Noise、Shot Noise、Impluse Noise、Fog、Brightness、Contrast、Pixelate、Jpeg Compression 八种变换上下降幅度低于 OHEM，平均下降幅度为 16.35% AP，也低于 OHEM（16.45% AP）。实验证明本方法能够更有效提升模型的鲁棒性。

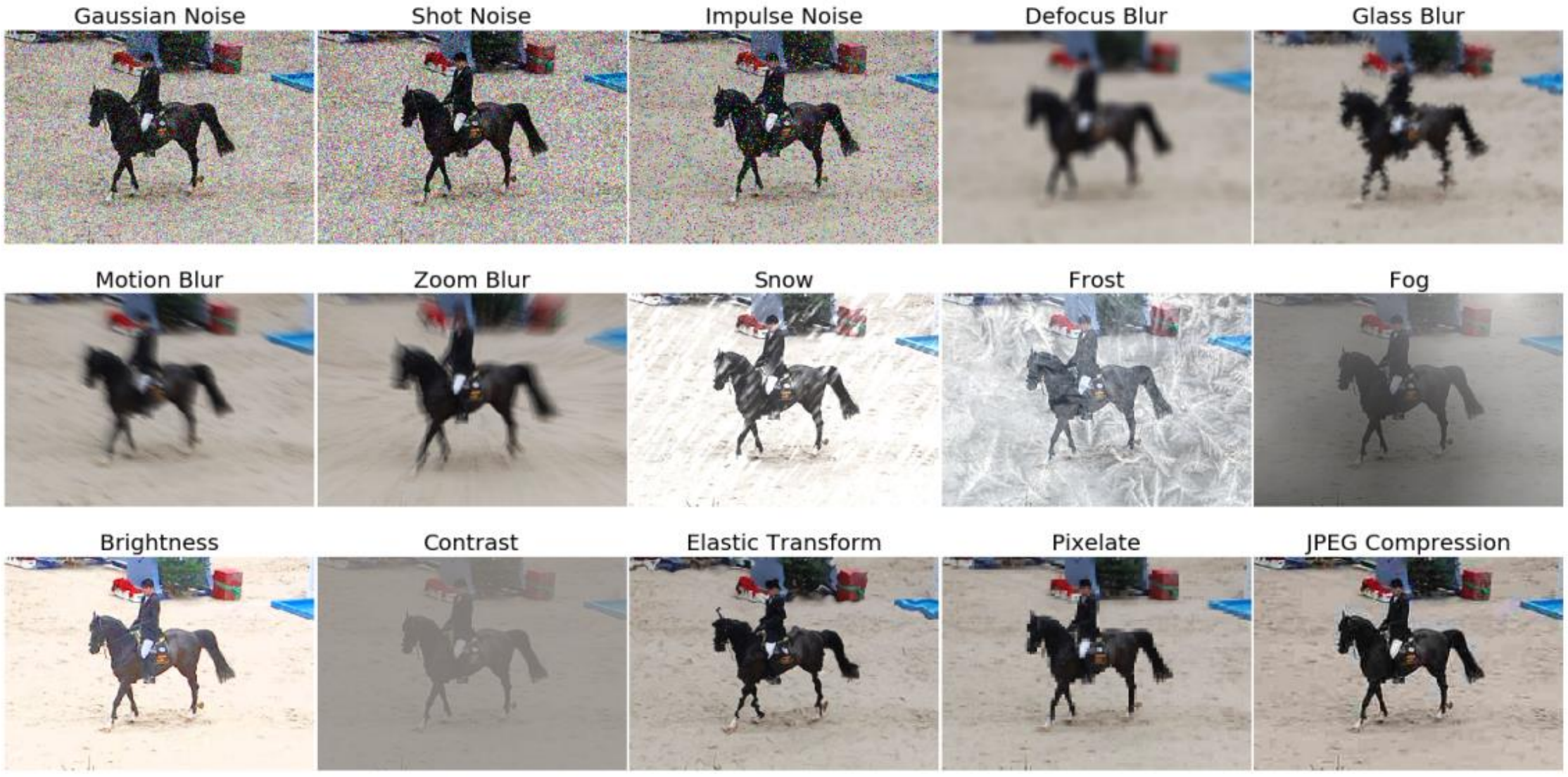

图 2.12 15 种畸变形式的可视化[123]

Figure 2.12 The visualization of 15 corruption types

表 2.9 鲁棒性测试
Table 2.9 Robustness test

| | Boosting R-CNN | OHEM |
|---|---|---|
| No Curruption | 48.2 | 45.1 |
| Gaussian Noise | 17.2 (-31.0) | 13.1(-32) |
| Shot Noise | 17.8 (-30.4) | 13.5 (-31.6) |
| Impluse Noise | 13.1 (-35.1) | 9.9 (-35.2) |
| Defocus Blur | 44.6 (-3.6) | 41.9 (-3.2) |
| Glass Blur | 46.7 (-1.5) | 43.9 (-1.2) |
| Motion Blur | 45.2 (-3.0) | 42.4 (-2.7) |
| Zoom Blur | 12.2 (-36.0) | 11.1 (-34.0) |
| Snow | 11.2 (-37.0) | 8.5 (-36.6) |
| Frost | 14.6 (-33.6) | 11.7 (-33.4) |
| Fog | 32.6 (-15.6) | 28.4 (-16.7) |
| Brightness | 47.3 (-0.9) | 44.0 (-1.1) |
| Contrast | 38.7 (-9.5) | 34.8 (-10.3) |
| Elastic Transform | 47.6 (-0.6) | 44.7 (-0.4) |
| Pixelate | 47.5 (-0.7) | 44.3 (-0.8) |
| Jpeg Compression | 41.4 (-6.8) | 37.6 (-7.5) |
| Average | 31.85 (-16.35) | 28.65 (-16.45) |

## 2.5 本章小结

本章介绍提出了一种概率型二阶段检测器 Boosting R-CNN，相比于传统的二阶段 Faster R-CNN 方法。该方法有三个核心创新部件：视网膜区域提议网络、概率型推断、提升再权重模块。视网膜区域提议网络是一个强力的基于 Anchor 的一阶段检测器，拥有更强的特征提取能力，同时执行三种监督任务，其提出的快速交并比损失相比于其他回归损失有着更快更好的收敛性能。概率型推理结合一阶段对于目标的先验概率估计和二阶段的类别似然分数，构建完整的边缘分布进行推理，能够充分对模糊目标的不确定性进行建模。提升再权重模块是一种困难样本挖掘方法，在训练时，二阶段会着重训练一阶段犯错的样本，在概率型二阶段检测器框架下的推理过程中，第二阶段可以修正第一阶段的错误。实验结果表明，提出的方法在各个数据集的均能达到很好的性能，各种消融实验验证了提出各模块的有效性。

# 第三章　水下目标检测的域泛化

当水下目标检测算法走向实用时，一旦已经针对某几项类别训练得到一个水下目标检测器，自然希望这个检测器可以被用于任何水域。换言之，建立一个通用性的水下目标检测器（General Underwater Object Detector，GUOD）是非常必要的。要建立一个 GUOD 面临着三大项挑战：

（1）获取水下图像要比获取通用目标的图像困难得多，而且这些标注任务通常需要专家来完成，成本很高。因此，水下目标检测的标记数据集非常有限，不可避免地会导致深度模型的过度拟合。数据扩充旨在解决数据不足的问题。有三种类型的增强。第一是空间转换，如翻转、旋转、仿射变换等；第二是基于裁剪-粘贴的方法，如 CutMix、Mixup 等；第三种是基于风格迁移的方法。

（2）速度与性能之间的矛盾会变的更加严峻。一个 GUOD 应该是要可以实时工作的，这在机器人领域上是一个非常常规的要求。但水下机器人往往体积不大，由于在水下极端环境工作，功耗有限且大部分功耗要提供给机器人的运动机能，因此在水下机器人上搭载一个高性能图形硬件来驱动高精度大模型是一件不太现实的事情。

（3）深度学习经常会受到域迁移的影响，但一个 GUOD 必须对水质的变化是鲁棒的，不仅仅只是适用于海洋中，在江河湖畔的水质中也可以同样适用。这可以被看作是域泛化问题：模型在一系列的源域数据上训练，但在一个未曾见过但有所联系的目标域上进行测试。

本章旨在使用一个带有很少域多样性的小数据集，实现一个 GUOD 的训练。为了解决挑战（1），本章提出一种新的数据增强方法水质转移（Water Quality Transfer，WQT）去扩大数据集和增强域多样性；为了解决挑战（2）和挑战（3），本章基于实时目标检测器 YOLOv3 提出了 DG-YOLO 模型以进一步提高模型的域不变性。本章的方法将在 S-URPC2019 数据集上进行实验，实验结果体现了本章方法的有效性。

## 3.1　数据集设置及域泛化问题的发现

### 3.1.1　URPC2019 数据集

本章采用了 URPC2019 数据，其来自于 2019 年水下机器人采集比赛（Underwater Robot Picking Contest 2019）。数据集有 5 个类别：海胆（Echinus）、海参（Holothurian）、海星（Starfish）、扇贝（Scallop）和海草（Waterweeds）。训练集中有 4707 张图像，但测试集的标注并不公开，因此手动将原来的训练集分割为 3765 张图片的训练数据和

942 张图片的验证数据。数据的图片有四种分辨率：3840×2160、1920×1028、720×405 和 586×480。在性能测试方面，本节采用 2.4.3 所提及的 Pascal VOC 的 mAP （即 COCO 评测下的 AP50）作为评测手段。

### 3.1.2 水质迁移

如图 3.1 所示，基于水下目标检测数据集 URPC2019，本章设计了一种数据增强方法水质迁移（Water Quality Transfer，WQT），并构建了 S-URPC2019 以测试模型在域迁移下的鲁棒性。首先，使用真实风格迁移 WCT2[124]模型，选取了 7 种不同的水质，对 URPC2019 的训练集进行处理生成[type1-type7]7 个域的训练集，被用作源域数据集。对 URPC2019 的验证集进行处理生成[val1-val8]8 个域的验证集，被用作目标数据集。模型在源域数据上训练，在验证数据上进行测试。模型从未接触过 val8 域的数据，因此可以认为在 type8 上的性能代表了模型的域泛化性能，在 val8 上的性能越好，说明模型对水质变化越鲁棒，即模型在未见过的水质上也能得到很好的效果。这种问题设置相比通常的目标检测任务更接近真实使用场景。

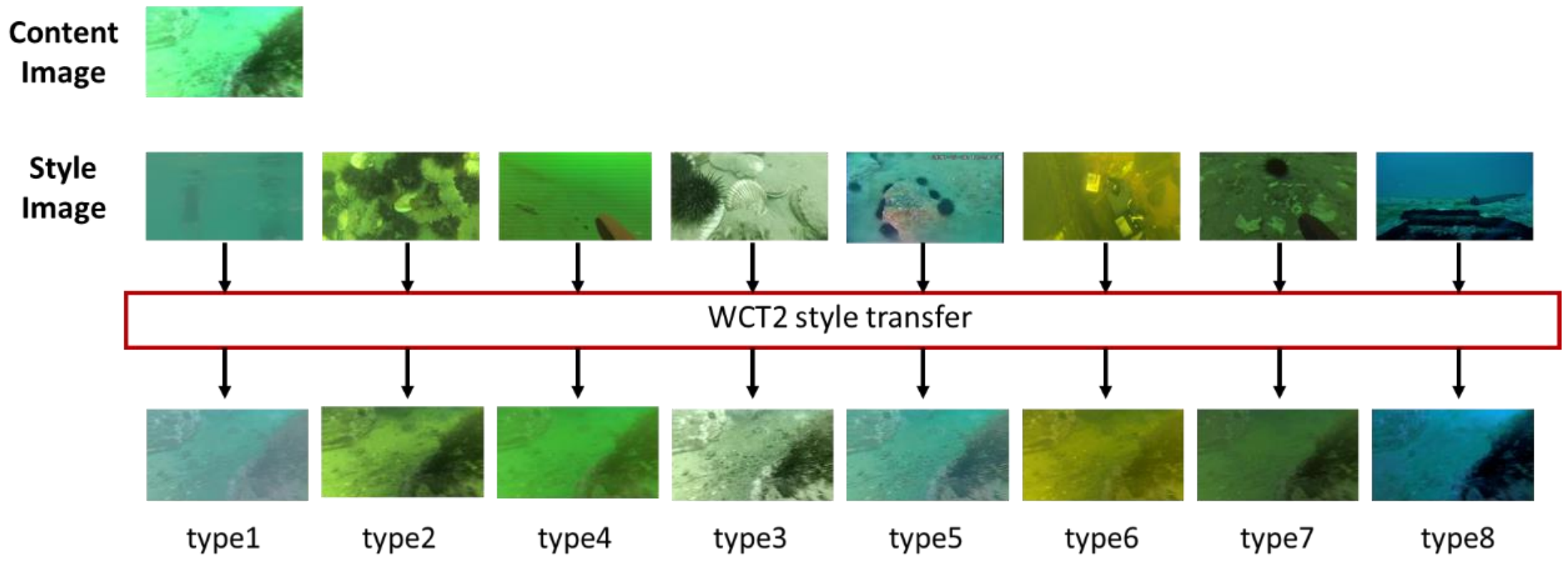

图 3.1 水质迁移

Figure 3.1 Water Quality Transfer

### 3.1.3 域泛化问题

本节使用 S-URPC2019 数据和 YOLOv3[20]算法进行实验，以验证域泛化问题在水下环境下的存在。如表 3.1 所示，“baseline”指的是 YOLOv3 在原 URPC2019 的训练集上进行训练，“ori+type1”指的是 YOLOv3 在原 URPC2019 数据集和 type1 数据集上训练，“Full WQT”指的是 YOLOv3 在 type1 到 type7 的 7 个数据集上进行训练。从表 3.1，可以得到一些有趣的发现：

（1）相比于 Baseline，可以得出以下结论：每一组的数据增强都有助于提升原数据集的性能。WQT 可以和其他的数据增强一起使用以获得更高的性能（表 3.1 的最后一行），这进一步证明了它的有效性。其次，存在一个现象：WQT 也帮助了模型更好地在

其他水质泛化。比如，ori+type7 在 type3 上测试的得到了 39.23%mAP 的性能，相比 baseline 高了 12.66%。

表 3.1 YOLOv3 在各种水质下训练的性能
Table 3.1 The performance of model augmented by different types of water quality

| 训练集 | Evaluation (mAP) | | | | | | | |
|---|---|---|---|---|---|---|---|---|
| | val | val1 | val2 | val3 | val4 | val5 | val6 | val7 |
| baseline | 56.45 | 18.72 | 16.83 | 26.57 | 10.71 | 23.66 | 9.38 | 29.04 |
| ori+type1 | 56.66 | 52.39 | 27.66 | 42.07 | 14.25 | 42.79 | 20.96 | 41.07 |
| ori+type2 | 56.71 | 18.90 | 51.86 | 39.44 | 24.89 | 34.51 | 6.21 | 45.85 |
| ori+type3 | 57.78 | 18.01 | 29.96 | 53.10 | 15.63 | 35.07 | 5.68 | 41.20 |
| ori+type4 | 58.33 | 16.80 | 33.50 | 41.57 | 53.85 | 35.52 | 3.72 | 42.30 |
| ori+type5 | 57.63 | 35.35 | 30.73 | 42.04 | 20.04 | 53.12 | 19.41 | 42.28 |
| ori+type6 | 57.19 | 21.64 | 35.63 | 42.19 | 24.37 | 36.04 | 51.22 | 46.15 |
| ori+type7 | 58.43 | 7.57 | 34.81 | 39.23 | 15.52 | 32.77 | 3.88 | 52.36 |
| FullWQT | **58.56** | **55.93** | **53.60** | **57.48** | **54.95** | **56.08** | **53.51** | **54.29** |
| Ori+rot+flip | 62.53 | 14.81 | 18.29 | 31.36 | 8.89 | 24.95 | 5.34 | 33.18 |
| FullWQT +rot+flip | **63.83** | **60.57** | **57.71** | **60.38** | **58.96** | **59.84** | **58.43** | **60.53** |

表 3.2 水质之间的风格距离
Table 3.2 Style distance between types of water quality

| | Type1 | Type2 | Type3 | Type4 | Type5 | Type6 | Type7 |
|---|---|---|---|---|---|---|---|
| Type1 | 0 | 0.6281 | 0.1105 | 0.6893 | 0.0495 | 0.7239 | 0.6286 |
| Type2 | 0.6281 | 0 | 0.2860 | 0.0052 | 0.3311 | 0.0077 | 0.0033 |
| Type3 | 0.1105 | 0.2860 | 0 | 0.3435 | 0.0411 | 0.3575 | 0.2977 |
| Type4 | 0.6893 | 0.0052 | 0.3435 | 0 | 0.3747 | 0.4024 | 0.3308 |
| Type5 | 0.0495 | 0.3311 | 0.0411 | 0.3747 | 0 | 0.4024 | 0.3308 |
| Type6 | 0.7239 | 0.0077 | 0.3575 | 0.0074 | 0.4024 | 0 | 0.0094 |
| Type7 | 0.6286 | 0.0033 | 0.2977 | 0.0037 | 0.3308 | 0.0094 | 0 |

（2）本节相信各水质下的性能和水质之间的相似性之间存在相关性。首先，使用风格损失去表示风格距离。将 type1 到 type7 的水下图像扔进 WCT2 中，从编码器和解

码器提取特定层的特征图，计算每两种水质之间的风格距离，风格距离的计算公式为：

$$d_{style}(f_1, f_2) = \left|\left|\boldsymbol{G}(f_1) - \boldsymbol{G}(f_2)\right|\right|_F^2, \tag{3.1}$$

$$\mathbf{G}(f)_{c,c'} = \frac{1}{C \times H \times W} \sum_{h=1}^{H} \sum_{w=1}^{W} f_{h,w,c}\, f_{h,w,c'}^{T}, \tag{3.2}$$

其中，$f_1$和$f_2$代表两种水质的特征图，G(f)为特征图 f 的格拉姆矩阵。每一对水质的风格距离计算结束之后可以得到风格距离矩阵$H_{dist}$。结果如表 3.2 所示。第二，将表 3.1 的第三列到第九列（val1 到 val7）和第二行到第八行（ori+type1 到 ori+type7）的一个 7×7 矩阵提取出来，再将此矩阵的每一行减去第一行（baseline）的对应值，可以得到性能增益矩阵$H_{perf}$。使用皮尔森相关系数法计算$H_{dist}$和$H_{perf}$的相关系数，可以得到 0.4634。从这一分析可以推理得到，WQT 泛化性能的增益来自于不同水质之间的相似性。

（3）为了进一步分析（2）中的发现，将 YOLOv3 在 type8 水质上进行测试（type8 与 type1 到 type7 的水质有显著差异）。使用完整 WQT 训练后的 YOLOv3 比任意一组水质上（包括原数据集）测试的结果都要高。然而，其仍未能在 type8 数据集上取得和非常有效的性能（尽管相比只在原数据集上训练的性能已经有显著提高）。因此，WQT 可以提升模型在封闭数据集上的性能，但仍然未能满足一个 GUOD 的要求。WQT 在开放复杂水下场景仍未能实现完全的泛化性。

## 3.2 DG-YOLO

### 3.2.1 DG-YOLO 的概况

图 3.2 展示了 DG-YOLO 的模型流程图。相比于 YOLOv3，加入了域无关模块和无关风险最小化惩罚项。详细地说，YOLOv3 的骨干网络 darknet-53 可以被视作一个特征提取器，darknet-53 提取出来的特征被送进梯度反转层（Gradient Reversal Layer，GRL），其在梯度反向传播的过程中将梯度取反。这之后，域分类器会将反转梯度后的特征图进行域分类。有着 GRL 和域分类器的辅助，骨干网络就会被强迫抛弃特征图中的水质的信息以欺骗域分类器。最后，DG-YOLO 可以更多地依赖于域无关的语义信息进行预测。除此之外，无关风险最小化惩罚项和 YOLOv3 原来的损失会同时计算得到最终的损失：

$$L_{total} = L_{yolo} + \lambda \cdot L_d + \lambda \cdot P_{IRM}, \tag{3.3}$$

其中，$\lambda_p$和$\lambda_d$在实验中被设为 1。在推断过程，因为域无关模块和无关风险最小化惩罚项可以被去掉，因此 DG-YOLO 和 YOLOv3 的推断速度是一致的。需要强调的是，因

为域标签来自于 WQT，因此 DG-YOLO 是不可以单独使用的。

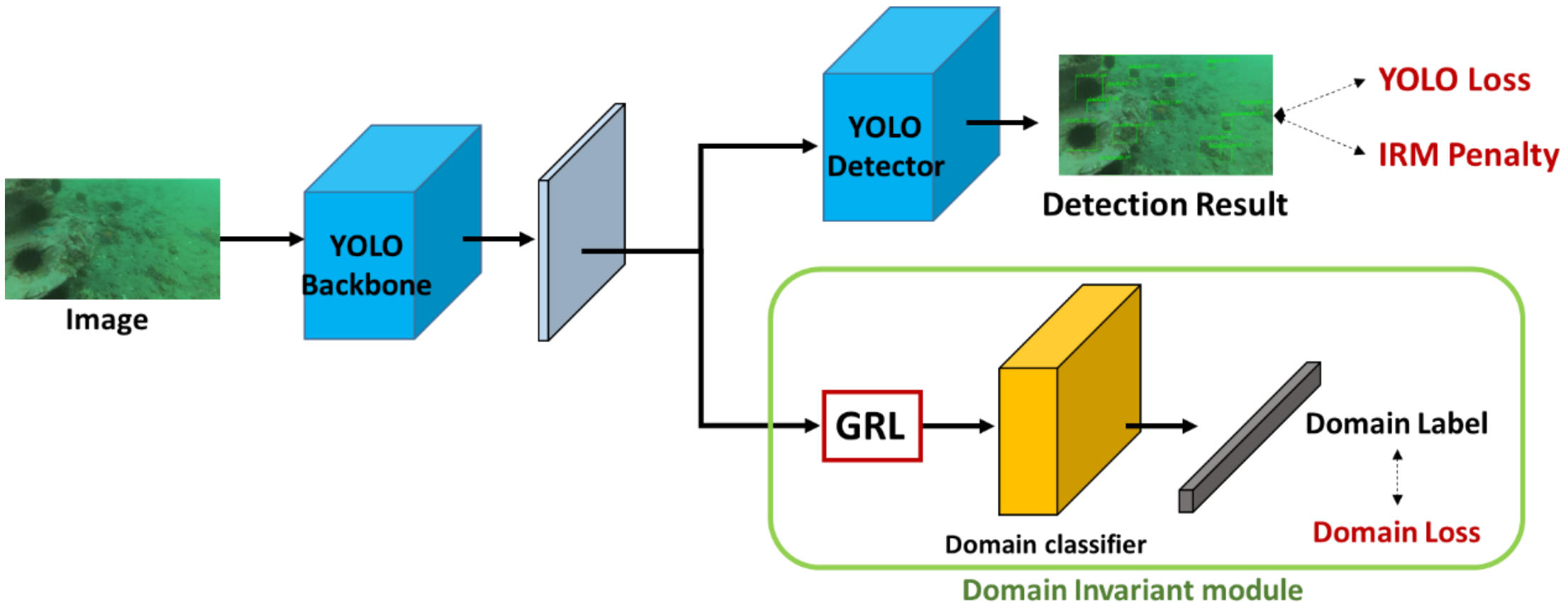

图 3.2 水质迁移和 DG-YOLO 的方法流程图
Figure 3.2 The pipeline of WQT and DG-YOLO

### 3.2.2 回顾 YOLOv3

因为水下机器人的微处理器往往性能受限，实时性目标检测器 YOLOv3 是一个较好的选择。YOLOv3 是一个一阶段检测器，使用 DarkNet-53 作为骨干网络。和 Faster R-CNN 相比，YOLOv3 不需要使用区域提议网络。它使用一个全卷积网络直接回归框的坐标以及类别信息。YOLOv3 将原图分为 S×S 个单元格，每一个单元格都负责一个目标（如图 3.3）。

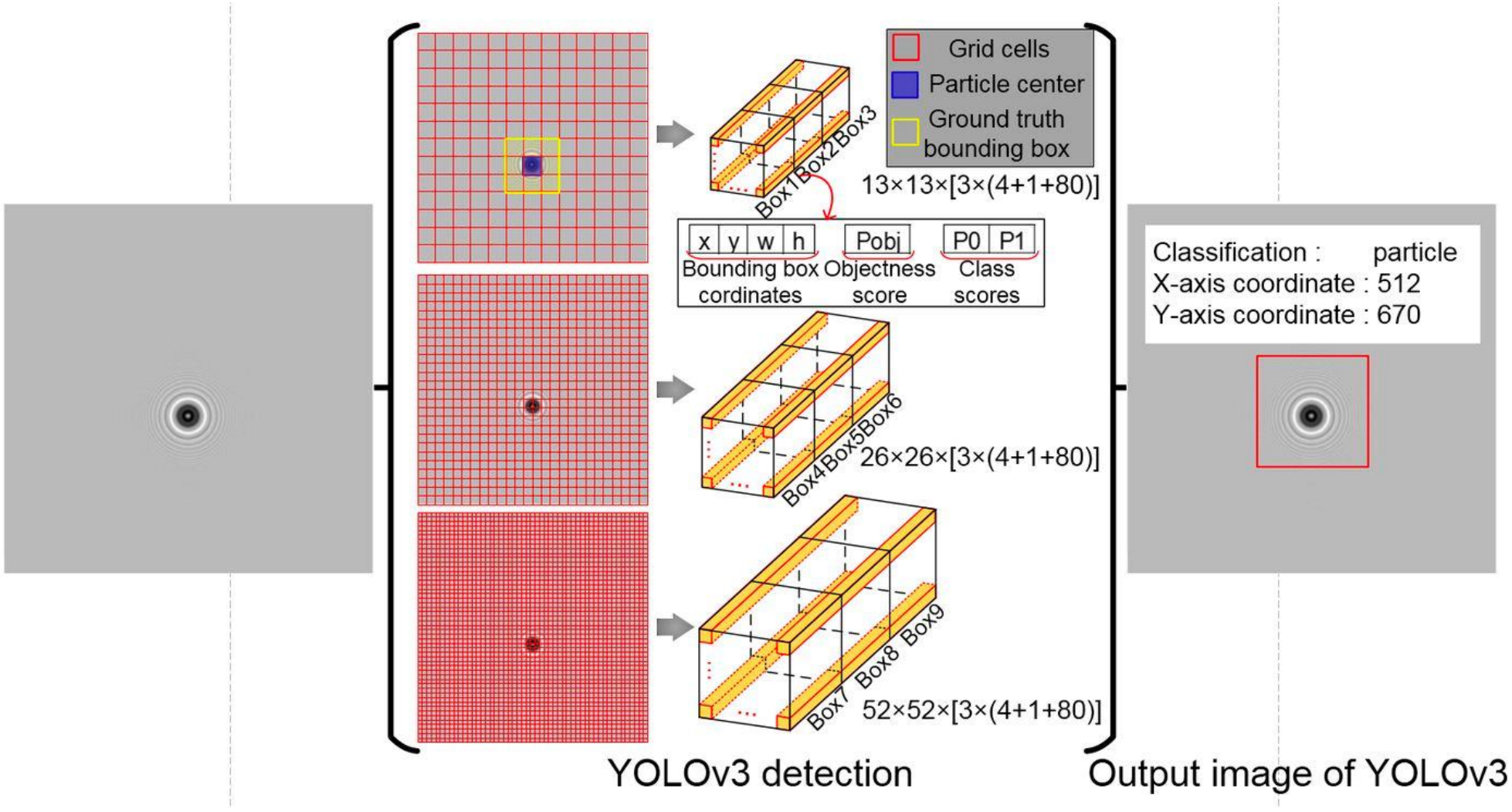

图 3.3 YOLOv3 检测器的训练方式
Figure 3.3 The training paradigm of YOLOv3

YOLOv3 的训练损失包含分类损失$L_{cls}$，坐标损失$L_{coord}$，目标损失$L_{obj}$以及无目标损失$L_{noobj}$：

$$L_{cls}=\sum_{i=0}^{S^2}\sum_{j=0}^{B}\mathbb{1}_{i,j}^{obj}\sum_{c\in classes}l_{CE}(p_i(c),\hat{p}_i(c)),\tag{3.4}$$

$$L_{coord}=\sum_{i=0}^{S^2}\sum_{j=0}^{B}\mathbb{1}_{i,j}^{obj}\sum_{i\in\{x,y,w,h\}}\left|\left|t_i-t_i^*\right|\right|_2^2,\tag{3.5}$$

$$L_{obj}=\sum_{i=0}^{S^2}\sum_{j=0}^{B}\mathbb{1}_{i,j}^{obj}l_{CE}(C_i,\hat{C}_i),\tag{3.6}$$

$$L_{noobj}=\sum_{i=0}^{S^2}\sum_{j=0}^{B}\mathbb{1}_{i,j}^{noobj}l_{CE}(C_i,\hat{C}_i),\tag{3.7}$$

$$\mathrm{L_{yolo}}=L_{cls}+\lambda_{coord}\cdot L_{coord}+L_{obj}+\lambda_{noobj}\cdot L_{noobj},\tag{3.8}$$

其中$\lambda_{coord}$和$\lambda_{noobj}$是协调参数，$\mathbb{1}_{i,j}^{obj}$表示如果预测框$j$由$i$单元格负责则设为 1，$\mathrm{l_{CE}}$代表类别交叉熵损失，$C_i\in\{0,1\}$，$t_i$和$t_i^*$的含义和 2.1.4 节的公式 2.5-2.6 编码方式一致。

### 3.2.3 域无关模块

2014 年，I. Goodfellow 提出了最初的生成式对抗网络[125]，第一次提出了对抗训练的概念。整体的训练包括两部分，生成器和判别器，对于生成器，其目标是随机地生成清晰且语义完好的图像，而对于判别器，则希望其能够分辨出哪一部分是生成器生成的图像，哪一部分是判别器生成的图像。而生成器在训练的过程中，则会想办法去骗过判别器，尽可能生成与真实图像无疑的图像。因此，两者之间存在一个完全相反的优化目标，其最终优化目标为：

$$\mathrm{G}^*=\arg\min_{\mathrm{G}}\max_{D}V(G,D),\tag{3.9}$$

$$\mathrm{V(G,D)}=\mathrm{E}_{\mathrm{x\sim P_{data}}}[\log(D(x)]+E_{x\sim P_G}[\log(1-D(x)],\tag{3.10}$$

其中 G 为生成器，D 为判别器，其输出值为[0,1]区间之间，$x$为输入图像，来自真实分布$P_{data}$或者生成器分布$P_G$。

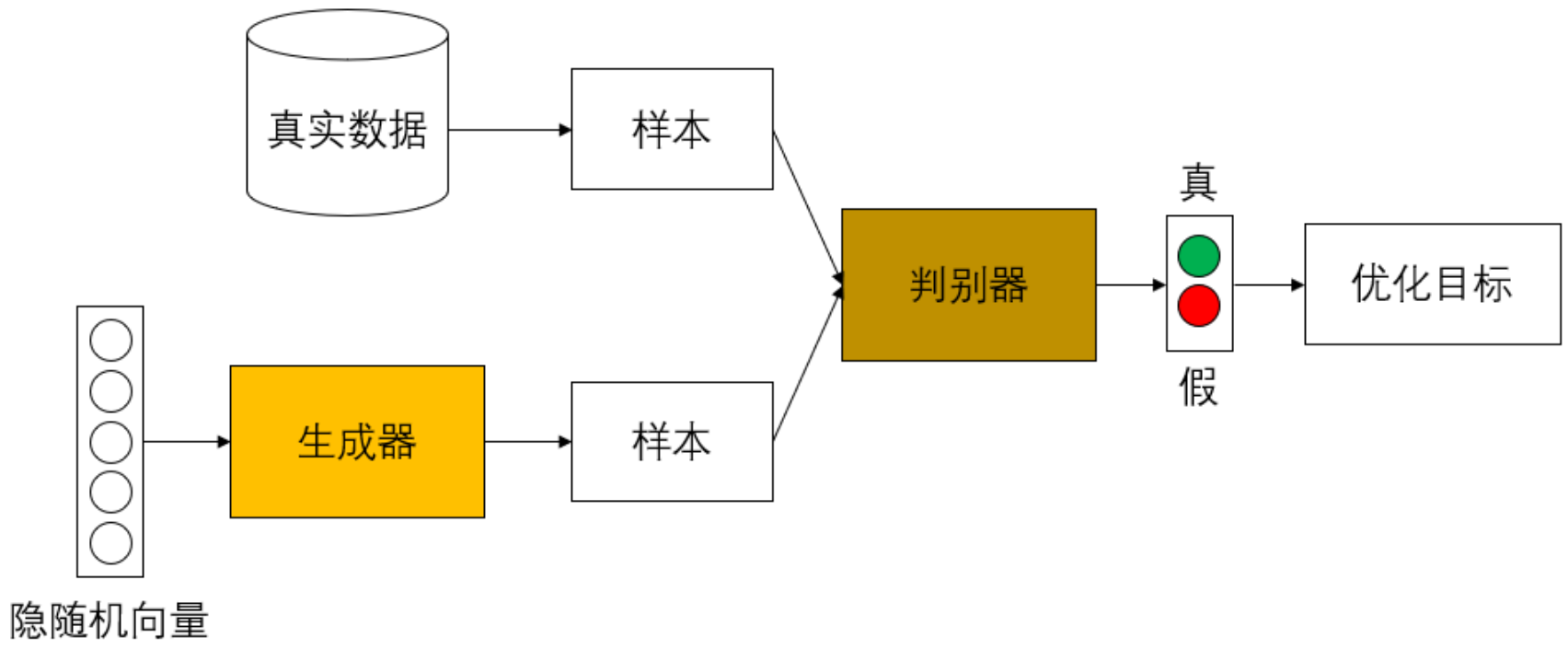

图 3.4 生成式对抗网络的流程图

Figure 3.4 The pipeline of generative adversarial network

在域自适应领域和域泛化领域，对抗式训练的思路可以使用于提取域无关的信息。和生成式对抗网络一样，可以将一个判别器加入到原网络之中，此判别器可以用于判别域的信息，然后使骨干网络往增大判别器损失的方向进行优化，便可以迫使骨干网络提取到域无关的信息，本节称此模块为域无关模块（Domain Invariant Module, DIM）。具体而言，给定一个批的来自 K 个不同域的图像$\{X_1, X_2, \ldots, X_N\}$，其对应的域标签为$\{d_1, d_2, \ldots, d_N\}$，N 是批的大小。将 G 表示为特征提取器，将 D 表示为域判别器，则可以将域损失$L_d$定义为：

$$L_d = \sum_i^N l_{CE}\big(D\big(G(X_i)\big), d_i\big). \tag{3.11}$$

而实际上，域标签即来自于 WQT 生成的不同数据集，K 为 7 代表 7 种不同的源域数据。对于源域的数据，并不计算其域损失。判别器的优化目标是减少域损失。但与生成式对抗网络相反的是，域无关模块不使用生成式对抗网络的生成器和判别器迭代训练的方式，而采用的是端到端的训练方式。将一个梯度反转层加入到域判别器和骨干网络之间，当梯度传到骨干网络的时候，梯度便会取负值，此时骨干网络的优化目标便改变了，朝着原梯度反向，即最大化域损失的方向去优化。在模型训练过程，域判别器会尽可能地挖掘特征中的域信息进行域分类，分类能力越来越强；而骨干网络为了使得域损失增大，使域判别器犯错，会尽可能在特征提取过程中抛弃掉无关的域信息。因此，域无关模块便实现了对域无关特征的提取。

### 3.2.4　无关风险最小化惩罚

无关风险最小化理论（Invariant Rish Minimization，IRM）[126]可以帮助得到在多个域的一个无关预测器。无关风险最大化的理论为：给定一组训练环境（和“域”同义）$e \in \mathcal{E}_{tr}$，其最终目标是得到一个大规模未见但有关系的环境$\mathcal{E}_{all}$ ($\mathcal{E}_{tr} \in \mathcal{E}_{all}$)上获得较好的性能。然而，直接使用经验风险最小化（Emprical Risk Minimization，ERM）将会导致在训练环境上过拟合，以及使得模型学习到数据集上的错误相关性）。为了能够更好泛化到未见的环境，IRM 是一个获得无关性的更好选择：

$$\min_{\Phi:\ \mathcal{X} \sim \mathcal{Y}} \sum_{e \in \mathcal{E}_{tr}} R^e(\Phi) + \lambda \cdot \left|\left|\nabla_{r|r=1.0} R^e(r \cdot \Phi)\right|\right|^2, \tag{3.12}$$

其中，$\Phi$是整个无关预测器，$R^e(\Phi)$是在环境 e 上的 ERM 项，$r = 1.0$是一个常数项，$\left|\left|\nabla_{r|r=1.0} R^e(r \cdot \Phi)\right|\right|^2$是无关惩罚，$\lambda \in [0, \infty)$是协调 IRM 和 ERM 项的参数。

为了将 IRM 理论应用到 YOLOv3 中，本节将为 YOLOv3 特意设计出 IRM 惩罚项，具体如下：

$$\mathrm{Pen_{coord}} = \left\|\nabla_{\mathrm{r|r=1.0}} \sum_{i=0}^{S^2} \sum_{j=0}^{B} \mathbb{1}_{i,j}^{obj} \sum_{i\in\{x,y,w,h\}} \left\| t_i \cdot r - t_i^* \right\|_2^2 \right\|^2, \tag{3.13}$$

$$\mathrm{Pen_{cls}} = \left\|\nabla_{\mathrm{r|r=1.0}} \sum_{i=0}^{S^2} \sum_{j=0}^{B} \mathbb{1}_{i,j}^{obj} \sum_{c\in classes} l_{CE}(p_i(c), \sigma(a_i(c)\cdot r))\right\|^2, \tag{3.14}$$

$$\mathrm{Pen_{obj}} = \left\|\nabla_{\mathrm{r|r=1.0}} \sum_{i=0}^{S^2} \sum_{j=0}^{B} \mathbb{1}_{i,j}^{obj} l_{CE}(C_i, \sigma(\widehat{S_i}\cdot r))\right\|^2, \tag{3.15}$$

$$\mathrm{Pen_{noobj}} = \left\|\nabla_{\mathrm{r|r=1.0}} \sum_{i=0}^{S^2} \sum_{j=0}^{B} \mathbb{1}_{i,j}^{noobj} l_{CE}(C_i, \sigma(\widehat{S_i}\cdot r))\right\|^2, \tag{3.16}$$

其中$r = 1.0$是常数值，σ是 sigmoid 操作，$\hat{a}_i(c)$是类 c 的在 sigmoid 前的分数值，$p_i \in \mathbb{R}^{K\times 1}$是类别标签，$\widehat{S_i}$是 sigmoid 前的目标分数。惩罚项是基于对应的 YOLOv3 的损失（公式 3.4-3.8）设计的。具体来说，常数 r 乘上了损失函数上模型的各个输出值，并且求得 r 对应的梯度平方项，得到对应的惩罚项。

## 3.3 实验与讨论

### 3.3.1 实验设置

本节对 YOLOv3 和 DG-YOLO 训练 300 个 epochs，并且在原验证集和合成数据验证集进行测试。输入的图片尺寸为 416×416。模型在 Nvidia GTX 1080Ti GPU 上进行训练，使用 PyTorch 进行实现，批大小被设为 8。模型使用 Adam 算法进行优化，学习率被设为了 0.001，$\beta_1$和$\beta_2$分别设为 0.9 和 0.999。IoU 阈值，置信度阈值和非极大值抑制阈值均设为 0.5。使用梯度累积技术，即每两次迭代的梯度加起来再集中进行一次下降。对 YOLOv3 和 DG-YOLO 不使用其他的数据增强方法，除非文中提及。

模型的评价指标采用本文 2.4.3 节所提及的 Pascal Voc2012 的评估指标。

### 3.3.2 DG-YOLO 的实验

#### 3.3.2.1 DG-YOLO 的有效性

如表 3.3 所示，WQT 可以辅助 YOLOv3 去学习域无关信息，在 type8 上，WQT 训练的模型比原来的 YOLOv3 高了 23.98%mAP，但模型仍然严重收到域迁移的影响，WQT 训练的模型的 type8 性能比原验证集性能低了 28.01%mAP。DG-YOLO 更进一步地从数据中挖掘了域无关信息，相比于只使用 WQT，在 type8 验证集提升了 3.21mAP 的性能。同时，和其他目标检测器相比，DG-YOLO 展现出更加有效的域泛化性能。

#### 3.3.2.2 消融实验

表 3.4 展示的是消融实验的结果。相比于 WQT-only，WQT+DIM 在 type8 数据集上

由 4.52%mAP 的性能下降，WQT+$P_{IRM}$由小部分提升（0.08%mAP）。然而，WQT+DG-YOLO 可以获得 3.21%mAP 的性能提升，这暗示了只有结合了 DIM 和$P_{IRM}$才能获得更高的性能。

表 3.3 DG-YOLO 和其他目标检测器的性能比较
Table 3.3 Comparison between DG-YOLO and other object detectors

| Method | ori | type8 (mAP) | | | | | |
|---|---|---|---|---|---|---|---|
| | | echinus | starfish | holothurian | scallop | waterweeds | ave. |
| baseline (YOLOv3) | 56.45 | 28.08 | 4.5 | 0.26 | 0 | 0 | 6.57 |
| WQT-only | 58.56 | 60.98 | 17.08 | 33.29 | **39.02** | 2.38 | 30.55 |
| Faster R-CNN+FPN | 58.20 | 29.49 | 5.91 | 9.13 | 1.07 | 10.40 | 11.23 |
| SSD300 | 50.66 | 27.31 | 14.57 | 13.62 | 3.01 | 2.98 | 12.31 |
| SSD512 | 56.51 | 26.62 | 14.44 | 18.07 | 1.41 | 14.51 | 15.22 |
| WQT+DG-YOLO | 54.81 | **63.84** | **27.37** | **35.64** | 36.88 | **5.11** | **33.77** |

表 3.4 DG-YOLO 的消融实验
Table 3.3 Ablation Study of DG-YOLO

| Method | ori | type8 (mAP) | | | | | |
|---|---|---|---|---|---|---|---|
| | | echinus | starfish | holothurian | scallop | waterweeds | ave. |
| WQT-only | **58.56** | 60.98 | 17.08 | 33.29 | **39.02** | 2.38 | 30.55 |
| WQT+DIM | 58.06 | 58.78 | 18.55 | 26.64 | 21.82 | 4.39 | 26.03 |
| WQT+$P_{IRM}$ | 57.01 | 54.99 | 25.98 | 32.90 | 29.25 | 0 | 30.63 |
| WQT+DG-YOLO | 54.81 | **63.84** | **27.37** | **35.64** | 36.88 | **5.11** | **33.77** |

#### 3.3.2.3 DG-YOLO 的可视化

无可忽视的是，WQT+DG-YOLO 相比于 WQT-only 在原验证集上有性能的下降（-3.75%mAP）。这是因为 WQT-only 捕捉到了原数据集上的一些错误的相关性，并以此相关性作为模型判断的标准。举个例子，水草在绿色的水质环境中是绿色的，但是在另外一种水质环境中会变成了黑色。因此，目标的颜色就不是一个域无关的信息，尽管利用此信息在一个封闭单域的数据集上做判断是取巧之法。DG-YOLO 的性能减少可以被解释为模型抛弃了域相关的信息，且尽可能学习到数据集中的域无关信息。本节使用 SmoothGrad[127]的可视化方法进行可视化以证明本节的上文的猜想，找到让模型认为相信此处有 95%的可能性认为是海胆的区域。如图 3.5 所示，baseline 关注到了第一

行图片的左上角阴影，疑似将此处当作海胆，而此处并没有任何海胆。WQT-Only 关注的像素非常散且不集中，这意味着 WQT-only 学习到了一些错误的相关性。而 DG-YOLO 关注的像素非常集中，且恰好关注到了有海胆的位置。这个可视化结果说明 DG-YOLO 比 baseline 和 WQT-only 学习到更多的语义信息。

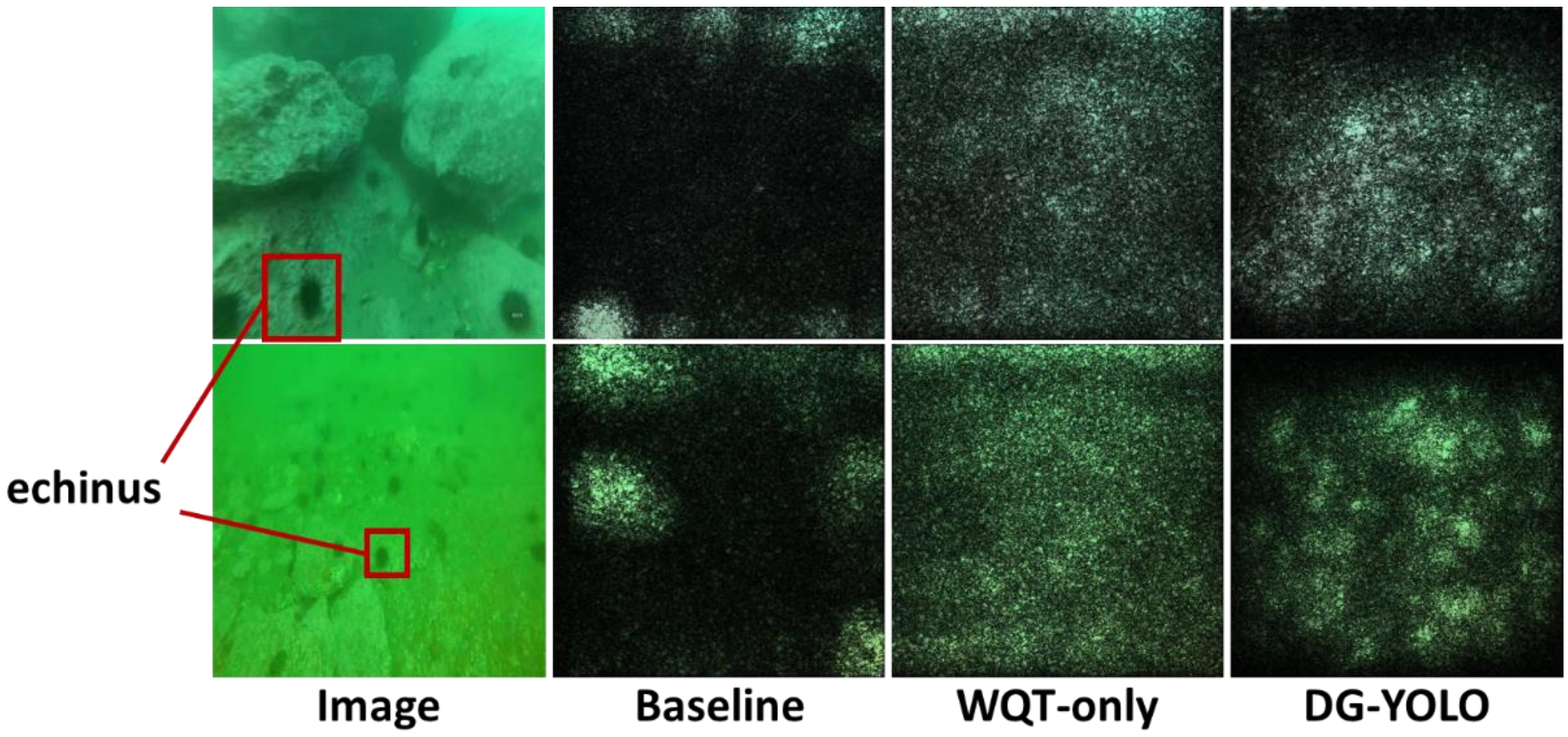

图 3.5 使用 SmoothGrad 对海胆进行可视化

Figure 3.5 SmoothGrad visualization on “echinus” samples

## 3.4 本章小结

本章提出了一种数据增强方法水质迁移，以及一个新的模型 DG-YOLO，以克服一个 GUOD 的三个挑战：受限的数据、实时性处理以及域迁移。水质迁移使用 WCT2 风格迁移模型，旨在增加数据集中的域多样性。DG-YOLO 是 YOLOv3 的改进，包含了两个新模块：域无关模块和无关风险最小化惩罚，两者的辅助使得 DG-YOLO 可以进一步挖掘数据集中的语义信息。在原 URPC2019 数据集和合成数据集的性能展现了本方法卓越的域泛化性能。然而，因为 DG-YOLO 在未知域的性能仍未能达到和已知域的性能的同样水平，在这一领域仍然有许多工作值得去探索。而下一章将会利用本章所构建起的研究方法，更深入地去探讨域泛化问题和域泛化问题的解决。

# 第四章　基于域混合和对比学习的水下域泛化训练框架

正如上一章所揭示的，当面临复杂水下环境带来的域迁移效应时，现存的水下目标检测方法的性能往往会严重下降，这也是当前需要构建一个 GUOD 所必须突破的难题。而在数据集有限数量的域中，深度模型会仅仅只是记住了这些见过的域，而非从这些域中真正提取出了总结性的、预测性的知识，这便会导致域过拟合（Domain Overfitting）和较低的域泛化性能。对于域泛化问题，常用的思路有两种。第一，如果模型能够在尽可能多的域上面进行训练，模型获得域无关的特性。第二，对于有着相同语义内容的却来自不同域的图像，其模型提取出来的特征应该是要一致的。本章将深挖这两个思路，并提出一个域泛化训练框架，称为域混合对比训练。首先，基于水下图像成像模型 IFM 的推理结果，一张水下环境的图像是可以由其他环境的水下图像通过线性加权的方式进行结合。因此，本章使用了一个风格迁移模型，其输出是一个线性变换矩阵而非直接输出一个完整的图像，并将此用于将水下图像从一种水质（域）转换到另外一种水质（域），增加训练数据的域多样性。其次，混合（Mixup）操作将不同域在特征层面上进行线性加权，可以在域流形上采样新的域。最后，对比损失有选择性地作用于来自不同域的特征，以捕捉域无关的特征，但同时保留了模型的辨别能力。使用了本方法可以极大提高模型的域泛化能能力。各种综合实验体现了本方法能够辅助模型学习域无关的表征，并且超越其他域泛化方法的性能。

## 4.1　研究现状

目标检测任务旨在图片中辨别和定位特定类别的目标。基于深度学习的目标检测方法也被用于水下机器人任务，如水下机器人采集任务、鱼的养殖、生物多样性监控等。和其他传统的场景相比，复杂的水下环境会给目标检测技术带来极大的挑战，因为域迁移问题常常会在水下环境出现。比如，一天中的不同时间有不同的光照条件，不同区域的水质环境也会剧烈变化，湖和海的水质就截然不同。除此之外，因为收集和标注水下图像的巨大困难，得到一个有着丰富水质多样性的大规模水下数据集是一个非常困难的事情。在实际操作中，使用一个有限的数据集去训练一个检测器，还要其在变化的水下环境中表现良好，是一个不太现实的事情，因为深度检测器对于域迁移极度不鲁棒。因此，提升模型对于域迁移的鲁棒性是必不可少的。

深度学习模型在自然水下环境上的实际应用可以被视作是域泛化任务：模型在一系列已知域上进行训练，却要在未知但相关的域上进行泛化。之前的大多数工作集中在将源域特征进行对齐，比如最小化最大均值差异（Maximum Mean Discrepancy, MMD）

或者使用对抗训练策略（如第三章）。除此之外，现有的域泛化工作集中在识别任务，对于检测任务的域泛化工作很少。

本章的方法基于两个基本假设：（1）增加在域空间的采样能够帮助改善域迁移的鲁棒性。如图 4.1 所示，数据分布$\mathcal{Z}$取决于两个先验分布：域分布$\mathcal{D}$和语义内容分布$\mathcal{S}$。$\mathcal{D}$构建了一个隐式流形。数据集中小部分的域并不能代表整个域流形的结构，这会导致域过拟合。如果能够采样更多的域进行训练，检测器可以成功消除域迁移的影响。（2）一个理想的特征提取器会将两张有着同样语义内容但来自不同域的图像视作是等价的，这是一个非常直接的思路。然而，简单地惩罚特征之间的差异是不可行的，这会导致网络使网络倾向于对任何输入都有相同的输出（崩溃解），使得网络的辨别能力的下降。因此，本章采样选择和间隔（Margin）的思路去维持网络原来的辨别能力。

本章提出域混合对比训练（Domain Mixup and Contrastive Learning，DMC）方法以解决水下目标检测的域泛化问题。第一，提出了名为条件双边风格迁移（Conditional Bilateral Style Transfer，CBST）风格迁移模型，可以将一张水下图像从一种水质转换到另外一种。第二，提出了域混合（Domain Mixup，DMX），作用于特征层面，将两个不同的域进行插值合成新的域的数据。CBST 和 DMX 都增加了训练数据的域多样性。第三，提出了空间选择性间隔对比损失（Spatial Selective Marginal Contrastive loss，SSMC loss），对模型学到的域特定特征进行正则化。最后，本章提出了一个新的基准 S-UODAC2020。S-UODAC2020 是一个用于水下目标检测的域泛化数据集，可以体现出检测器面对水下域迁移时的鲁棒性。为了构建这一基准，本章将许多主流的域泛化技术迁移到这一数据集中，以和本章提出的方法进行对比。

总而言之，本章的主要贡献为：（1）提出了一个风格迁移模型 CBST，可以将图像从一种域转换成另外一种域，以提高训练数据的域多样性；（2）提出了一个数据增强方法 DMX，旨在通过对现有域进行插值对域流形进行采样；（3） 提出了一个正则化项 SSMC loss，以对骨干网络中得到的域特定信息进行正则化；（4） 提出了一个新的数据集 S-UODAC2020，此数据集是为水下目标检测的域泛化任务而设计的；（5）综合性的实验不仅表现出本方法能够得到更加优越的性能，更证明了提出模块的有效性。

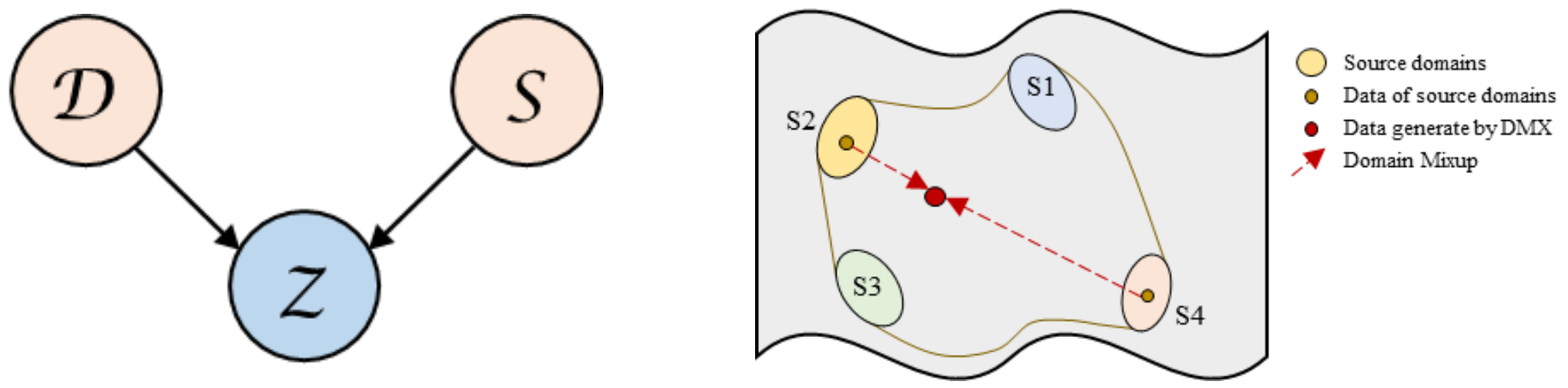

（a）贝叶斯模型　　（b）隐式域流形

图 4.1 域混合的贝叶斯解释

Figure 4.1 The bayesian interpretation of Domain Mixup

(a) Bayesian model; (b) The latent domain manifold

## 4.2　域混合对比训练

本章提出的域混合对比训练流程图如图 4.2 所示，所采用检测器为 Faster R-CNN，主要包含三个主要模块：条件双边风格迁移、域混合和空间选择性间隔对比损失。首先，从一个域获得的图像先被送进 CBST 中转换成另外一种源域，得到成对图像。第二，成对图像送进参数共享的骨干网络中，提取出成对特征，DMX 将成对特征进行插值，可以从源域构成的域凸包中获取新的域。第三，SSMC loss 作用于成对特征上，可以对域相关信息进行正则化。

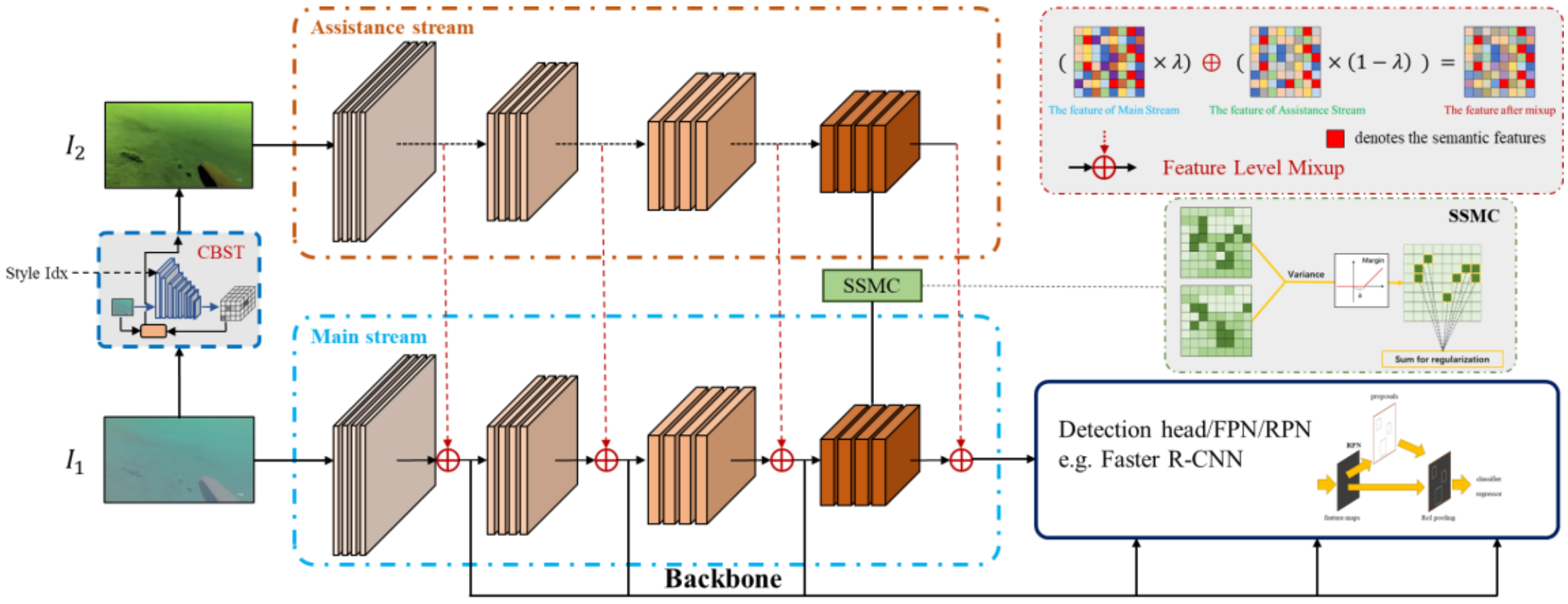

图 4.2　域混合对比训练的流程图

Figure 4.2 The overview of DMC

### 4.2.1　条件双边风格迁移

根据 1.3.1 节所提及的水下图像成像原理，水下图像的形成可以建模为：

$$\mathrm{I}_\lambda(x) = J_\lambda(x)\cdot t_\lambda(x) + \big((1 - t_\lambda(x))\big)\cdot B_\lambda,\quad \lambda \in \{red, green, blue\}, \tag{4.1}$$

其中 x 是水下图像上的一个坐标点，$\mathrm{I}_\lambda(x)$表示被摄像头获取的图像上 x 点的值，$J_\lambda(x)$是隐式清晰图像在点 x 上的值，$t_\lambda(x)$是点 x 上从水下场景反射到相机的残差能量比，$\mathrm{B}_\lambda$是背景光。根据上述公式可知，水下图像 I 可以被视作是清晰图像 J 的在颜色空间上的线性变换，换句话说，两个水下图像之间也可以通过特定的线性变换互相转换。因此，提出了条件双边风格迁移以合成水下图像，其从低分辨率风格图像中学习一个局部颜色仿射变换，应用于高分辨率内容图像上。CBST 的设计是基于实时风格迁移模型 BST[128]的。相比于 BST，CBST 有两个主要的改进。第一个是使用了条件实例归一化[129]（Conditional Instance Normalization，CIN）替代了自适应实例归一化[130]（Adaptive Instance Normalization，AdaIN）。原 BST 模型中所有的 AdaIN 都被置换成了 CIN。CIN 的数学表达式为：

$$\mathrm{CIN}(x; s) = \gamma_s \left(\frac{x-\mu(x)}{\sigma(x)}\right) + \beta_s, \tag{4.2}$$

其中，$\mu$和$\sigma$是 x 的空间维度的均值和标准差，$\gamma_s$和$\beta_s$是风格 s 的缩放和平移参数。使用 CIN 的模型会更加容易训练，消耗更少的推理时间，因为省略了一次从 VGG 提取特征的过程。其次，受到了 Region loss 的影响，CBST 提出了掩码损失。风格迁移会让图像变得非常风格化，导致最终输出图像的语义内容都产生了变化。检测器在过度风格化到已经改变了实际内容的图像上训练反而会造成和标注的不匹配，最终导致性能下降。

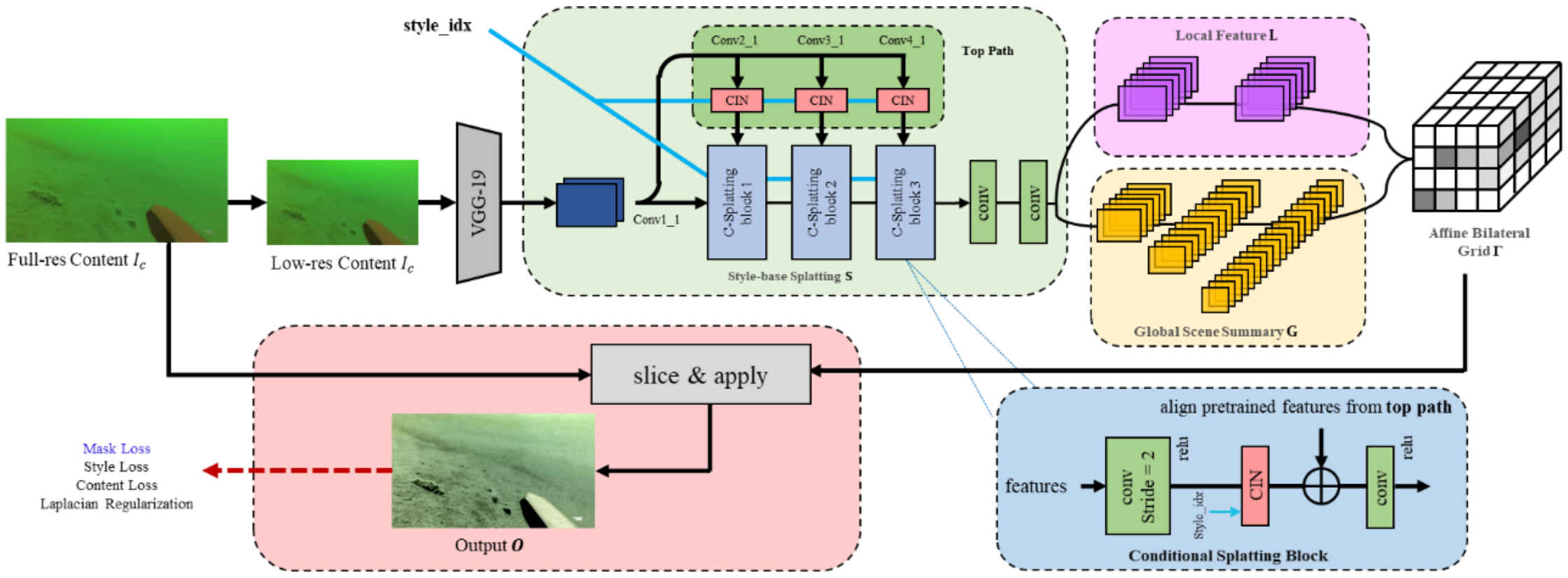

图 4.3 条件双边风格迁移的模型框架

Figure 4.3 The framework of Conditional Bilateral Style Transfer

CBST 的模型结构如图 4.3 所示。首先，一个低分辨率内容图像$I_{c,low}$送进已经预训练好的 VGG19 提取特征$F_{conv1_1}$，$F_{conv2_1}$，$F_{conv3_1}$和$F_{conv4_1}$。第二，$F_{conv1_1}$被送进三个连续的印迹模块（Splatting Block，SB）。每一个 SB 包含一个步幅为 2 的卷积层、一个 CIN 和一个步幅为 1 的卷积层。另外，在经过 CIN 之后的$F_{conv2_1}$，$F_{conv3_1}$和$F_{conv4_1}$会被加到三个 SB 的顶部通路。在三个 SB 之后，跟着两个额外的卷积。其公式如下：

$$F_1 = S_1(F_{conv1_1}, F_{conv2_1}), \tag{4.3}$$

$$F_2 = S_2(F_1, F_{conv3_1}), \tag{4.4}$$

$$F_3 = S_3(F_2, F_{conv4_1}), \tag{4.5}$$

$$F_4 = C_1(C_2(F_3)), \tag{4.6}$$

其中，$S_1$，$S_2$和$S_3$表示三个 SB，$C_1$和$C_2$表示两个额外的卷积层。第三，网络被分割成两个不对称的路径：一个局部路径，一个全局路径。局部路径由全卷积网络构成，用以学习颜色变换，以及设置网格分辨率。全局路径包含卷积和全连接层，考虑了所有像素的特征去学习了场景的一个总结性的特征，可以对颜色变换进行正则化：

$$A = F\left(T\left(L(F_4) \oplus G(F_4)\right)\right), \tag{4.7}$$

其中，L 和 G 分别是是局部特征和全局场景总结，$\oplus$表示 concatenation 操作。F和T是两个卷积层，A 是输出特征图，$A \in \mathcal{R}^{16\times16\times96}$。A 可以被视作一个 16×16×8 的双边网格，每一个网格单元包含 12 个元素，可以被写成一个 3×4 的仿射颜色矩阵，即$A[i,j,k] \in \mathbb{R}^{3\times4}$。

第四，一个全分辨率的内容图像$I_{c,full}$送进了指导图辅助网络（Guidance Map Auxiliary Network，以 P 代替）去获得指导图 g，

$$g = P(I_c, full), \tag{4.8}$$

$$\bar{A}[x,y] = \sum_{i,j,k} \tau(s_x x - i)\tau(s_y y - j)\tau(d \cdot g[x,y] - k)A[i,j,k], \tag{4.9}$$

其中，$I_{c,full} \in \mathbb{R}^{h\times w\times 3}$，$g \in \mathbb{R}^{h\times w\times 1}$，h 和 w 是$I_{c,full}$的高和宽，$\bar{A}[x,y] \in \mathbb{R}^{3\times4}$，[x,y]是内容图像的像素坐标。$s_x$和$s_y$是网格维度相对于全分辨率图像的维度的宽比率和高比率，$\tau(\cdot) = \max(1 - |\cdot|, 0)$。

第五，将颜色仿射变换应用到内容图像上，

$$O[x,y] = I_c[x,y] \otimes \bar{A}[x,y][:,0:3]^T + \bar{A}[x,y][:,3]^T, \tag{4.10}$$

其中，$O \in \mathbb{R}^{h\times w\times 3}$是输出图像，$O[x,y] \in \mathbb{R}^{1\times3}$，$\otimes$指矩阵乘法。

CBST 的损失函数分为四部分：内容损失、风格损失、双边空间拉格朗日正则项和掩码损失。内容损失和风格损失的表达式为：

$$L_c = \sum_{i=1}^{N_s} \left|\left|F_i[O] - F_i[I_c]\right|\right|_2^2, \tag{4.11}$$

$$L_{sa} = \sum_{i=1}^{N_s} \left|\left|\mu(F_i[O]) - \mu(F_i[I_s])\right|\right|_2^2 + \sum_{i=1}^{N_s} \left|\left|\sigma(F_i[O]) - \sigma(F_i[I_s])\right|\right|_2^2, \tag{4.12}$$

其中，$N_C$和$N_S$是从预训练 VGG19 中选出来以代表图像内容和风格的中间层的数量。双边空间拉格朗日正则项的表达式为：

$$L_r(A) = \sum_s \sum_{t\in N(s)} \left|\left|A[s] - A[t]\right|\right|_F^2, \tag{4.13}$$

其中，A[s]是双边网格的一个单元格，A[t]是它其中的一个近邻。双边空间拉格朗日正则项惩罚双边网格 6 个近邻的不同。掩码损失旨在阻止重要语义信息的变化：

$$L_{mask} = \frac{1}{h\cdot w\cdot 3}\left|\left|\left(O - I_{c,full}\right)\cdot M\right|\right|, \tag{4.14}$$

$$M(i,j) = \begin{cases} 1, & if\ (i,j)\ is\ in\ bboxes\ of\ objects \\ 0.01, & \text{otherwise} \end{cases}, \tag{4.15}$$

其中，$M \in \mathbb{R}^{h\times w\times 1}$表示目标框创造的掩码。CBST 的总损失为：

$$L = \lambda_c L_c + \lambda_{sa} L_{sa} + \lambda_r L_r + \lambda_{mask} L_{mask}, \tag{4.16}$$

在实验中，超参数$\lambda_c$、$\lambda_{sa}$、$\lambda_r$和$\lambda_{mask}$分别被设为 0.5，1，0.015 和 1。

### 4.2.2 域混合

CBST 可以实现图像层次的数据增强，但数据集中源域提供的域多样性仍然很有限。为了进一步丰富数据集中的域多样性，可以在特征层面上采样更多的域。通过 CBST，可以得到原始图像$I_1$以及它对应的生成图像$I_2$。在骨干网络中，$I_1$和$I_2$的特征包括两部分的信息：域信息和语义信息。因为$I_1$和$I_2$的标注框是一样的，因此可以认为其语义信息也是一样的，而其无关的域信息是不同的。如果将$I_1$和$I_2$的隐层特征进行插值，语义信息不会改变，而域信息进行插值。因为域流形在隐层空间上更加平缓，线性插值可以在域凸包生成更多新域。$K = [k_1, k_2, \dots, k_n]$表示骨干网络中选中用于进行混合的层。将$I_1$和$I_2$同时送进骨干网络中（主流和支流，如图 4.2 所示）。支流上的特征可以用于增强主流上的特征。在主流上，第 k 层的隐层特征可以表示为：

$$h_{1,k} = \begin{cases} \lambda_k \cdot f_k(h_1, h_{k-1}) + (1-\lambda_k) \cdot f_k\left(h_{2,k-1}\right),\ k \in K \\ f_k\left(h_{1,k-1}\right),\ otherwise \end{cases}, \tag{4.17}$$

在支流上，隐层特征可以表示为：

$$h_{2,k} = f_k\left(h_{2,k-1}\right), \tag{4.18}$$

其中，$h_{1,k}$和$h_{2,k}$分别是是第 k 层主流和支流特征图，$\lambda_k \sim Beta(\alpha,\alpha)$是第 k 层的混合比率，$\alpha \in (0,\infty)$。可以使用主流中域混合前的特征（第 k 层的$f_k(h_{1,k-1})$）或者混合后的特征（第 k 层的$h_{1,k}$）送进检测头中。还有两种梯度反向传播的路径，第一种是只做主流的梯度反向传播，第二种是同时对主流和支流做梯度反向传播。使用 DMX 后的特征来做检测以及对主流和支流同时做反向传播是本章主实验中所采用的策略，但所有的变体的实验仍然会在消融实验中进行。

接着，本节剩余部分将探讨 DMX 为何有效。用$\mathcal{Z}$来代表所有层的隐式特征。可以假设数据分布$\mathcal{Z}$由两个独立的先验分布：域分布$\mathcal{D}$和语义内容分布$\mathcal{S}$。$\mathcal{D}$构建了一个隐式域流形，从贝叶斯模型的观点，可以得到：

$$p(\mathcal{S}|\mathcal{Z}) = \int p(\mathcal{S},\mathcal{D}|\mathcal{Z})d\mathcal{D}$$

$$= \int \frac{p(\mathcal{Z}|\mathcal{D},\mathcal{S}) \cdot p(\mathcal{D}) \cdot p(\mathcal{S})}{p(\mathcal{Z})} d\mathcal{D}$$

$$= \mathrm{E}_{\mathcal{D}}\left[\frac{p(\mathcal{Z}|\mathcal{D},\mathcal{S})\cdot p(\mathcal{S})}{p(\mathcal{Z})}\right], \tag{4.19}$$

检测器可以表示为$q(\mathcal{S}|\mathcal{Z},\theta)$，$\theta$是可学习参数。检测器的优化目标是减少$p(\mathcal{S}|\mathcal{Z})$和$q(\mathcal{S}|\mathcal{Z},\theta)$。借助式子（4.19），优化目标可以被表示为：

$$\begin{aligned} &\operatorname{argmin} \mathrm{KL}(p(\mathcal{S}|\mathcal{Z})||q(\mathcal{S}|\mathcal{Z},\theta)) \\ &= \operatorname{argmin} \int p(\mathcal{S}|\mathcal{Z})\cdot\log\frac{p(\mathcal{S}|\mathcal{Z})}{q(\mathcal{S}|\mathcal{Z},\theta)}d\mathcal{S} \\ &= \operatorname{argmin} \mathrm{E}_{\mathcal{D}}\big[E_{\mathcal{S}}[p(\mathcal{Z}|\mathcal{D},\mathcal{S})\cdot\log q(\mathcal{S}|\mathcal{Z},\theta)]\big]. \end{aligned} \tag{4.20}$$

在离散形式，可以将$p(\mathcal{Z}|\mathcal{D},\mathcal{S})$近似为经验数据分布$\frac{1}{N_{D,S}}\sum_{n=1}^{N_{D,S}}\delta_{Z_n}(Z)$（$N_{D,S}$是特定域$\mathcal{D}$和特定$\mathcal{S}$下的数据的数量），然后可以通过在域分布采样的操作近似式子（4.20）优化目标中的期望值。假设数据集中的源域有限，只有 6 个域，其便覆盖了整个域流形的 6 个小区域。没有大规模的样本，上文提到的近似是不准确的。因此，模型很可能只是“记住”了这些域。这个问题的一个解决方案是去尽可能多地去采样域。CBST 可以生成同样语义不同域的图像，这可以被视作给定$\mathcal{S}$去采样$\mathcal{D}$。假设源域在域流形上构建了一个凸包，DMX 可以通过插值在域流形上采样更多的域。

DMX 选择在特征层面上插值而不在图像层面上插值的原因如下：在 Mixup 的原文中揭示了 Mixup 之所以能实现良好的泛化性在于其迫使模型得到线性的流型特征分布。在分类任务中，当高维的图像输入到图像中，在输出端能够得到一个 C 维的向量（C 为类别数量）。当对输入图像进行插值，并使用插值的标签进行训练，会迫使模型得到某种线性性质。即可以假设深度学习模型有能够将复杂高维的流型空间展平的能力（如图 4.4）。因为在特征层面的流型线性性质更为优良，因此在特征层面上插值相比于在图像层面上插值更为合理，后续的实验也将进一步证明这一观点。

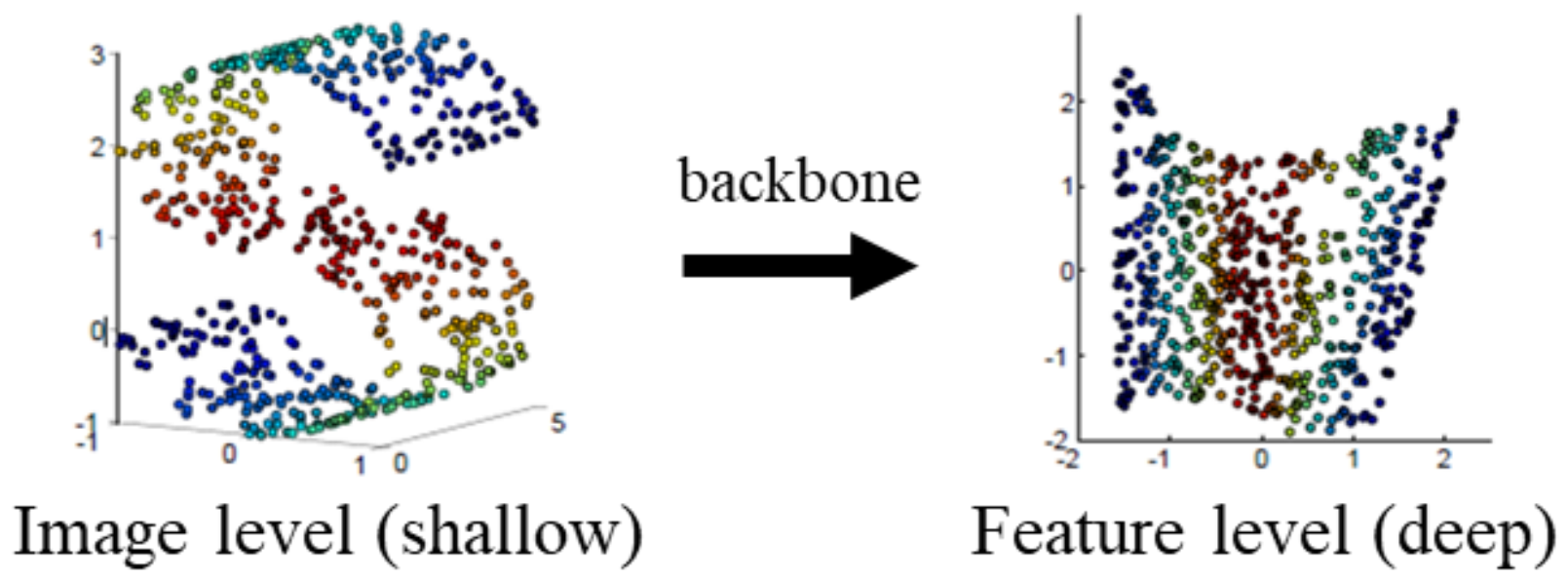

图 4.4 图像空间和特征空间的流型
Figure 4.4 The manifold of images and features

### 4.2.3　空间选择性间隔对比损失

给定两张有着同样语义但来自不同域的图像，可以假定其通过骨干网络提出的隐式

特征都是相同的。本节使用了对比学习的思路设计了空间选择性间隔对比损失（SSMC loss）。具体而言，对于原图像$I_1$和其对应图像$I_2$，空间对比损失（Spatial Contrastive loss，SC loss）的计算方式如下：

$$L_{SC} = \left|\left|F(I_1) - F(I_2)\right|\right|_2^2, \tag{4.21}$$

其中，F 表示骨干网络，$F(I_1), F(I_2) \in \mathbb{R}^{H\times W\times C}$，H，W，C 分别是特征图的高、宽和通道大小。$||\cdot||$表示 L2 归一化。然而，过于受限的正则化会对检测器的辨别能力有负面影响。为了解决这一问题，本节提出了两个解决方法。第一，可以选择那些有着最高变化值的像素进行正则化。因此可以将 SC loss 改进为空间选择性对比损失（Spatial Selective Contrastive loss，SSC loss）：

$$V = \left(F(I_1) - F(I_2)\right)^2, \tag{4.22}$$

$$L_{SSC} = kMaxpooling\left(\frac{1}{C}\sum_j^C V_j\right), \tag{4.23}$$

其中，V 表示方差矩阵，$V_j$是 V 的第 j 个通道，kMaxpooling 可以定义为：

$$\text{kMaxpooling(H)} = \frac{1}{k}\sum_i^k topk(H), \tag{4.24}$$

其中，topk(H)是 H 中最高的 k 的值，$H \in \mathbb{R}^{H\times W}$，$k = (H \times W/16)$。其次，再将间隔概念引入到 SSC loss 中，让所有空间方差都限制在一定的间隔之内，而非直接优化至 0，可以得到空间选择性间隔对比损失（Spatial Selective Marginal Contrastive loss，SSMC loss）：

$$L_{SSMC} = kMaxpooling\left(\max\left(\frac{1}{C}\sum_j^C V_j - \delta, 0\right)\right). \tag{4.25}$$

提出的 SSMC loss 可以正则化那些域相关的像素，并且允许一定的空间以维持网络的辨别能力。SSMC loss 的流程图如图 4.5 所示。

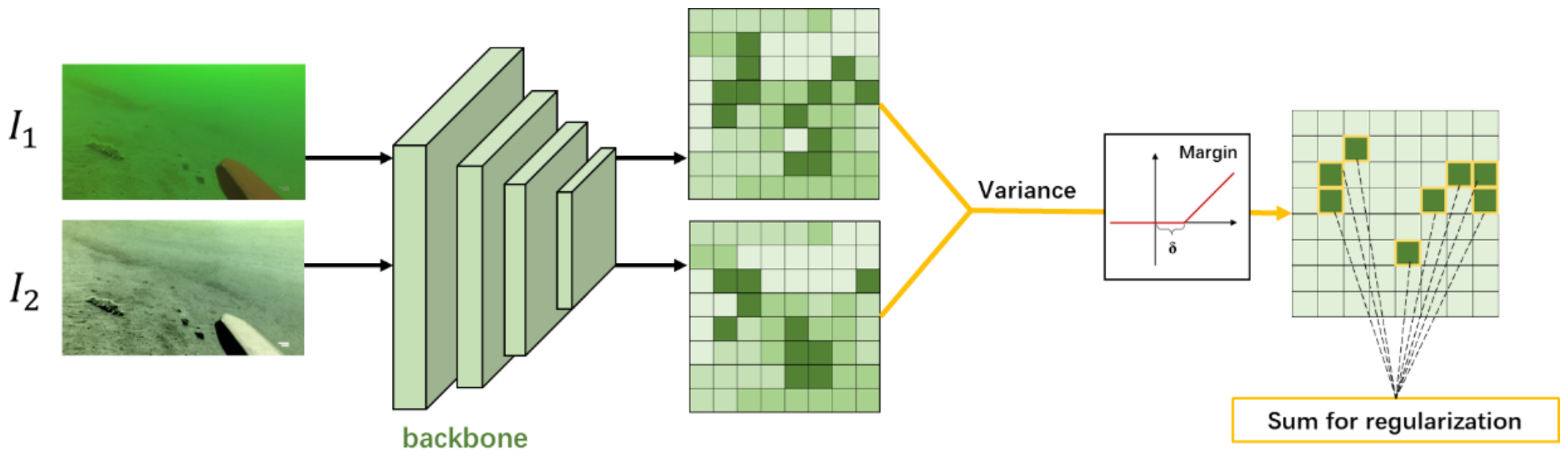

图 4.5 空间选择性间隔对比损失的流程图
Figure 4.5 The pipeline of Spatial Selective Marginal Contrastive loss

整个检测网络的总损失为：

$$\mathrm{L_{total}} = L_{SSMC} + L_{rpn_cls} + L_{rpn_reg} + L_{cls} + L_{reg}, \tag{4.26}$$

其中，$\mathrm{L_{rpn_cls}}$和$\mathrm{L_{rpn_reg}}$表示 RPN 的分类损失和回归损失，$\mathrm{L_{cls}}$和$\mathrm{L_{reg}}$表示 R-CNN 头的分类损失和回归损失。

## 4.3　实验与讨论

### 4.3.1　数据集介绍

本节对本章提出 DMC 的性能进行测试，并测试本章提出的每一个模块的行性能。因为不存在域泛化的检测任务数据集，因此本节提出了一个用于域泛化的水下目标检测的数据：S-UTDAC2020 数据集。并且在另外两个分类任务的域泛化数据集：PACS 数据集和 VLCS 数据集。

**S-UTDAC2020 数据集**是基于 UTDAC2020 数据集（见 2.4.1）生成的。其生成方式仿照 3.1.2 进行生成，其不同点在于：并非将所有数据都生成各种水质，而是先将数据集均分成 7 份，每一份 791 张图片。本节将这 7 份分别用 WCT2 模型转换成 7 个域（type1 到 type7），其中 type1 到 type6 是源域$\mathcal{S}$，type7 是目标域$\mathcal{T}$（如图 4.6 所示）。对于源域数据$\mathrm{x_s} \in \mathcal{S}$，类标签和目标框$\mathrm{b} = [\mathrm{x}, \mathrm{y}, \mathrm{w}, \mathrm{h}]$坐标是可知的。模型应该在源域上训练，而在目标域上进行测试。目标域的数据在训练中完全无法接触。

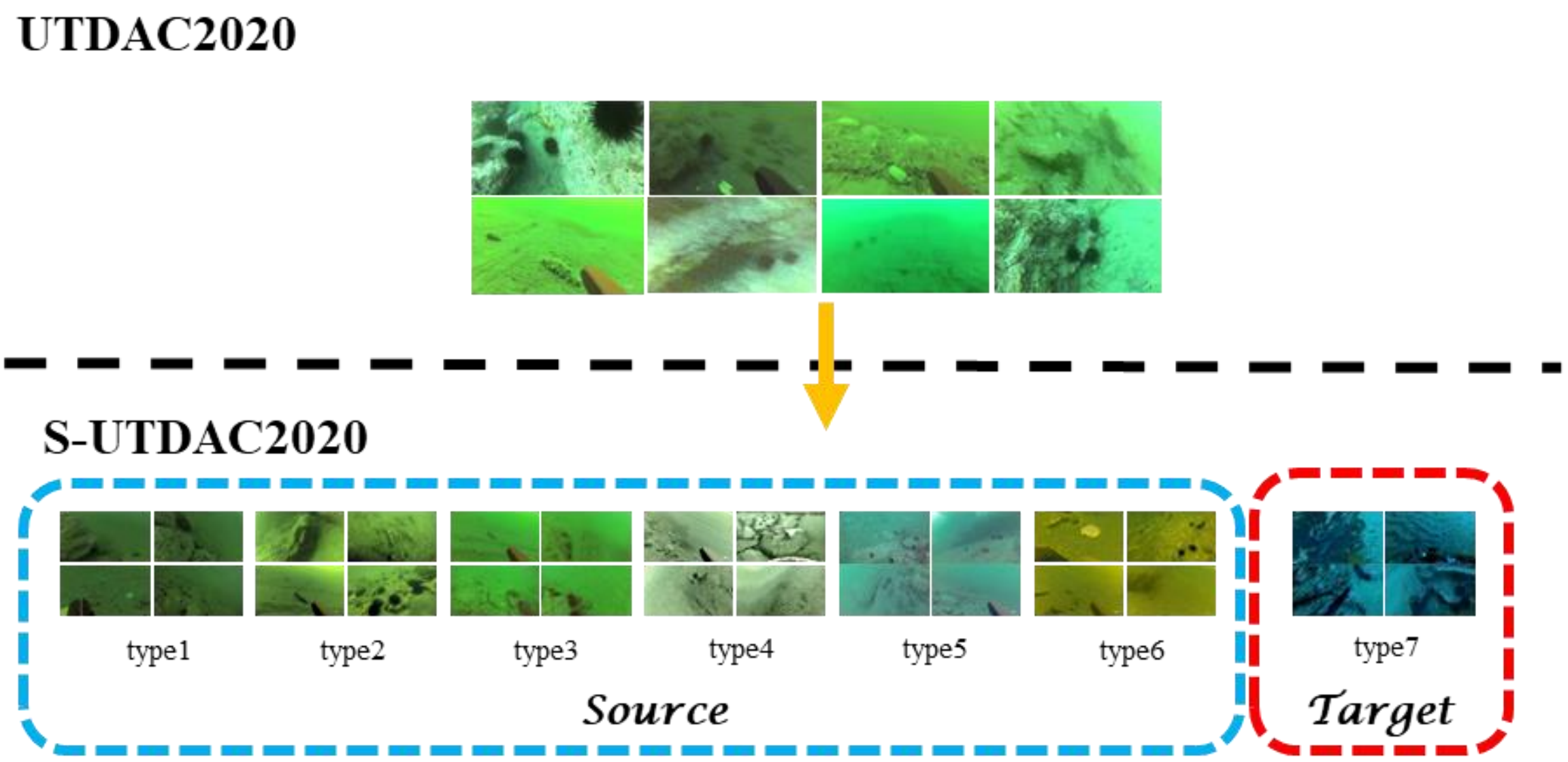

图 4.6 S-UTDAC2020 数据集的介绍

Figure 4.6 The overview of UTDAC2020 dataset

**PACS 数据集**[63]。因为没有其他公开的面向目标检测的域泛化数据集，本节使用传统的面向目标识别的域泛化数据集。PACS 是一个经典的域泛化数据集（如图 4.7），域

之间有着非常大的差异。其包含 7 个类别和 4 个域：照片（Photo）、艺术画作（Art Paintings）、卡通（Cartoon）和草图（Sketches）。每一个域的数据集都被分为训练集、验证集和测试集。测试时，选取三个域作为训练域，剩下的一个域作为测试域，在训练域的训练集中进行训练，训练域的验证集中选取最好结果，再测试其在测试域的测试集性能。

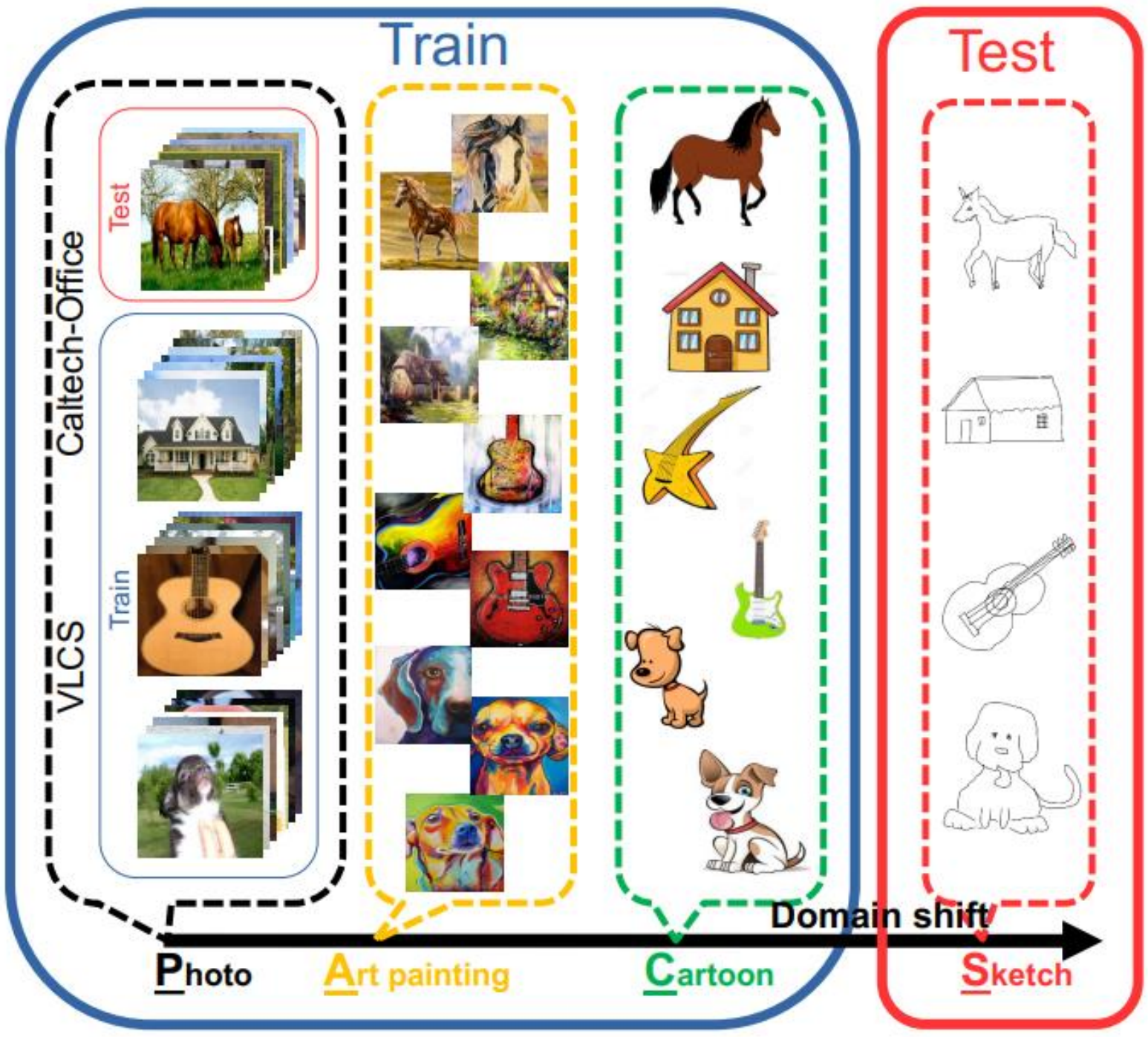

图 4.7 PACS 数据集的介绍[63]

Figure 4.7 The overview of PACS dataset

**VLCS 数据集**也是一个经典域泛化数据集（如图 4.8），其均为现实所采集图像，只是来自不同的四个数据集（Pascal VOC 2007，LabelMe，Caltech 和 Sun），因此域之间的差异性很小。每一个域的数据集都被分为训练集和测试集，本节将原训练集继续切分 10%的比例作为验证集。测试时，选取三个域作为训练域，剩下的一个域作为测试域，在训练域的训练集中进行训练，训练域的验证集中选取最好结果，再测试其在测试域的测试集性能。

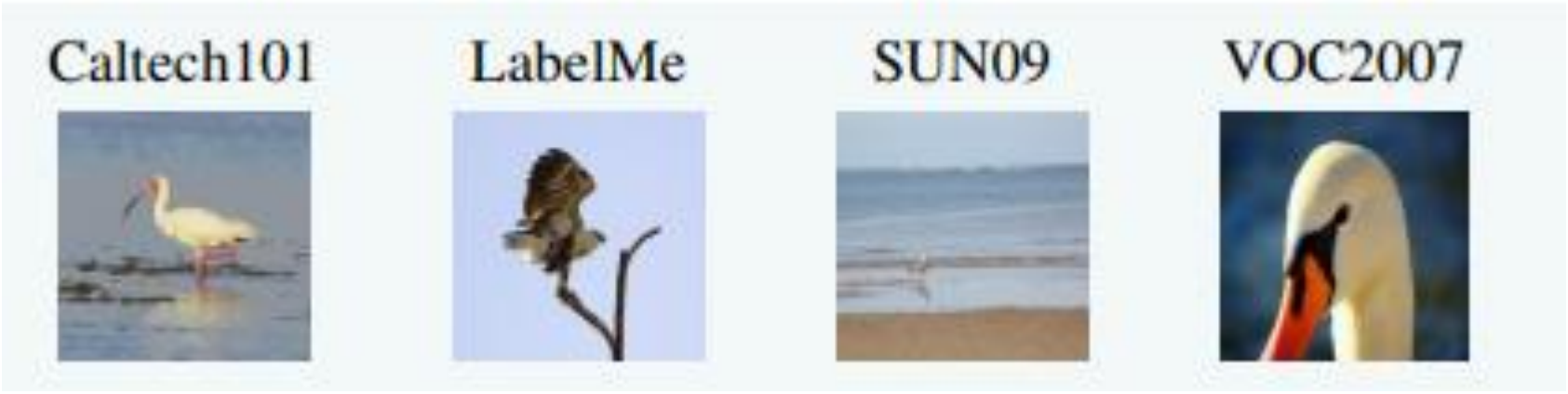

图 4.8 VLCS 数据集的介绍

Figure 4.8 The overview of VLCS dataset

### 4.3.2　在 S-UTDAC2020 数据集实验结果

#### 4.3.2.1 实验设置

本节将 DMC 应用在 YOLOv3 和 Faster R-CNN+FPN 上，这两个是分别是广泛使用的一阶段检测器和二阶段检测器。两者都在 Nvidia GTX 1080Ti GPU 和 PyTorch 环境进行训练。

对于 YOLOv3，模型会以批大小 8 来训练 100 个 epochs。使用 Adam 作为优化器，学习率设置为 0.001，$\beta_1$和$\beta_2$分别被设为 0.9 和 0.999。使用多尺度训练。IoU、置信度和非极大值抑制值分别设为 0.5，0.02 和 0.5。使用累积梯度技术，每两次迭代才进行一次梯度下降。DMX 在 DarkNet 的第 36、46、55 和 71 层进行。SSMC loss 应用到第 71 层。

对于 Faster R-CNN，模型会以批大小 4 来训练 24 个 epochs。使用 SGD 作为优化器，学习率、动量和权重衰减分别设为 0.02，0.9 和 0.0001。IoU、置信度、非极大值抑制阈值分别被设为 0.5、0.05 和 0.5。DMX 在 ResNet50 的最后三个阶段之后进行。SSMC loss 应用在 ResNet50 的最后一层。

对于以上两个模型，除非刻意提及，不采用除了水平翻转以外的数据增强。

#### 4.3.2.2 在 S-UTDAC2020 数据集上的性能比较

因为基本上没有面向目标检测的域泛化方法，本章选择一些面向目标识别的域自适应方法或者域泛化方法，通过进行一些改动使它们能够应用于目标检测上。本章所选择的对比方法有：

**DeepAll** 是最传统的训练范式，仅仅只是简单地在所有的源域上训练检测器。

**DANN**[53]是一个域自适应方法，其使用一个域分类器以区分数据是来自源域的还是来自目标域的，然后进行对抗训练以获得域无关特征。在此任务中，此方法被改动为可以辨别数据是来自哪一个源域。在 Faster R-CNN 中，第二阶段的特征被选择用于对抗训练，对于 YOLOv3 则是第 36 层的特征被用于对抗训练。

**CCSA**[58]设计了一个对比语义对齐损失（Contrastive Semantic Alignment loss，CSA loss）以对齐源域的特征。CSA 原本是特地为分类任务所设计的（其需要类别标签），但其可以很容易地用于 Faster R-CNN 的 R-CNN 上，因为 RPN 和 RoI Align 将分类和回归两个任务解缠了，R-CNN 上进行了单独的分类任务。但其并不适用于 YOLOv3。负提议（背景样本）不参与 CSA loss 的计算，因为其会导致无法收敛。

**CrossGrad**[40]借鉴了对抗攻击防御的思路，借用两个分类器，用域分类器的对抗样本训练目标识别，目标识别器的对抗样本训练域分类器，从而生成一些源域以外的数据。本章将分类损失替换成检测损失。域分类器的设置和 DANN 在同一个地方。

**MMD-AAE[54]**使用对抗训练和 MMD loss 对齐源域的特征。其仅仅只在 Faster R-CNN 上实现（和 CCSA 同样的理由）。R-CNN 上共享的全连接层可以被看作有着 1024 个神经元的编码器，分类器的全连接层对应 R-CNN 的分类器。解码器和两个全连接层的辨别器加进了 R-CNN 之中。

**CIDDG[55]**使用类条件和类先验归一化的域分类器以进行域对抗训练。其仅仅只在 Faster R-CNN 上实现（和 CCSA 同样的理由）。所有域分类器应用在 R-CNN 的共享全连接层。负样本不参与对抗训练。

**JiGEN[78]**设计了一个辅助的拼图任务去增加域泛化的性能。在 Faster R-CNN 中，ResNet 最后一层的特征被用于进行拼图分类任务，在 YOLOv3 中，DarkNet 第 36 层的特征用于拼图分类。YOLOv3 和 Faster R-CNN 的输入尺寸分别为 513×513 和 768×768，便于拼图分割。

**DG-YOLO** 是上一章提出的域泛化方法，使用了对抗训练和 IRM 惩罚项去对齐源域特征。本节将其重新在新设计的 S-UODAC2020 上进行训练。

CCSA，MMD-AAE 和 CIDDG 都只在 Faster R-CNN 的 R-CNN 上实现，这里的检测任务退化成了分类任务。在表 4.1 中，可以得出结论，无论是在 Faster R-CNN 上还是在 YOLOv3 上，本方法相比于其他方法能够得到更高的性能。同时，可以观察到的是，用于图像层面上的方法（DANN、CrossGrad、JiGEN）都比在实例层面上的方法（CCSA、MMD-AAE、CIDDG）相对要高一点。可以推断出，适用于分类的域泛化方法并不一定适用于检测。

#### 4.3.2.3 消融实验

消融实验的结果如表 4.2 所示。如果只使用源域和 CBST 产生的数据进行训练（“Only CBST”），性能将从 48.86%mAP 急剧提升至 58.17%mAP。另外，增加了 DMX 训练范式（“CBST+DMX”）更进一步增加性能到 60.32%mAP。但如果是只在输入图像上做混合（“CBST+Input Mixup”）只会将性能降低到 54.23%mAP。这证明了隐式特征空间更加平坦，因此通过在特征层面上进行插值合成新域是一个更为合理的方式。而随机选择一个层进行混合（“CBST+DMX*”）能够获得 59.30%mAP，相比“Only CBST”有提升，但是相比“CBST+DMX”还是较低。使用域混合后的特征做检测（“Output Before Mixup”）相比于“CBST+DMX”没有性能提升。如果只对主流进行梯度反向传播（“Detach Mixup”），其性能（60.26%mAP）和“CBST+DMX”基本没有区别。

表 4.1 在 S-UTDAC2020 数据集上的实验结果
Table 4.1 Comparisons with other object detection methods on S-UTDAC2020 dataset

| 检测器 | 方法 | 输入尺寸 | 海胆 | 海星 | 海参 | 扇贝 | 平均 |
|---|---|---|---|---|---|---|---|
| YOLOv3 | DeepAll | 416×416 | 70.28 | 31.83 | 27.67 | 41.74 | 42.88 |
| | DANN | 416×416 | 63.67 | 24.01 | 25.78 | 30.28 | 37.32 |
| | CrossGrad | 416×416 | 71.21 | 31.41 | 32.17 | 32.46 | 41.81 |
| | JiGEN | 513×513 | 73.15 | **38.56** | 34.57 | 35.06 | 45.34 |
| | DG-YOLO | 416×416 | 62.74 | 26.83 | 32.84 | 34.54 | 39.24 |
| | **DMC (Ours)** | 416×416 | **73.26** | 36.69 | **43.79** | **59.61** | **53.34** |
| | DeepAll | 1333×800 | 74.79 | 36.59 | 43.12 | 40.94 | 48.86 |
| | DANN | 1333×800 | **78.62** | 42.76 | 50.60 | 43.48 | 53.87 |
| | CCSA | 1333×800 | 76.71 | 36.85 | 40.58 | 37.46 | 47.90 |
| | CrossGrad | 1333×800 | 77.67 | 45.43 | 49.80 | 42.40 | 53.83 |
| | MMD-AAE | 1333×800 | 75.73 | 35.00 | 43.31 | 44.86 | 49.73 |
| | CIDDG | 1333×800 | 76.37 | 39.89 | 42.27 | 43.65 | 50.55 |
| | JiGEN | 768×768 | 76.15 | 39.06 | 50.27 | 41.44 | 51.73 |
| | **DMC (Ours)** | 512×512 | 78.44 | **54.62** | **53.15** | **59.23** | **61.36** |

表 4.2 消融实验
Table 4.2 Ablation Studies

| 方法 | 海胆 | 海星 | 海参 | 扇贝 | 平均 |
|---|---|---|---|---|---|
| DeepAll | 74.79 | 36.59 | 43.12 | 40.94 | 48.86 |
| Only CBST | 76.88 | 49.09 | 50.49 | 56.23 | 58.17 |
| CBST+Input Mixup | 75.56 | 47.72 | 46.43 | 47.19 | 54.23 |
| CBST+DMX* | 76.79 | 51.96 | 51.26 | 57.20 | 59.30 |
| CBST+DMX | 77.11 | 54.73 | 51.70 | 57.74 | 60.32 |
| Output Before Mixup | 76.24 | 50.37 | 51.31 | 56.74 | 58.67 |
| Detach Mixup | 76.49 | **55.10** | 51.70 | 57.74 | 60.26 |
| DMC w. SC loss | **78.85** | 47.53 | 54.83 | 56.66 | 59.47 |
| DMC w. SSC loss | 78.18 | 54.53 | 51.74 | **59.43** | 60.97 |
| DMC (Ours) | 78.44 | 54.62 | **53.15** | 59.23 | **61.36** |

使用完整的 DMC 能够获得 61.36%mAP。如果把 SSMC loss 替换成 SC loss，性能会下降到 59.47%mAP，因为其限制过于严格。相比于 SC loss，SSC loss 放松了一点限

制，因此性能回复到 60.97%。然而，在后续的训练过程，来自不同域的特征已经非常相似，以至于 topk 方差都已经足够小。迫使模型去维持这一限制会降低性能。SSMC loss 使用间隔停掉了对特征的限制，维持了检测器的辨别力，提高了性能。

表 4.3 是混合比率参数的实验。SSMC loss 将不会在这个实验中被使用。如果$\alpha$接近 1，Beta 分布就会更接近于均匀分布。当$\alpha = 2$时，可以达到最佳性能 60.32%mAP，当$\alpha = 0.5$或者$\alpha = 1$时，取得一个相对低的性能，分别为 59.20%mAP 和 59.41%mAP，因为其接近于均匀分布了。因此，可以推断均匀分布采用不适用于 DMX。

表 4.3 混合比率参数α的实验
Table 4.3 Ablation study of the choice of α

| α | 0.1 | 0.5 | 1 | 2 |
|---|---|---|---|---|
| mAP (%) | 60.15 | 59.20 | 59.41 | **60.32** |

#### 4.3.2.4 ResNet50 中域混合层的实验

如表 4.4 所示，实验表明了不同域混合层的选择对于最终检测性能的影响，本实验也不使用 SSMC loss。实验在以 ResNet50 为骨干网络的 Faster R-CNN 上进行。ResNet50 有四个阶段，DMX 可以在任意一个阶段上进行。可以从表 4.4 观察到的是，如果只在最后三个阶段执行 DMX，可以达到最高性能 60.32%mAP。然而，增加 DMX 的层数并不能保证性能的提升。从第四行到第六行可以得出结论，中间深度的层次做 DMX 的效果会比在浅层或者在深层要好。只在第一个阶段后做 DMX 会得到一个相比于在其他阶段更低的性能（58.45%mAP），这更证明了在浅层做域混合并非一个合理的选择。原因如下：（1）深度模型的能力是将结构复杂的图像流形信息在深层之中展平，因此在浅层，域流形还没有足够平坦，因此很难用线性插值去生成新的域；（2）在深层，根据信息瓶颈理论，大多数无关的域信息都被扔掉了，因此域多样性被大大减少，浅层的参数也无法获得数据增强的训练以增强鲁棒性。

#### 4.3.2.5 在更少的域上进行训练

本节多进行了两组实验以验证本方法在更少源域的情况下的域泛化性能。如表 4.5 所示，本文的方法能够在各种设定中超越 DeepAll，而 DANN 在只有两个域的情况下低于了 DeepAll。然而，DMC 的性能仍然随着源域数量的增加而大幅减少，因为 DMX 严重依赖于输入的域多样性以在域流形上采样。

表 4.4 域混合层的实验

Table 4.4 Ablation study of Mixup stage

| 行 | 阶段 1 | 阶段 2 | 阶段 3 | 阶段 4 | 海胆 | 海星 | 海参 | 扇贝 | 平均 |
|---|---|---|---|---|---|---|---|---|---|
| 1 | √ | √ | √ | √ | 77.06 | 54.16 | 52.08 | 57.85 | 60.29 |
| 2 | √ | √ | √ | | 77.38 | 50.45 | **52.82** | 54.76 | 58.85 |
| 3 | | √ | √ | √ | 77.11 | 54.73 | 51.70 | 57.74 | **60.32** |
| 4 | √ | √ | | | 76.66 | 52.62 | 51.59 | **58.73** | 59.90 |
| 5 | | √ | √ | | **77.69** | **55.49** | 50.37 | 57.26 | 60.20 |
| 6 | | | √ | √ | 76.20 | 52.53 | 51.41 | 58.59 | 59.68 |
| 7 | √ | | | | 75.75 | 53.52 | 49.60 | 54.93 | 58.45 |
| 8 | | √ | | | 76.74 | 54.50 | 49.03 | 57.40 | 59.42 |
| 9 | | | √ | | 76.17 | 52.57 | 50.74 | 56.60 | 59.02 |
| 10 | | | | √ | 76.52 | 54.75 | 51.18 | 57.04 | 59.87 |

#### 4.3.2.6 实时风格迁移的实验

本节将 CBST 和另外三种实时风格迁移模型做比较：AdaIN[130]、MST[129]、BST[128]。CBST 在 Nvidia GTX 1080Ti GPU 和 PyTorch 环境下进行训练。其训练数据是 UODAC2020 的训练集。从 S-UODAC2020 的每一个域中随机选取一种图片作为风格图像，这将会采用六个固定的风格图片。训练过的模型将在 UODAC2020 的验证集上进行验证。在训练阶段和推理阶段中，图像的输入尺寸均为 512×512。将 Adam 优化器作为模型优化器，学习率设置为 0.001，$\beta_1 = 0.9$和$\beta_2 = 0.999$。CBST 在批大小 8 的条件下训练 10 个 epochs。对于 MST，训练的细节和 CBST 基本一致。因为 AdaIN 已经开源了，本节决定直接使用其开源模型参数。对于 BST，因其没有开源，本节将重新在和 CBST 的同样的训练设定上重新实现，但其风格图片是随机从 S-UODAC2020 上选取的。

图 4.9 所示的是图像质量比较实验。AdaIN 过于风格化，和真实图像不相似。MST 会将目标模糊，不利于模型的训练。BST 生成的图像已经改变了目标本身的样子，这样会对训练产生不利影响。而所提出的 CBST 能够产生真实的、清晰的图像，同时保证了重要的目标的语义信息。除此之外，表 4.6 所示的是当将这些风格迁移模型应用到 Faster R-CNN 的训练中时的性能表现和推理速度（基于 Nvidia 1080Ti GPU，推理 100 张图像得到的平均速度）。CBST 相比于其他风格迁移模型能够获得最高的性能，这和图像质量比较实验的结论是一致的，而且 CBST 也可以得到最高的推理速度。

表 4.5 更少源域的实现

Table 4.5 The experiments on fewer domains

| 源域 | 方法 | type8 上的 mAP |
|---|---|---|
| 1-6 | DeepAll | 48.86 |
| 1-6 | DANN | 53.87 |
| 1-6 | **Ours** | **61.36** |
| 1-4 | DeepAll | 49.54 |
| 1-4 | DANN | 49.67 |
| 1-4 | **Ours** | **52.95** |
| 1-2 | DeepAll | 42.58 |
| 1-2 | DANN | 41.26 |
| 1-2 | **Ours** | **43.25** |

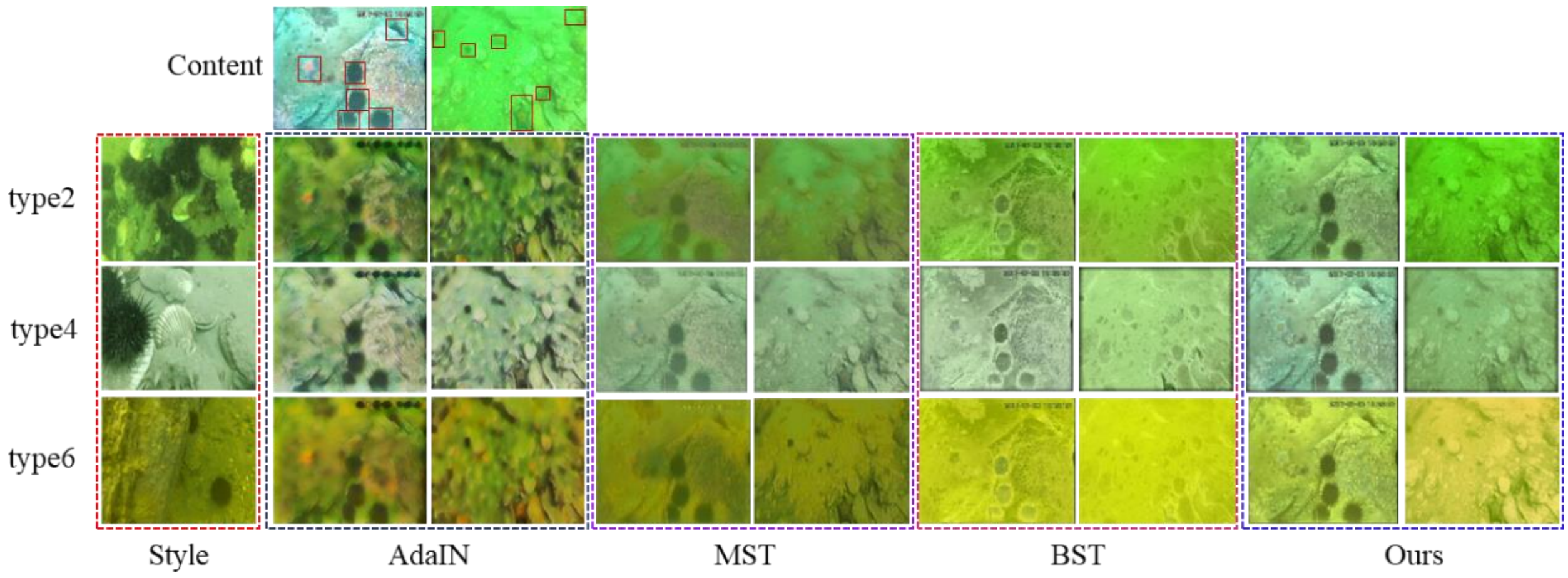

图 4.9 和其他风格迁移方法的质量比较

Figure 4.9 The qualitative comparison of our method against three state-of-the-art baselines on some challenging examples

表 4.6 不同风格迁移模型的性能对比

Table 4.6 The comparison of different style transfer models

| 方法 | MST | AdaIN | BST | CBST |
|---|---|---|---|---|
| FPS | 39.98 | 15.63 | 78.62 | **105.71** |
| mAP | 55.71 | 51.91 | 56.00 | **58.17** |

#### 4.3.2.7 使用 t-SNE 进行特征可视化

本节采用 t-SNE 投影将特征分布可视化。随机从源域和目标域分别选取 300 张和 100 张图像，将他们输入到使用了各种域泛化方法训练的 Faster R-CNN 之中，并将 ResNet50 的最后一个阶段的特征提取出来投射到 2 维平面上（如图 4.10 所示）。蓝色的点代表源域数据，红色的点代表目标域数据。如果红色的点和蓝色的点很好地混合在一起，则可以说明模型更多的捕获到了域无关的特征。从图 4.10 可知，DeepAll、CCSA、CrossGrad、CIDDG 的源域和目标域的特征之间完全分离，DANN、MMD-AAE 的源域和目标域的特征已经较为接近，而 DMC 的源域和目标域的数据完全混合在一起，说明本方法真正地去除了域迁移的影响，而其他方法仍然捕获到了一些错误的相关性。

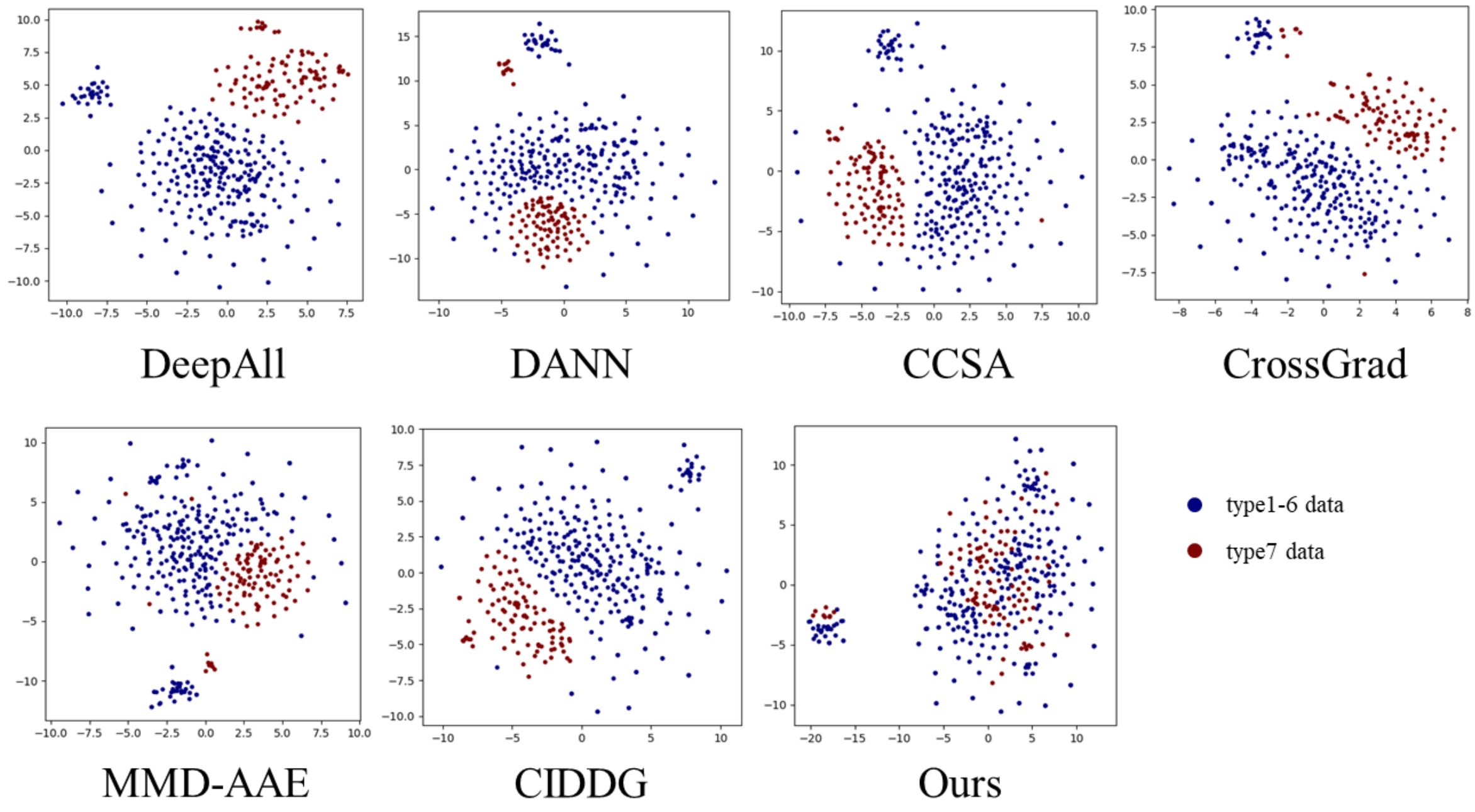

图 4.10 t-SNE 的可视化结果
Figure 4.10 t-SNE visualization of different methods

#### 4.3.2.8 特征统计值分析

大量风格迁移和正则化的工作都揭示了正则参数(γ，β)和风格之间的关联性，这意味着激活值的统计值和域之间也存在关联。本章在每一个域选取 10 张图片，输入到 YOLOv3 之中，提取 12 层到 60 层之间的特征进行统计值分析。对于图 4.11 (a-b)，X 轴表示模型层的编号，左侧的折线图中 Y 轴表示特征图均值的大小，右侧的折线图中 Y 轴表示特征图方差的大小。在(a)中，模型在 7 个域中同时训练，然后分别在不同的域上进行统计值分析。(b)是模型应用所提出的方法进行训练，在各个域上进行分析的结果。如果不同域之间得到的激活值的统计值（均值、方差）相似，那么模型对于域的

变化是鲁棒的。因此每一层的均值和方差在每一层的差异（标准差）展示在(c)中，标准差越小，模型的域泛化性能越好。由(c)图可以看到，由所提出的方法训练出来的模型，均值和方差的波动都是最小的，也进一步证明了所提出的方法能帮助模型学习到域无关的信息。

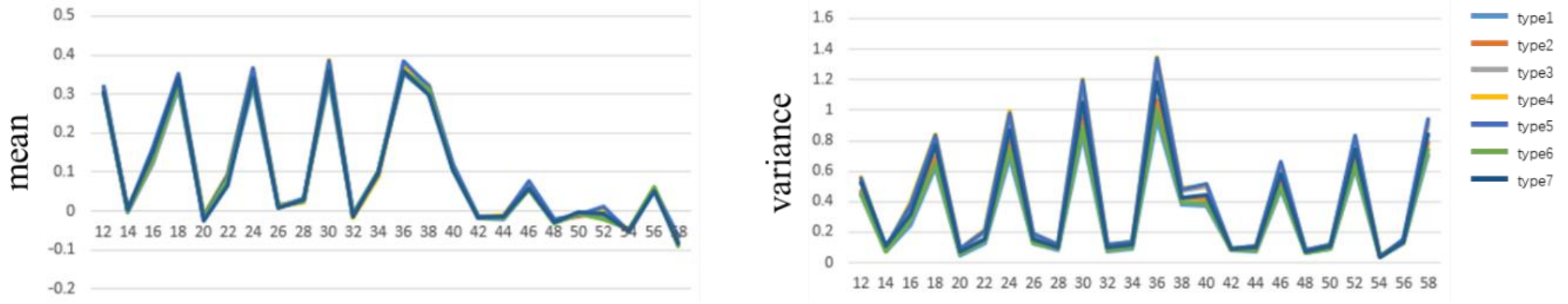

（a） DeepAll

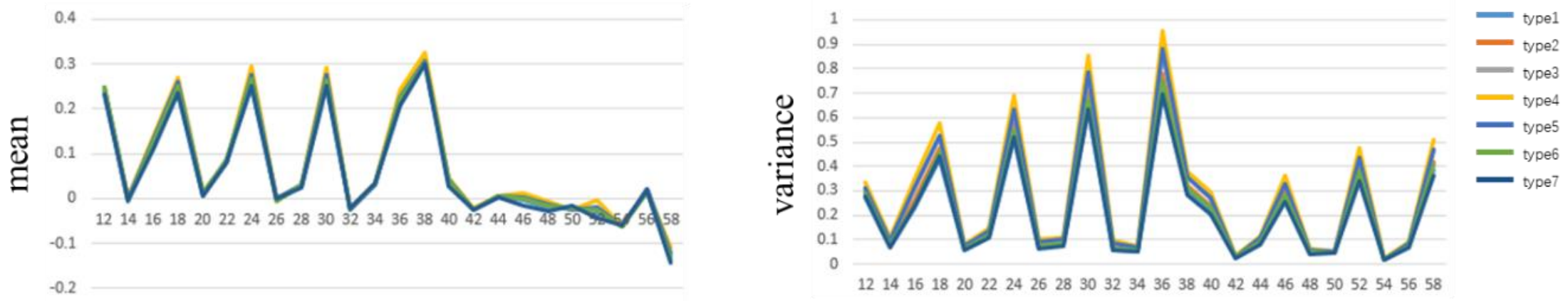

（b） DMC (Ours)

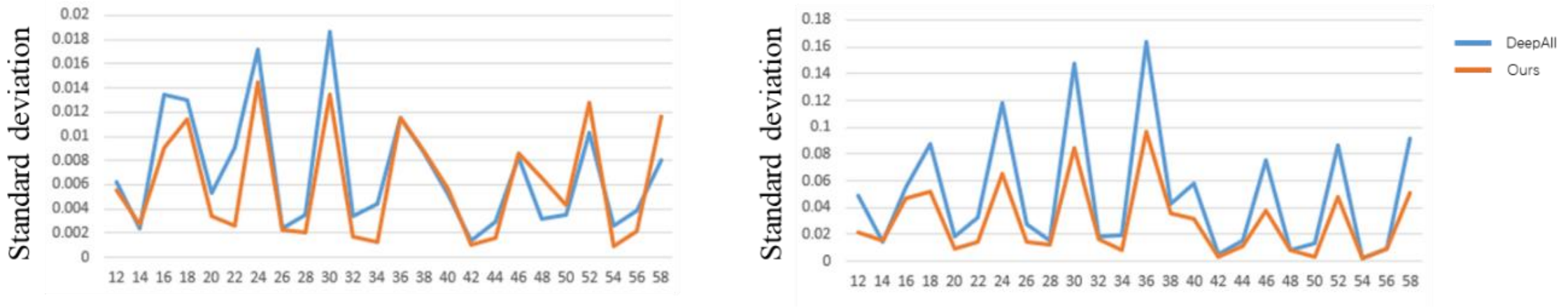

（c） 特征统计值的标准差

图 4.11 特征统计值的可视化

Figure 4.11 Visualization of the feature statistics

(a) DeepAll; (b) DMC (Ours); (c) Standard deviation of features statistics

### 4.3.3 在 PACS 数据集的实验结果

#### 4.3.3.1 实验设置

因为 PACS 的数据并非全是真实图像，为真实图像风格迁移设计的 CBST 不适用于此数据集，因此将 CBST 替换成了 AdaIN，然后为每一个训练图像随机挑选一个不同源域的图像进行变换。本节将在 ImageNet 训练的 ResNet18 上测试本方法的有效性。DMX 应用在最后三个阶段的特征之中。模型以批大小 64 训练 100 个 epochs，使用 Adam 优化器，学习率为 0.0005，且$\beta_1 = 0.9$和$\beta_2 = 0.999$。使用余弦学习率衰减进行训练。SSMC loss 在 ResNet18 的最后一个阶段使用。因为此数据集是一个分类任务，因此采用准确率（公式 2.26）作为性能指标，测试每一个数据集的域泛化准确率及平均

域泛化准确率。

#### 4.3.3.2 在 PACS 数据集上的性能比较

表 4.7 所示的是于其他域泛化方法在 PACS 数据集上的对比结果。参与对比的方法有 MASF[76]、Epi-FCR[81]、D-SAM[69]、JiGEN[78]、MetaReg[75]、EISNet[79]和 L2A-OT[44]。本方法在平均域泛化性能上取得了最高结果 85.14%准确率，虽然本方法在照片和艺术画作上并没有取得最好的性能，但在卡通和草图上取得了最高的性能（78.37%和 85.21%），尤其是在草图上，比 D-SAM 高出了 7.38%准确率。

### 4.3.4　在 VLCS 数据集的实验结果

#### 4.3.4.1 实验设置

和 VLCS 一样，将 CBST 替换成了 AdaIN，然后为每一个训练图像随机挑选一个不同源域的图像进行变换。本节将在 ImageNet 训练的 AlexNet 上测试本方法的有效性。DMX 应用第三个 Relu 操作和最后一层的 MaxPooling 之后。模型以批大小 64 训练 10 个 epochs，使用 Adam 优化器，学习率为 0.0005，且$\beta_1 = 0.9$和$\beta_2 = 0.999$。使用余弦学习率衰减进行训练。在本数据集采用随机裁剪（保留比率为 0.8-1.0 中随机采样）和水平翻转的数据增强方法。SSMC loss 在 AlexNet 的第一个全连接层之前使用。因为此数据集是一个分类任务，因此采用准确率（公式 2.26）作为性能指标，测试每一个数据集的域泛化准确率及平均域泛化准确率。

#### 4.3.4.2 在 VLCS 数据集上的性能比较

表 4.8 所示的是于其他域泛化方法在 VLCS 数据集上的对比结果。参与对比的方法有 CIDDG[55]、CCSA[58]、MMD-AAE[54]、D-SAM[69]和 JiGEN[78]。与在 PACS 上得到的高性能不同，本方法在 VLCS 上并没有达到最优性能，其原因在于本方法的有效性极度依赖于训练数据的域多样性，用此来挖掘隐式的域信息。然而，VLCS 数据集上的数据都是真实照片样子的，其差异来自于采集的地点和时间的不同，因此本方法并不能在域空间上充分采样，因此性能并不卓越。

表 4.7 在 PACS 数据集上的域泛化性能比较
Table 4.7 Domain generalization results on PACS dataset

| 方法 | 照片 | 艺术画作 | 卡通 | 草图 | 平均 |
|---|---|---|---|---|---|
| MASF | 94.99 | 80.29 | 77.17 | 71.69 | 81.03 |
| Epi-FCR | 93.90 | 82.10 | 77.00 | 73.00 | 81.50 |
| D-SAM | 95.30 | 77.33 | 72.43 | 77.83 | 80.72 |
| JiGEN | 96.03 | 79.42 | 75.25 | 71.35 | 80.51 |
| MetaReg | 95.50 | **83.70** | 77.20 | 70.30 | 80.51 |
| EISNet | 95.93 | 81.89 | 76.44 | 74.33 | 82.15 |
| L2A-OT | **96.20** | 83.30 | 78.20 | 73.60 | 82.80 |
| **DMC (Ours)** | 95.08 | 81.93 | **78.37** | **85.21** | **85.14** |

表 4.8 在 VLCS 数据集上的域泛化性能比较
Table 4.8 Domain generalization results on VLCS dataset

| 方法 | Pascal | LabelMe | Caltech | Sun | 平均 |
|---|---|---|---|---|---|
| CIDDG | 64.38 | 63.06 | 88.83 | 62.10 | 69.59 |
| CCSA | 67.10 | 62.10 | 77.00 | 59.10 | 70.15 |
| MMD-AAE | 67.70 | 62.60 | 94.40 | 64.40 | 72.28 |
| D-SAM | 58.59 | 56.95 | 91.75 | 60.84 | 67.03 |
| JiGEN | **70.62** | 60.90 | **96.93** | 64.30 | **73.19** |
| **DMC (Ours)** | 67.52 | **63.49** | 95.52 | **65.78** | 73.08 |

## 4.4 本章小结

本章关注面向水下目标检测的域泛化问题，这是一个少有研究而极为关键的问题。本章提出的域泛化训练范式 DMC 基于两个基本思路：数据采样和对比学习。使用风格迁移模型和域混合可以在域分布上进行采样，增加训练数据的域多样性，而 SSMC loss 可以对域相关信息进行正则化。使用本方法进行训练，检测器可以捕获到域无关的信息，并以此作为预测标准。此外，为了验证本方法的有效性，本章提出了一个正式的用于水下目标检测的域泛化数据集 S-UODAC2020，并且将多个主流的基于分类域泛化方法迁移至一阶段检测器 YOLOv3 和二阶段检测器 Faster R-CNN 中，建立了此数据集的基准。实验中，本方法在 S-UODAC2020 和另外的两个域泛化分类数据集 PACS 和 VLCS

数据集上取得较好的性能。相信本方法可以推动其他域泛化或者域自适应方法的研究，诸如在特殊天气下的自动驾驶（雾天或者雨天）、不同光照条件下的行人重识别（白天和晚上）等。

# 第五章 结论与展望

海洋中拥有着非常丰富的生物资源和矿产资源，对海洋资源的开采对人类的生活有着极其重要的意义。我国拥有极长的海岸线，领海面积巨大，对于和海洋的利用对于我国的生产和发展具有战略性的意义。我国自“十三五”以来，就非常重视海洋资源开发对我国发展的战略性意义，并提出了“海洋强国”的发展理念。由于深海环境的危险性，水下机器人对于海洋研究具有非常重要的意义，搭载了视觉系统的水下机器人能够大大促进海洋研究的发展，其中目标检测技术就是其中的一个关键技术。目标检测是发展时间最久、目前应用最为广泛的深度学习技术之一，其目的在于找到图像中感兴趣的目标的位置，并对此进行分类。然而不同于通用目标检测任务，水下场景的多样性和复杂性，给目标检测任务带来了新的挑战：水质变化、光照变化、遮挡、生物拟态问题。本文针对上述的挑战，从多个方面提出了关键性的技术，旨在提升目标检测任务在多个困难场景下的鲁棒性，实现一种高性能的、通用性的、鲁棒的目标检测器。以下是本文工作的结论和展望。

## 5.1 结论

本文的主要工作可以总结为以下三点：

（1）针对水下场景的极端光照、低对比度、遮挡、生物拟态问题带来的大量困难样本，本文基于困难样本挖掘的思路提出了一种概率型二阶段水下目标检测器 Boosting R-CNN，其构建了一个强力的一阶段检测器作为区域提议网络，大大提升了二阶段模型在第一阶段的识别和先验概率建模的准确率，并从贝叶斯的角度出发，结合一阶段的先验概率和二阶段的类别似然概率，还原边缘分布，使用概率型推理流程，修正最终检测，有效提升模型准确率。最后，提出了一个依据区域提议网络的错误的软采样方法提升再权重，当区域提议网络错误估计了某一个样本的先验概率，提升再权重模块会根据错误的程度放大其在最终损失的大小，集中训练了水下的困难样本，使模型更加鲁棒。

（2）针对水下检测器在不同水质的适应性，提出了一个新的概念：通用性的水下目标检测器，其意味着目标检测器一旦训练结束便可以无缝地在任何水域下实时性地使用。并且提出了阻碍构建通用性的水下目标检测器的障碍：水下数据集的受限导致模型无法获得足够的鲁棒性。本文使用了合成数据揭示了由水质变化导致的模型脆弱性并非是一个无关紧要的问题，并且借助了数据增强、对抗训练和无关风险最小化理

论构建了检测器 DG-YOLO，相比于基线在跨水质情况获得了极大的鲁棒性提升。

（3）基于上一点提出的贡献，本文进一步尝试解决由复杂水下环境带来的域迁移效应，指出在数据集有限数量的域的情况下，深度学习会仅仅只是记住了这些见过的域，而非从这些域中提取中总结性的、预测性的知识，这便会导致域过拟合问题。因此本文基于两个简单的思路：数据增强和对比学习，提出了一种域泛化训练范式。利用风格迁移模型和特征层面上的插值操作在域流形空间的源域构成的域凸包中尽可能多地采样域数据，并且通过一个参数共享的孪生网络，对不同域的数据有选择性地进行正则化，以捕捉域无关的特征，构建一个强鲁棒性的水下检测器。

## 5.2　展望

尽管本文基于水下目标检测和水下域泛化问题的研究取得了一些进展，但也只是在封闭数据集或者合成数据集上取得了较好的性能，而面对真实场景中可能出现的更多挑战和场景，本文仍存在一定的局限性。基于本文研究工作可能存在的局限，针对水下目标检测的域泛化研究工作可以关注以下几个方面展开：

（1）水下目标检测器的设计思路。对于通用目标检测的研究众多，因此大量的研究者以及总结出某些设计指导思路能够大概率在通用目标上提升性能，而水下目标检测的工作大多数还是停留在借鉴通用目标检测技术之上，并没有形成自己的设计思路和指导理论。然而，水下目标检测和通用目标检测在各检测模型上的巨大差异，意味着仅仅只是搬用通用目标检测的技术是不够的。仍需要继续深入地理解目标检测技术，并且和水下的场景结合起来，基于水下场景的特定难点发展出指导性的水下检测器设计思路。

（2）真实场景的水下域泛化数据。本文对于水质变化的模拟是通过数学模型和风格迁移模型所合成的数据与真实采集的数据之间没有差别的假设进行的。然而合成的数据和真实的数据可能仍存在不小的差距，使用构建和使用真实场景的域泛化数据仍然是非常重要的一个工作，能够为未来在水下域泛化任务上的研究做奠定式的基础。

（3）针对检测的域泛化工作。目前大多数的域泛化工作都围绕着目标识别任务进行，本文的工作也指出了用于识别任务的域泛化技术并非完全适用于检测任务之上，因此将过往工作从识别迁移至检测的，并构建一个有挑战、评估标准合理的基准数据集是一个重要的工作。更重要的是，需要研究出一种通用性的，能在各个任务上（识别、检测、分割、生成）都能够有有效的通用性域泛化方法更是亟待研究。

希望通过本文的研究，能够引起相关研究者的对于水下目标检测域泛化问题的重视，同时也希望能够相关领域的研究者提供新的研究思路和研究方法。

# 参考文献

# 攻读硕士学位期间的科研情况

## 已发表的学术论文

[1] Hong Liu, **Pinhao Song** and Runwei Ding. Towards domain generalization in underwater object detection. IEEE International Conference on Image Processing (ICIP), pp. 1971-1975, 2020, October 25-28, Abu Dhabi, United Arab Emirates.

[2] Tao Wang, Hong Liu, **Pinhao Song**, Tianyu Guo and Wei Shi. Pose-guided Feature Disentangling for Occluded Person Re-identification Based on Transformer. Proceedings of the AAAI Conference on Artificial Intelligence (AAAI), 2022, Ferbruary 22 - March 1, Vancouver, Canada.

## 已申请国家发明专利

[1] 刘宏、**宋品皓**、丁润伟、戴林辉，国家发明专利：基于小目标搜索缩放技术的水下目标检测方法和系统，申请号：CN202011096905.5，申请日：2020 年 10 月 14 日

[2] 刘宏、戴林辉、丁润伟、**宋品皓**，国家发明专利：目标检测方法、终端设备及存储介质，申请号：CN202110550672.X，申请日：2021 年 05 月 19 日

## 参与的科研项目

[1] 鹏城实验室项目：水下敏捷机器人协同作业平台

[2] 国家自然科学基金项目：面向混合增强智能运动规划的机器人位姿空间建模方法，No.62073004

[3] 深圳市高等院校稳定支持计划重点项目：面向复杂场景机器人高效作业的混合增强智能，No.GXWD20201231165807007-20200807164903001

## 荣誉奖励

[1] 2019-2020 北京大学优秀科研奖

# 致谢

三年前，我踏入了北大校门，对未来的科研生涯充满了无限的幻想和激情，如同杰克踏上泰坦尼克号的船头，大喊“我是世界之王”，那样的意气风发，那样的跃跃欲试，感慨着儿时要成为科学家的梦想近在咫尺触手可及。

三年过去了，长时间的工作让摧毁了我的肩颈，一次又一次的拒稿打击了我的信心，看不到性能的实验浇灭了我的热情。三年来，一次又一次的，我在半夜里思考我的工作的意义，几近绝望到崩溃，又在第二天当作无事发生来到实验室继续科研工作。我所做的成果，对人类的进步，哪怕有那么一丝毫的意义吗，我的努力是能够被世人所见，还是会终究在时代的潮汐之中，被冲到不留一丝痕迹。但是在浩渺的宇宙、无限的真理以及历史的伟人面前，我不过是蝼蚁般渺小，发出些微不足道的无病呻吟罢了。终究，我还是谦虚谨慎小心翼翼的，将我三年的所有心血凝结成这最后的一篇毕业论文，以奢求其能在历史巨木的年轮上留下自己的微小刻印。

首先，最需要感谢的是我的导师刘宏教授。三年前刘老师毫不犹豫地把我招进实验室，并且在刚开学没到第一周的时候就让我进入了鹏城实验室的水下机器人项目，并让我有机会成为鹏城实验室的一员，和全国各地高校的优秀科研工作者进行交流合作，让我非常感激，没有刘老师把我参与这一项目，就没有这篇论文的存在。刘老师对我们的科研一直保持高要求，强调做科研需要“顶天立地”，一直以最优秀的北大学生的要求来指导和鞭策我们，让我们把目标都放在顶刊顶会上。同时，刘老师要求我们每次讨论班都需要开口演讲英文，这对我的英文学术演讲和交流能力有力极大的提高，为我们在参加学术会议进行学术交流锻炼了语言能力，更为我后来博士申请的面试打下了很好的语言基础。同时，刘老师在我出国申请博士的事宜上给予了极大的支持和帮助。感谢刘老师一直以来的教导和帮助，对于刘老师为我们做的一切，表示深深的感激和敬仰。

感谢丁润伟师姐为实验室尽心尽力的贡献，丁姐负责实验室各种大大小小的事务，是实验室运转的主轴。我们能够没有忧虑的沉浸在科研之中，都是因为丁姐替我们解决了绝大多数的繁琐事务，我们在实验室中如同活在了丁姐建立的温室之中。在生活上，丁姐也给予了我们许多帮助，组织了实验室大大小小的聚会，为我们实验室的生活增添了许许多多的乐趣。

感谢博士后陈阳师姐在科研上和生活上对我的鼓励和帮助，在我人生最黑暗最难过的时间拉了我一把，给予我鼓励和支持，就如同太阳一样照亮了我心里阴暗的角落。正是陈阳师姐对我的支持，才让我有决心和勇气继续出国深造，继续

走科研道路，我才不至于一直沉沦和堕落下去。也感谢师姐能成为我最好的朋友，和师姐一起看电影的日子，是我枯燥的北大生活中的一抹亮色。

感谢已经毕业的博士生石伟师兄、杨冰师姐、黄伟波师兄对我三年来论文写作的指导和帮助。感谢李一迪、吴璐璐、叶汉荣、任斌、宋美佳、管礼思、任丛雅旭、李夏、戴林辉、袁佩佩、魏鹏、华国亮、张琳琳、张倩、孙永恒、王文帅等师兄师姐们的帮助和照顾，感谢一同毕业的同级生陈湛、陈争妍、徐琬璐、苗子凌、王亚伟和直博生李文豪，感谢你们在我刚进入实验室的时候对我的关心和照顾，感谢你们一直以来的帮助和支持。感谢师弟师妹，朱颖、石邢越、周紫惠、王韬、郭天宇、邱峻寅、班渺椐、吴剑兵、蔡家伦、王体、张琬若、游盈萱、郭静文、苏桐，和你们平时一同做超市，一同玩耍，祝愿你们在未来前程似锦，成果斐然。感谢我的舍友张艺明、张文宏、朱子玉，我们每晚都深夜做饭，让我释放了很多压力。感谢和我一起打球的胡妤婕师妹和乔鹏冲师弟，一同去打球的时光真的很快乐。

感谢学院的朱跃生老师、王荣刚老师、邹月娴老师、李革老师、李挥老师、张健老师、陈杰老师，从你们的课程中我打下了坚实的知识基础。感谢学院的辅导老师戴铭志、卢志明、杨柳老师等，你们在疫情期间兢兢业业地通知我们打卡，关心学生安全，为学生处理各项事务，给我们的研究生生涯带来了很多便利。

感谢我的父母一直支持我，从来没有干涉我的任何事情，让我自由地做自己想做的事情。

# 北京大学学位论文原创性声明和使用授权说明

## 原创性声明

本人郑重声明：所呈交的学位论文，是本人在导师的指导下，独立进行研究工作所取得的成果。除文中已经注明引用的内容外，本论文不含任何其他个人或集体已经发表或撰写过的作品或成果。对本文的研究做出重要贡献的个人和集体，均已在文中以明确方式标明。本声明的法律结果由本人承担。

论文作者签名：　　　　　　日期：　　　年　　月　　日

## 学位论文使用授权说明

（必须装订在提交学校图书馆的印刷本）

本人完全了解北京大学关于收集、保存、使用学位论文的规定，即：

- 按照学校要求提交学位论文的印刷本和电子版本；
- 学校有权保存学位论文的印刷本和电子版，并提供目录检索与阅览服务，在校园网上提供服务；
- 学校可以采用影印、缩印、数字化或其它复制手段保存论文；
- 因某种特殊原因需要延迟发布学位论文电子版，授权学校□一年/□两年/□三年以后，在校园网上全文发布。

（保密论文在解密后遵守此规定）

论文作者签名：　　　　　　导师签名：

日期：　　　年　　月　　日